%% file: thesis.tex
\documentclass[12pt]{report}   
\usepackage{graphicx}  
\usepackage[letterpaper, left=1in, right=1in, top=1in, bottom=1in]{geometry}
\usepackage{setspace}  
\usepackage{times}  
\usepackage[explicit]{titlesec}  
\usepackage[titles]{tocloft}  
\usepackage[backend=bibtex, sorting=none, bibstyle=ieee]{biblatex}  
\usepackage{appendix}  
\usepackage{rotating}  
\usepackage[normalem]{ulem}  
\usepackage{textcomp} 
\usepackage{indentfirst} 
\usepackage{amsmath} 
\usepackage[T1]{fontenc} 
\usepackage[utf8]{inputenc} 
\usepackage{newtxmath} 
\usepackage[hidelinks]{hyperref}

\usepackage{multicol}
\usepackage{balance}
\DeclareMathOperator*{\argmax}{argmax}

\usepackage{epsfig} 
\usepackage{booktabs}
\usepackage{float}
\usepackage{url}
\usepackage{makecell}
\usepackage[colorinlistoftodos]{todonotes}
\usepackage{wrapfig}
\usepackage{array,multirow,graphicx}
\usepackage{xspace}

\usepackage[linesnumbered,ruled]{algorithm2e}
\usepackage{graphics}
\usepackage{epsfig} 
\usepackage{gensymb}
\usepackage{multirow}
\newcount\Comments 
\usepackage{array,multirow,graphicx}
\usepackage{color}
\usepackage{wrapfig}
\definecolor{darkgreen}{rgb}{0,0.5,0}
\definecolor{purple}{rgb}{1,0,1}
\newcommand{\kibitz}[2]{\ifnum\Comments=1\textcolor{#1}{#2}\fi}

\bibliography{references}

\AtEveryBibitem{\clearfield{issn}}
\AtEveryBibitem{\clearlist{issn}}

\AtEveryBibitem{\clearfield{language}}
\AtEveryBibitem{\clearlist{language}}

\AtEveryBibitem{\clearfield{doi}}
\AtEveryBibitem{\clearlist{doi}}

\AtEveryBibitem{\clearfield{url}}
\AtEveryBibitem{\clearlist{url}}

\AtEveryBibitem{%
  \ifentrytype{online}
    {}
    {\clearfield{urlyear}\clearfield{urlmonth}\clearfield{urlday}}}

\begin{document}
\doublespacing  

\input{titlePage.tex}


\input{CopyrightPage.tex}

\pagenumbering{gobble}
\input{Abstract.tex}

\pagenumbering{roman}
\setcounter{page}{1} 
\renewcommand{\cftchapdotsep}{\cftdotsep}  
\renewcommand{\cftchapfont}{\normalfont}  
\renewcommand{\cftchappagefont}{}  
\renewcommand{\cftchappresnum}{Chapter }
\renewcommand{\cftchapaftersnum}{:}
\renewcommand{\cftchapnumwidth}{5em}
\renewcommand{\cftchapafterpnum}{\vskip\baselineskip} 
\renewcommand{\cftsecafterpnum}{\vskip\baselineskip}  
\renewcommand{\cftsubsecafterpnum}{\vskip\baselineskip} 
\renewcommand{\cftsubsubsecafterpnum}{\vskip\baselineskip} 

\titleformat{\chapter}[display]
{\normalfont\bfseries\filcenter}{\chaptertitlename\ \thechapter}{0pt}{\large{#1}}

\renewcommand\contentsname{Table of Contents}

\begin{singlespace}
\tableofcontents
\setlength{\cftparskip}{\baselineskip}
\listoffigures
\listoftables
\end{singlespace}

\clearpage

\phantomsection
\addcontentsline{toc}{chapter}{Acknowledgments}
\input{acknowledgements.tex}

\phantomsection
\addcontentsline{toc}{chapter}{Dedication}
\input{Dedication.tex}



\clearpage
\pagenumbering{arabic}
\setcounter{page}{1} 


\titleformat{\chapter}[display]
{\normalfont\bfseries\filcenter}{}{0pt}{\large\chaptertitlename\ \large\thechapter : \large\bfseries\filcenter{#1}}  
\titlespacing*{\chapter}
  {0pt}{0pt}{30pt}	
  
\titleformat{\section}{\normalfont\bfseries}{\thesection}{1em}{#1}

\titleformat{\subsection}{\normalfont}{\thesubsection}{0em}{\hspace{1em}#1}



\input{chapters/introduction}


\input{chapters/related_work}

\input{chapters/part1}


\input{chapters/tandem}


\input{chapters/tandem3d}


\input{chapters/geotact}

\input{chapters/part2}


\input{chapters/semiemg}


\input{chapters/metaemg}


\input{chapters/reciprocal}


\input{chapters/chatemg}


\clearpage
\phantomsection
\addcontentsline{toc}{chapter}{Conclusion}
\input{Conclusion}

\clearpage
\phantomsection 
\titleformat{\chapter}[display]
{\normalfont\bfseries\filcenter}{}{0pt}{\large\bfseries\filcenter{#1}}  
\titlespacing*{\chapter}
  {0pt}{0pt}{30pt}

\begin{singlespace}  
	\setlength\bibitemsep{\baselineskip}  
	\addcontentsline{toc}{chapter}{References}  
	\printbibliography[title={References}]
\end{singlespace}


  



\end{document}

%% file: titlePage.tex

\begin{titlepage}
\begin{center}

\begin{singlespacing}
\vspace*{6\baselineskip}
{\LARGE \textbf{Robot Learning with Sparsity and Scarcity}}\\
\vspace{3\baselineskip}
Jingxi Xu\\
\vspace{18\baselineskip}
Submitted in partial fulfillment of the requirements \\
for the degree of Doctor of Philosophy\\
under the Executive Committee\\
of the Graduate School of Arts and Sciences\\
\vspace{3\baselineskip}
COLUMBIA UNIVERSITY\\
\vspace{3\baselineskip}
\the\year
\vfill

\end{singlespacing}

\end{center}
\end{titlepage}

%% file: CopyrightPage.tex
\begin{titlepage}
\begin{singlespacing}
\begin{center}

\vspace*{35\baselineskip}

\textcopyright  \,  \the\year\\
\vspace{\baselineskip}	
Jingxi Xu\\
\vspace{\baselineskip}	
All Rights Reserved
\end{center}
\vfill

\end{singlespacing}
\end{titlepage}

%% file: Abstract.tex

\begin{titlepage}
\begin{center}

\vspace*{5\baselineskip}
\textbf{\large Abstract}

Robot Learning with Sparsity and Scarcity

Jingxi Xu
\end{center}

\hspace{10mm} Unlike in language or vision, one of the fundamental challenges in robot learning is the lack of access to vast data resources. 
We can further break down the challenge into (1) \textit{data sparsity} from the angle of data representation and (2) \textit{data scarcity} from the angle of data quantity. The data sparsity problem means that there is a large proportion of empty space or non-relevant information in the data we collected. Robotics is the science of interaction. We have a piece of software or an algorithm embodied inside a piece of hardware, and then the robot needs to interact actively with the environment to collect useful information. The sequential manner of such interaction and the lapse between two consecutive actions make robotic data inherently very sparse. On the other hand, the data scarcity issue is that the sheer amount of data we can collect in the domain of interest is very limited. In contrast to the richness and accessibility of text, image, and video data available on the Internet, it is extremely difficult to collect data from physical hardware or humans on a large scale. 

\hspace{10mm} In this thesis, I will discuss my PhD work on two selected domains: (1) \textit{tactile manipulation} and (2) \textit{rehabilitation robots}, which are exemplars of data sparsity and scarcity, respectively. Tactile sensing is an essential modality for robotics, but tactile data are often sparse, and for each interaction with the physical world, tactile sensors can only obtain information about the local area of contact. I will discuss my work on learning vision-free tactile-only exploration and manipulation policies through model-free reinforcement learning to make efficient use of sparse tactile information. On the other hand, rehabilitation robots are an example of data scarcity to the extreme due to the significant challenge of collecting biosignals from disabled-bodied subjects at scale for training. I will discuss my work in collaboration with the medical school and clinicians on intent inferral for stroke survivors, where a hand orthosis developed in our lab collects a set of biosignals from the patient and uses them to infer the activity that the patient intends to perform, so the orthosis can provide the right type of physical assistance at the right moment. My work develops machine learning algorithms that enable intent inferral with minimal data, including semi-supervised, meta-learning, reciprocal learning, and generative AI methods. 

\vspace*{\fill}
\end{titlepage}

%% file: acknowledgements.tex

\clearpage
\begin{center}

\vspace*{5\baselineskip}
\textbf{\large Acknowledgements}
\end{center}

\hspace{10mm} 
First and foremost, I would like to express my deepest gratitude to my advisor, Matei Ciocarlie. He is not only a distinguished researcher and mentor with exceptional vision and taste, but also a compassionate and generous human being. Under his guidance, I have gained countless technical skills, but more importantly, I have learned the values of responsibility, professionalism, and intellectual rigor. The PhD journey can be long and challenging, and I am forever grateful for his support, care, and presence through it all. He is a truly elegant person, and I look up to him a lot.

\hspace{10mm} I would like to thank my co-advisor, Shuran Song, another person who I deeply admire and look up to. She is such a gifted researcher with endless ground-breaking ideas. I couldn't wait to work with Shuran the moment she joined Columbia, and she has always been nice with a welcoming smile. She is one of the most energetic and encouraging people that I have ever met. She makes impactful research look easy and effortless. I am very grateful for her support along my journey in robotics. 

\hspace{10mm} I would also like to thank the person who brought me into robotics research, Peter Allen. I reached out to Peter three months before I even arrived in the U.S. for opportunities to work in his lab as a master's student. He took a chance on me, who had zero research experience with robotics. He led me into the fascinating world of robots. He gave me an opportunity, and I am always grateful.

\hspace{10mm} I would like to thank my committee, Matei Ciocarlie, Shuran Song, Carl Vondrick, Lerrel Pinto,
and Brenna Argall, for reading my thesis, their support, and their feedback. I couldn’t have asked for a better or more varied group of experts to review the work in this thesis.

\hspace{10mm} I am immensely grateful to undertake this journey as a member of the MyHand team. Joel Stein and Dawn Nilsen, our clinical co-PIs, taught me how to conduct research that makes sense for impaired populations. I am so happy to have shared PhDs with Ava Chen and Lauren Winterbottom, who worked on all MyHand projects with me, and to see us grow together as interdisciplinary scholars through this research. I am also grateful to all of the stroke participants for their patience, enthusiasm, and feedback.

\hspace{10mm} This thesis was made possible by the brilliant mentors and collaborators throughout my journey in robotics research. I had a great time working as a research intern at the Boston Dynamics AI Institute under David Watkins, who was also a close mentor during my master's. I would like to thank Dinesh Jayaraman and Nikolai Matni for hosting me as a visiting student at the University of Pennsylvania, and also Leslie Pack Kaelbling and Tomás Lozano-Pérez for hosting me for a summer at MIT. In addition, I am deeply grateful to Iretiayo Akinola, Yinsen Jia, Dongxiao Yang, Siqi Shang, Runsheng Wang, Katelyn Lee, Han Lin, Pedro Leandro La Rotta, Yolanda (Xinyue) Zhu, Bruce Lee, Patrick Meng, Eric Chang, Huy Ha, Cassie Meeker, Junyao Shi, Wenxi Chen, Yuan Qing, David Choi, Shreenithi Navaneethan, Bingyao Du, Luke Hsu, and Zihan Guo for contributing to the work in this thesis. 

\hspace{10mm} The journey would have been much less fun without my lovely lab mates from both Matei and Shuran's groups. Huge thanks to Ava Chen, Eric Chang, Zhanpeng He, Sharfin Islam, Joaquin Bernardo Palacios, Runsheng Wang, Lennart Schulze, Pedro Piacenza, Zhenjia Xu, Cheng Chi, Huy Ha, Samir Gadre, Zeyi Liu, Mandi Zhao, and Mengda Xu.

\clearpage


%% file: Dedication.tex

\begin{center}

\vspace*{5\baselineskip}
\textbf{\large Dedication}
\end{center}

\begin{flushleft}
\hspace{10mm} To Mom and Dad, I will always remain indebted for your love and unwavering support.
\end{flushleft}


%% file: chapters/introduction.tex
\chapter{Introduction}
One of the major challenges in building versatile generalist robots that can handle a variety of complicated tasks is the lack of data. With the advent of data-driven machine learning methods and vast data resources from the Internet, we have witnessed unprecedented progress in vision and language applications, such as ChatGPT~\cite{openai2024chatgpt}, Sora~\cite{sora2024}, Claude~\cite{claude2023}, etc. However, robotics has seen fewer breakthroughs, primarily due to the lack of accessibility to vast data resources such as texts, images, or videos.

\begin{figure}[h!]
	\includegraphics[width=\textwidth]{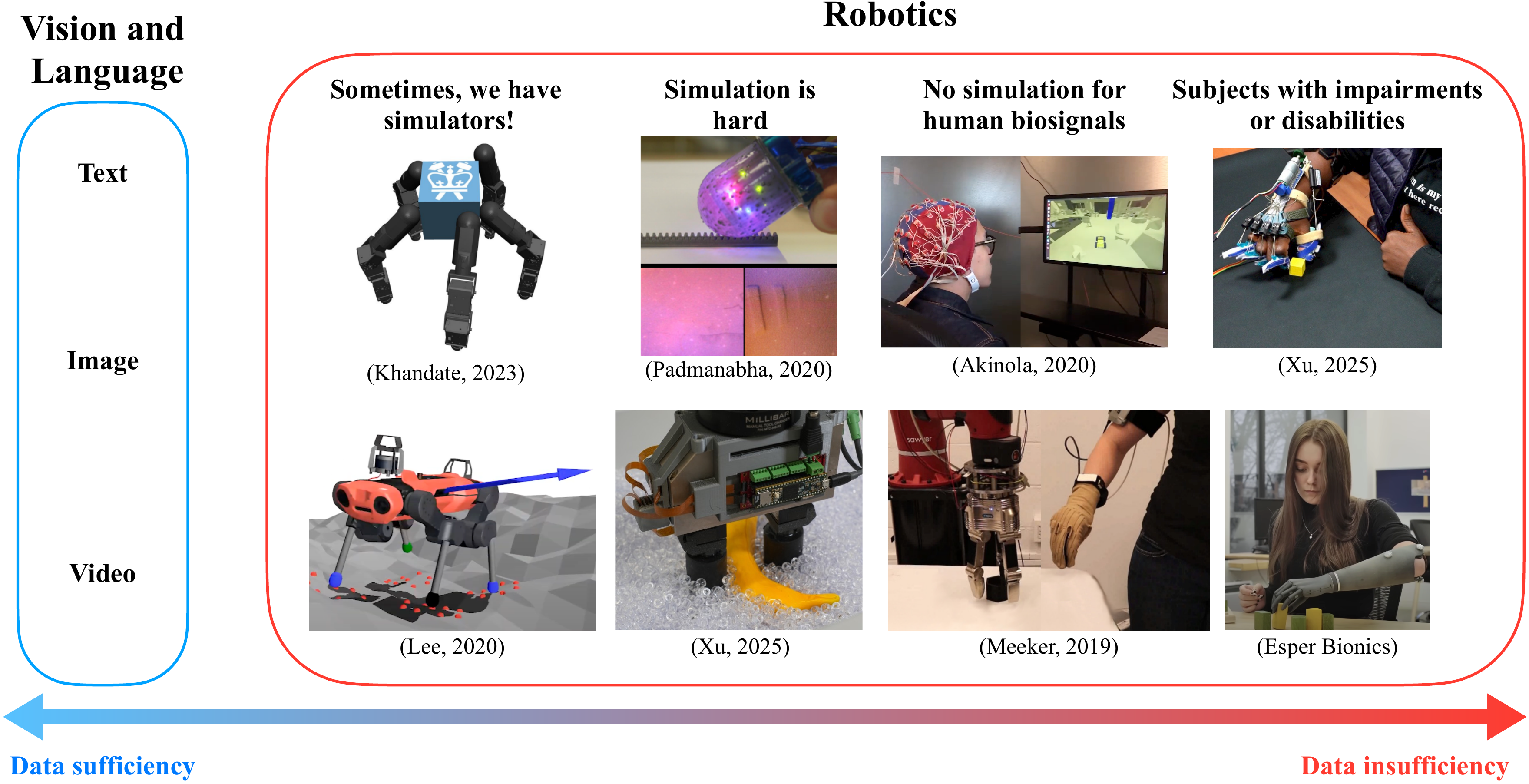}
	\caption{How much data do we have for different AI applications? From left to right, the data goes from sufficient to insufficient.}
	\label{fig:data_spec}
\end{figure}

Fig~\ref{fig:data_spec} shows the data spectrum for different AI applications. In terms of data sufficiency, text, images, and videos are definitely on the relatively more sufficient side, due to the Internet and online social platforms. When we move to robotics, data gets more limited. Robotics often requires customized physical hardware and specialized knowledge to operate them, making data collection on a large scale difficult. However, thanks to the recent progress on physics simulators, if we formulate our task smart enough, we can use simulators, which then enables large-scale data collection~\cite{ha2020learning,akinola2021dynamic,akinola2020accelerated,watkins2020learning,xu2021learned,jia2024dynamic,aoyama2024asynchronously}. But for a wide range of robotic tasks, simulation is hard to build. For example, the simulation of tactile sensing has always been a challenge despite recent pioneering work. For another example, the simulation of granular media, liquid, or deformable objects is traditionally very time-consuming, especially when accuracy is needed, which undermines the original purpose of super-fast simulation speed. 

To make things worse, some robotics tasks are not simulatable at all, such as tasks that involve human biosignals, for which a simulator is completely out of the question. Example applications include brain-computer interfaces, or intent prediction with electromyographic (EMG) or electroencephalographic (EEG) signals. Finally, when our tasks involve people with disabilities, that is the data problem to the extreme. Collecting human biosignals is challenging and time-consuming, let alone collecting them from disabled-bodied subjects. People with disabilities tend to get fatigued very fast, and data collection session often requires the presence of medical specialists, making sessions hard to coordinate and schedule. These hard-to-schedule, precious session time significantly limits the amount of data that we can collect.

We can further break down the data problem into (1) \textbf{\textit{data sparsity}} from the angle of data representation and (2) \textbf{\textit{data scarcity}} from the angle of data quantity. The data sparsity problem means that there is a large proportion of empty space or non-relevant information in the data we collected. Robotics is the science of interaction. We have a piece of software or an algorithm embodied inside a piece of hardware, and then the robot needs to interact actively with the environment to collect useful information. The sequential manner of such interaction and the lapse between two consecutive actions make robotic data inherently very sparse. On the other hand, the data scarcity issue is that the sheer amount of data we can collect in the domain of interest is very limited. Unlike vision and language, which have richer, more transferable, and more accessible data on the Internet, it is extremely difficult to collect data from physical hardware or humans on a large scale.

In this thesis, I will discuss my PhD work on two selected domains: (1) \textit{\textbf{tactile manipulation}} and (2) \textit{\textbf{rehabilitation robots}}, which I believe are representative domains where the data sparcity and scarcity issues are most pronounced, respectively.

\section{Data Sparsity Challenge in Tactile Manipulation}

Tactile manipulation is a task domain where data sparsity brings a lot of challenges. Tactile sensing is a powerful modality for robotic manipulation, but it is at the same time very challenging to use efficiently. Fundamentally, tactile sensing is an active sensing modality. The robot needs to drive the sensor to make active interaction with the physical world in order to obtain useful contact information. Every time the sensor contacts the target object, it can only obtain sensor reading from the local area of the contact, making tactile signals very sparse. 

Efficient exploration and guidance are critical: tactile sensors need to be physically moved by a robotic manipulator to obtain new signals, introducing additional costs for every sensor measurement. Without smart guidance, we can only blindly scan/grope on a surface~\cite{okamura2001feature, allen1985object} or continuously make repetitive and high amounts of contacts at tightly controlled positions~\cite{meier2011probabilistic, allen1988integrating, bierbaum2009grasp, gaston1984tactile, skiena1989problems}. These strategies are extremely inefficient and often incur prohibitively high costs and burdens. Furthermore, it is also important to have an intelligent way to rearrange or encode such local and sparse signals into a global representation.

In this thesis, we aim to address the data sparsity challenge in tactile manipulation, and specifically, we study the task of tactile object identification and tactile object retrieval from granular media. Humans are experts at using tactile sensing efficiently; for example, we can easily identify and retrieve small objects from our pockets, without looking. We want to ask if robots, equipped with tactile sensors, can achieve something similar. Towards that goal, we first study tactile object identification, where the robot interacts and makes contact with the target objects, and identifies the object out of a set of known objects with as few touches as possible. This task provides a great test bench to study the fundamental challenge of data scarcity. We propose TANDEM~\cite{xu2022tandem} (Chapter~\ref{chap:tandem}) and TANDEM3D~\cite{xu2023tandem3d} (Chapter~\ref{chap:tandem3d}), which uses an effective and novel co-training framework to jointly learn a decision-making policy and an exploration policy. With a trained exploration policy using this framework, they can correctly recognize both 2D and 3D objects with very few touches, greatly reducing the cost of moving the robot arm. 

We then take a step further and move to an even more challenging and sensor-deprived environment where we work on retrieving potentially unseen objects completely buried inside granular media. In such a sensor-deprived environment, vision is completely out of the question, and there is huge uncertainty caused by ubiquitous contact with granular media. As a result, the problem of tactile data sparsity is even more pronounced. To tackle this challenging task, we introduce GEOTACT~\cite{xu2024tactile} (Chapter~\ref{chap:geotact}), the first robotic system capable of retrieving objects from granular media relying purely on tactile sensing. We achieve such a system through model-free reinforcement learning in simulation, and we design the action space such that a learned emergent pushing behavior helps reduce uncertainty and funnels the object to a good graspable pose. 

\section{Data Scarcity Challenge in Rehabilitation Robots} 

Rehabilitation robot domain suffers extremely from the problem of data scarcity. Learning on wearable assistive or rehabilitative devices faces a tremendous scarcity of both raw data and reliable ground-truth data~\cite{meeker2017emg,xu2022adaptive,la2024meta,xu2024chatemg,wang2025reactemg,chen2024volitional}. The scheduling for sessions with disabled-bodied people is very time-consuming, and each session often requires the presence of medical experts. Moreover, there is a huge variation in the biosignals of different individuals, making models trained using the data of one patient difficult to generalize to another patient. There is also variation across different sessions and days, even for the same subject (note that a \textit{session} in this thesis is defined to be a single use of the device between donning and doffing). These compounding factors exacerbate the data scarcity challenge even more.

One fundamental application in the rehabilitation robot domain is intent inferral, or the process by which a robotic orthosis or prosthesis collects a set of biosignals from the user, and uses them to infer the activity that the user intends to perform, so it can provide physical assistance at the right moment. 
We collaborate with clinical experts from Columbia Medical School and build a robotic hand orthosis~\cite{park2018multimodal,chen2022thumb,lee2025fabric} (Fig.~\ref{fig:updated_orthosis}) that can help stroke patients open their hands through intent inferral. The stroke population we study in the thesis loses their ability to open their hands, but they can still close their hands voluntarily. Our tandon-driven motorized device is equipped with different sensing modalities, including primarily the armband that can measure forearm electromyographic (EMG) signals. The goal of intent inferral, is to read from the sensing input and predict when the stroke subject intends to open their hands, such that our device can help them extend their fingers, overcoming the muscle spasticity.
Widely considered to be a key problem in assistive and rehabilitative robotics~\cite{beckerle2017}, an effective intent inferral mechanism can be an intuitive way to control a robotic device.

Our series of work studies data-driven machine learning methods for intent inferral for stroke. Machine learning has revolutionized how computers recognize images and understand speech, but it has yet to show a similar effect on rehabilitation robots such as intent inferral, primarily due to the data scarcity challenge. In this thesis, we will discuss our line of work in studying various machine learning paradigms, such as semi-supervised learning, meta-learning, and generative models, to tackle the difficulty of data collection. 

Our first attempt, dubbed SemiEMG~\cite{xu2021learned} (Chapter~\ref{chap:semiemg}), uses semi-supervised learning to learn from unlabeled data, reducing the number of labeled training data needed from stroke patients. This paradigm uses a disagreement-based semi-supervision algorithm to handle intrasession concept drift. \textit{Concept drift} is the phenomenon where the distribution of the measured biosignals changes over time, which can occur between different sessions (\textit{intersession drift}) or gradually within a single session (\textit{intrasession drift}). In order to also handle intersession drift, we then introduce MetaEMG~\cite{la2024meta} (Chapter~\ref{chap:metaemg}), which models intent inferral as a multitask learning problem, and learns an intermediate representation that can fastly adapt to new sessions or subjects with limited new data and training epochs. In addition, we believe the learning between the device and the subject should be bi-directional. While the intent classifier learns from the subject to predict its intent, the subject should also adapt to the behavior of the classifier by producing more distinguishable signal patterns. We call this new paradigm reciprocal learning~\cite{xu2025reciprocal} (Chapter~\ref{chap:reciprocal_learning}), and our preliminary results show that it can lead to more distinguishable EMG patterns and better intent classification performance. 

Despite the recent progress in high-capacity learning models such as Transformers~\cite{vaswani2017attention}, machine learning research in rehabilitation robots still highly relies on low-capacity models such as linear discriminant analysis (LDA) and support vector machine (SVM). High-capacity models are data-hungry, but the data scarcity challenge in this domain makes it difficult to realize the potential of high-capacity models with enough training data. Toward the goal of bringing the full potential of the latest and greatest machine learning advances into the rehabilitation domain, we propose ChatEMG~\cite{xu2024chatemg} (Chapter~\ref{chap:chatemg}), a generative model that can produce context-specific synthetic EMG signals. These synthetic data can be added to the training set of any type of classifier to increase intent inferral accuracy. We are the first to bring the entire pipeline of data collection, synthetic data generation, and classifier training to help stroke subjects with functional tasks such as pick-and-place under clinical supervision.

\section{Contributions}
Throughout my PhD research addressing data sparsity and scarcity challenges, we have made several contributions to the field, resulting in a wide range of publications at various top conferences/journals.
In tactile manipulation, we developed efficient exploration and manipulation policies using model-free reinforcement learning (RL), supported by physics simulators to enable large-scale data collection. Despite sparse tactile signals, our exploration policy strategically selects the next region to explore, reducing the movement cost of robotic arms.
In the context of rehabilitation robotics, our work spans a range of learning paradigms --- from semi-supervised learning to generative AI. Over time, we have progressed from handling only intrasession drift to also handling intersession drift, and from relying on low-capacity models to successfully deploying high-capacity models. 

Specifically, the main contributions of this thesis are listed below.

\begin{itemize}
    \item We propose TANDEM~\cite{xu2022tandem} (Chapter~\ref{chap:tandem}), a novel framework to co-train a tactile exploration policy along with the task-related decision-making module. We demonstrate the effectiveness of this framework on tactile object identification and show that the learned policy can recognize 2D objects with very few actions.
    \item We extend the co-training framework to 3D objects in TANDEM3D~\cite{xu2023tandem3d} (Chapter~\ref{chap:tandem3d}) and we are the first to co-train a 6DOF exploration and decision-making strategy for 3D object tactile recognition. 
    \item We develop GEOTACT~\cite{xu2024tactile} (Chapter~\ref{chap:geotact}), the first robotic system that can grasp and retrieve unseen objects completely buried inside granular media using only touch sensing. 
    \item We propose SemiEMG~\cite{zhou2010semi} (Chapter~\ref{chap:semiemg}), an adaptive semi-supervised learning algorithm that learns from unlabeled multi-modal sensing data to control an orthosis for stroke. 
    \item We propose MetaEMG~\cite{la2024meta} (Chapter~\ref{chap:metaemg}), a novel meta-learning framework and we are the first to formulate intent inferral on stroke subjects as a multi-task learning problem. MetaEMG achieves fast adaptation to a new session or subject with only a few training samples and epochs.
    \item We propose a new learning paradigm called reciprocal learning~\cite{xu2025reciprocal} (Chapter~\ref{chap:reciprocal_learning}) for training EMG-based intent inferral, consisting of interwoven sessions that alternate between updating ML models based on human biosignal generation and encouraging human exploration and adaptation to models with the use of augmented feedback.
    \item We propose ChatEMG~\cite{xu2024chatemg} (Chapter~\ref{chap:chatemg}), an approach for producing synthetic EMG data via generative training on data also collected from stroke patients. We are the first to deploy an intent classification model trained partially on synthetic data for functional orthosis control by a stroke patient under real clinical supervision. 
\end{itemize}

We believe the insights gained extend beyond the two domains discussed in this thesis and are relevant to a wide array of robotics applications where data sparsity and scarcity remain critical and fundamental challenges.



%% file: chapters/related_work.tex
\chapter{Related Work}

\section{Tactile Manipulation}

In order to address the challenge of data sparsity, an efficient tactile exploration policy is critical to guide the exploration to reduce the cost incurred by driving the robotic arms. Moreover, when feasible, simulation serves as a powerful tool for enabling large-scale training with reinforcement learning. In this section, we will discuss the rich literature of various tactile exploration polices, and how previous works make efficient use of local and sparse tactile signals to develop tactile pushing and grasping behaviors. We then focus specifically on the very challenging granular media environment, where vision is not available, and we have to rely heavily on tactile sensing, providing an interesting task for us to study the challenge of using tactile sensors. 

\subsection{Tactile Exploration}

Tactile sensors need an intelligent exploration policy to actively interact with the target object and efficiently gather sparse and local tactile readings. These exploration policies can be roughly divided into heuristic-based and learning-based. The Exploration Policy (EP) is the sequence of exploratory actions the agent executes to gather tactile information. Since tactile information can only be obtained by interacting with the target object, the EP plays a critical role. An effective EP can gather information much more efficiently for the current manipulation task. We divide tactile sensing EPs into three major categories.

\begin{itemize}
    \item \textbf{Passive mode}. The robotic manipulator is fixed, and the human operator hands over the object to the manipulator, often times in random orientations and/or translations to collect tactile data~\cite{schmitz2014tactile, strub2014using, pastor2019using}.
    \item \textbf{Semi-active mode}. The manipulator interacts with the object according to a prescribed trajectory and does not need to react based on sensor data, maybe except being compliant to avoid damage. Examples include poking the object from uniformly sampled directions or grasping it multiple times with a predefined set of grasps~\cite{allen1988haptic, watkins2019multi, meier2011probabilistic}.
    \item \textbf{Active mode}. The manipulator finds the object and explores it reactively in a closed-loop fashion. The exploratory action is a function of current and/or past sensor data. EPs can be heuristic- or learning-based. 
\end{itemize}

In this thesis, we are mostly interested in studying the active EPs, which can then be grouped into heuristics-based and learning-based categories. Some of the most popular heuristic-based exploration policies range from contour following~\cite{martinez2013active, yu2015shape, suresh2020tactile, pezzementi2011tactile} to information gain (uncertainty reduction)~\cite{hebert2013next,xu2013tactile, schneider2009object,driess2017active}. 
Other heuristics to decide the regions of interest to explore include attention cubes~\cite{rajeswar2021touch}, Monte Carlo tree search~\cite{zhang2017active} and dynamic potential fields~\cite{bierbaum2009grasp}. However, while heuristic-based EPs require no training and can reduce the number of actions effectively, they are also sensitive to sensor noise and the performance of a particular heuristic can be task-dependent. In contrast, learning-based EP can be trained with sensor noise, and thus outperforms heuristic-based baselines when such noise is present in the evaluation.

Other works in active EP combine exploration and decision making, whereby a classifier is pre-trained from pre-collected data and used to estimate action quality with Bayesian methods to reduce uncertainty~\cite{fishel2012bayesian, lepora2013active, martinez2017active, kaboli2017tactile, kaboli2019tactile}. Most of these make effective use of high-dimensional or multimodal tactile data. In our work discussed in this thesis~\cite{xu2022tandem,xu2023tandem3d}, we use relatively simple contact signals that allow training an exploration policy through trial and error in simulation, with zero-shot transfer to real robots, eliminating the need for training on physical objects. We achieve high recognition accuracy with relatively few actions, which we attribute in part to the fact that, unlike in previous methods, our discriminator is constantly updated as the exploration policy improves.

\subsection{Tactile Pushing and Grasping}

Pushing is a key element of many manipulator operations. It is a good way to reduce uncertainty in the locations of objects, to move many objects at once, and to move objects that are hard to grasp. Combining both non-prehensile (e.g., pushing) and prehensile (e.g., grasping) manipulation policies has received increasing attention from the community. Brost~\cite{brost1988automatic} presents an algorithm that plans parallel-jaw push-grasping motions for polygonal objects with pose uncertainty, where the object is pushed by one plate towards the second one, and then squeezed between the two. Mason~\cite{mason1986mechanics} investigates the mechanics and planning of pushing in object manipulation under uncertainty. The seminal work of Dogar et al.~\cite{dogar2010push,dogar2012planning} presents a robust planning framework for push-grasping (non-prehensile motions baked
within grasping primitives) to reduce grasp uncertainty.

While the policies in the works above remain largely hand-crafted, other methods~\cite{omrvcen2009autonomous,clavera2017policy} explore the model-free planning of pushing motions to move objects to target positions that are more favorable for pre-designed grasping algorithms. Boularias et al.~\cite{boularias2015learning} explore the use of reinforcement learning for training control policies to select among push and grasp proposals represented by hand-crafted features. However, their method models perception and control policies separately (not end-to-end); it relies on model-based simulation to predict the motion of pushed objects and to infer its benefits for future grasping. Similarly, Yu et al.~\cite{yu2023novel} also separate push planning and grasp planning by training two networks. More recently, some works~\cite{zeng2018learning,yang2021collaborative,deng2019deep,tang2021learning} have been training model-free end-to-end policies using reinforcement learning to push and separate objects in clutter for better grasping. Our method~\cite{xu2024tactile} is also model-free and trained end-to-end, making no assumptions about the shapes or dynamics of the objects. However, unlike all the previous works cited here, ours is the first to demonstrate a pushing behavior for grasping inside granular media.

\subsection{Robot Interaction with Granular Media}

Unlike manipulating objects buried inside granular media, there are many more works focusing on manipulating granular materials directly, such as scooping and
bulldozing~\cite{sarata2004trajectory,millard2023granular,zhu2023few}, plate dragging (soil-tool interaction)~\cite{kobayakawa2020interaction,swick1988model,gravish2014force}, and pouring~\cite{wu2020squirl,yamaguchi2015pouring,matl2020inferring,tuomainen2022manipulation}. In addition to tactile feedback, some works~\cite{clarke2018learning,clarke2019robot} also incorporate visual and audio feedback for scooping and pouring tasks. Other works~\cite{li2013terradynamics,hauser2016friction,zhu2019data} study legged locomotion on granular materials. In terms of hardware design, there are works that use granular materials and particle jamming to design gripper~\cite{brown2010universal,fakhri2023systematic}, manipulator~\cite{cheng2012design}, swarm robots~\cite{karimi2020boundary}, or haptic displays~\cite{brown2020soft,stanley2013haptic}.

In contrast, interacting with objects buried inside granular materials has only recently begun receiving attention
from the research community, but is \textit{only} limited to tactile perception such as object mapping, localization or identification. Jia, et al.~\cite{jia2021tactile} teleoperate a robotic hand with the BioTac tactile sensors to detect contact with a
cylinder fixed inside a bed of granular media. They take advantage of the multi-modal sensors to distinguish between contact with an object embedded in granular media and ubiquitous contact with surrounding granular media. In their follow-up work~\cite{jia2022autonomous}, instead of teleoperation, they develop a tactile exploration policy for efficient localization and mapping of the buried object. Patel, et al.~\cite{patel2021digger} develop a new tactile finger built on top of GelSight technology~\cite{johnson2009retrographic,johnson2011microgeometry} that facilitates penetration in granular media with the help of mechanical vibrations. They use this sensor to identify four simple 2D shapes i.e. triangle, square, hexagon, and circle. More recently, Zhang, et al.~\cite{zhang2023grains} propose a proximity sensing system to detect an object's presence without contact and finally use a Bayesian-optimization-algorithm-guided exploration strategy to localize buried objects. Different from all these aforementioned works, our work~\cite{xu2024tactile} goes beyond tactile perception and focuses on retrieving any buried objects. To the best of our knowledge, ours is the first work to enable a robotic gripper to retrieve buried objects from granular media.

\section{Rehabilitation Robots}

Rehabilitative wearable devices are an important area that has a real impact on people's daily life, especially for people with disabilities; however, there is a significant challenge of data scarcity in this domain. Compared to autonomous robots, human biosignals are even harder to collect, and there is huge variation across different patients and sessions. In this section, we will lay out some previous works on intent inferral for stroke, and various techniques being used in the literature to address the data scarcity challenge in the rehabilitation robot domain. 

\subsection{Semi-supervised Learning for Intent Inferral}

Traditionally, controls for prosthetics and orthotics have been supervised - trained on a relatively small test set and used during a longer session~\cite{castellini2009, powell2013, lee2011}. However, leveraging unlabeled data may make the control algorithm more robust to fatigue, different arm poses and abnormal muscle coactivation. 

Semi-supervised learning has been shown to be an effective control paradigm for hand prosthetics. In this field, semi-supervision uses the assumption that the user will perform gestures at a low frequency to correct the classifier~\cite{jain2012improving, zhai2017self}. Semi-supervised formulations have been described for EMG classifiers, updating based on confidence, or when there were rapid changes in the prediction stream~\cite{sensinger2009adaptive}. Other semi-supervised paradigms for EMG include using a post-processing neural network for prediction correction~\cite{amsuss2013self}, using all new data for updates~\cite{chen2013application}, reinforcement learning~\cite{edwards2016application}, and cycle substitution paired with probability weighting~\cite{zhang2013adaptation}. He et al. showed updating when confidence was high improved classification of wrist movement~\cite{he2012adaptive}. Liu et al. found common characteristics among EMG classifiers  
to eliminate classifier retraining in later sessions~\cite{liu2015towards}.

All of the above works are specific to EMG. Many of them leverage the low-frequency nature of human hand motions in order to correct the classifiers~\cite{sensinger2009adaptive, zhai2017self, jain2012improving}. The above works are designed for prosthetic controls and have only been tested on healthy subjects or amputees. Only three studies look at intersession accuracy~\cite{amsuss2013self, liu2015towards, zhai2017self}.
However, controlling a hand orthosis for stroke is a different problem than controlling prosthetics for amputees, because of the abnormal coactivation in stroke subjects~\cite{miller2012}. Prosthetic controls do not have to contend with this phenomenon. 

There is a family of semi-supervision which we believe could also be applied to rehabilitation robotics. Disagreement-based semi-supervised learning asks multiple learners to collaborate in order to exploit unlabeled data~\cite{zhou2010semi}. Disagreement semi-supervision uses confident learners to train less confident learners, and is well suited to ensemble learning~\cite{javed2005online,kolter2007dynamic,jiang2012semi,wang2006exploiting,xu2021active}. To the best of our knowledge, our work~\cite{xu2021learned} is the first to apply semi-supervised learning to control a hand orthosis for stroke. 

\subsection{Meta-learning for Intent Inferral}

Meta-learning has been shown in biomedical research to be an effective framework to fast adapt high-capacity models to new individuals, reducing the need to collect large-scale subject-specific training data. Different from the group of works on semi-supervised learning discussed before, which requires user-defined heuristics for labeling unlabeled data, meta-learning pretrains the model on offline data in a way that allows it to quickly adapt to a new distribution. In MetaPhys~\cite{prorokovic2020meta}, the authors train a model to predict heart measurements from video data alone, and they use meta-learning to show how the models trained on a group of individuals can be fine-tuned on a smaller dataset to adapt to a new unseen individual. Similarly, both MetaSleepLearner~\cite{banluesombatkul2020metasleeplearner} and Prorokovich et al.~\cite{prorokovic2019} utilize meta-learning to adapt models to new individuals, in the contexts of sleep stage classification and speech recognition, respectively. These works all focus on healthy subjects while we are working with stroke subjects, which is a much more challenging task due to their abnormal muscle activation and spasticity. 

Similar to our work MetaEMG~\cite{la2024meta}, Prorokovich et al.~\cite{prorokovic2020meta} also work with disabled-bodied subjects (amputees), and they employed meta-learning for orthosis recalibration in the context of proportional control. They sought to train a regression model which predicts the strength of the movement generated by the user. In this work, the authors train regressors on data from both healthy subjects and amputees. However, in their work, the training and testing tasks contain the same individuals performing the same movements, which is different from MetaEMG, which seeks to adapt classifiers to new subjects and sessions. To the best of our knowledge, we are the first to use meta-learning to mitigate the burden of data collection needed to adapt high-capacity neural networks on stroke subjects.

\subsection{Generative AI in Biomedical Research}
The lack of accurate and reliable data is not specific to intent inferral, but to the general machine learning research in the medical community. Similar to ours, there is some previous work that studies generative models for synthetic data generation. 
A majority of them~\cite{shin2018medical,hazra2020synsiggan,anicet2020parkinson,coelho2023novel,piacentino2021generating,festag2022generative,yang2023ts} use Generative Adversarial Network (GAN)~\cite{goodfellow2014generative} and its variants. 
Notably, Shin et al.~\cite{shin2018medical} use GAN to generate synthetic MRI images to detect brain tumors. Hazra et al.~\cite{hazra2020synsiggan} propose SynSigGAN with a novel preprocessing stage that generates four types of biomedical signals, including chin EMG signals during sleep for able-bodied subjects.  Zanini et al.~\cite{anicet2020parkinson} use style transfer and DCGANs to generate EMG signals for Parkinson’s Disease. Pinto et al.~\cite{coelho2023novel} use WGAN-GP to generate EMG signals for six basic hand gestures for healthy subjects.

GAN and its variants are not autoregressive and can only generate a fixed length of signals from random seeds. In comparison, our work (ChatEMG~\cite{xu2024chatemg}) can generate an unlimited sequence of EMG signals conditioned on EMG prompts, which are also of arbitrary length. This is extremely useful in applications where personalization is necessary. Stroke subjects exhibit different hand functionality or muscle coactivation patterns, 
and ChatEMG allows us to bias data generation via prompting. 

Most similar to ours, Bird et al.~\cite{bird2021synthetic} train a GPT-2 model to generate EMG signals for hand open/close classification for healthy subjects. However, their method also does not allow conditioning on prompts, and they only study random forest classifiers with healthy subjects. We specifically focus here on stroke patients whose intents are much more challenging to infer
and the signal variations across subjects make conditional data generation from prompts critical.



%% file: chapters/part1.tex


\part{Addressing Data Sparsity in Tactile Manipulation}
\label{part:sparsity}

In the first part of this thesis (Chapters~\ref{chap:tandem}, \ref{chap:tandem3d}, \ref{chap:geotact}), we discuss our effort in addressing the data sparsity challenges in tactile manipulation. While tactile signals are very local and sparse, we show that we can learn efficient exploration strategies to meticulously choose the next area to explore, reducing the cost of driving the robotic manipulator. We also formulate the problem to use easy-to-simulate tactile information so that our work can make extensive use of physical simulators, enabling large-scale training and the zero-shot transfer onto the real hardware. 

In TANDEM (Chapter~\ref{chap:tandem}) and TANDEM3D (Chapter~\ref{chap:tandem3d}), we work on tacitle object identification, we learn an efficient exploration policy through co-training with a decision-making policy, achieving high accuracy in identifying 2D and 3D objects with very few touches. We effectively address the data scarcity challenge by learning a smart guidance of exploration in simulation and also by developing a learned encoder that can rearrange and encode the local and sparse tactile data into a global representation. 

In GEOTACT (Chapter~\ref{chap:geotact}), we move to an even more challenging environment and we work on retrieving objects buried completely under granular media, relying only on tactile sensing. In such a sensor-deprived environment, apart from the data sparsity challenge, there is also huge uncertainty and significant sensor noise. We address these challenges with model-free RL in simulation, and we design our action space such that an intelligent pushing policy can emerge to efficiently explore, reduce uncertainty, and funnel the target object to a graspable pose.  

%% file: chapters/tandem.tex
\chapter{Co-training Exploration and Decision Making for Tactile Object Identification}
\label{chap:tandem}

In this chapter, we discuss TANDEM, which co-trains an exploration policy and decision-making policy for object identification with only touch sensing. 

\section{Motivation}

Using tactile sensors for robotic manipulation tasks in the absolute absence of vision is an example of a robot learning application that needs to address data sparsity. Tactile sensing is fundamentally an active sensing modality, and we need to move the manipulator to interact with the target objects to collect tactile information, which is a costly process, making large-scale data collection difficult. In addition, tactile data is often local and sparse. We need intelligent encoding techniques to fuse such sparse and local information into a global representation. 

\begin{figure}[h!]
    \centering
    \includegraphics[width=0.65\textwidth]{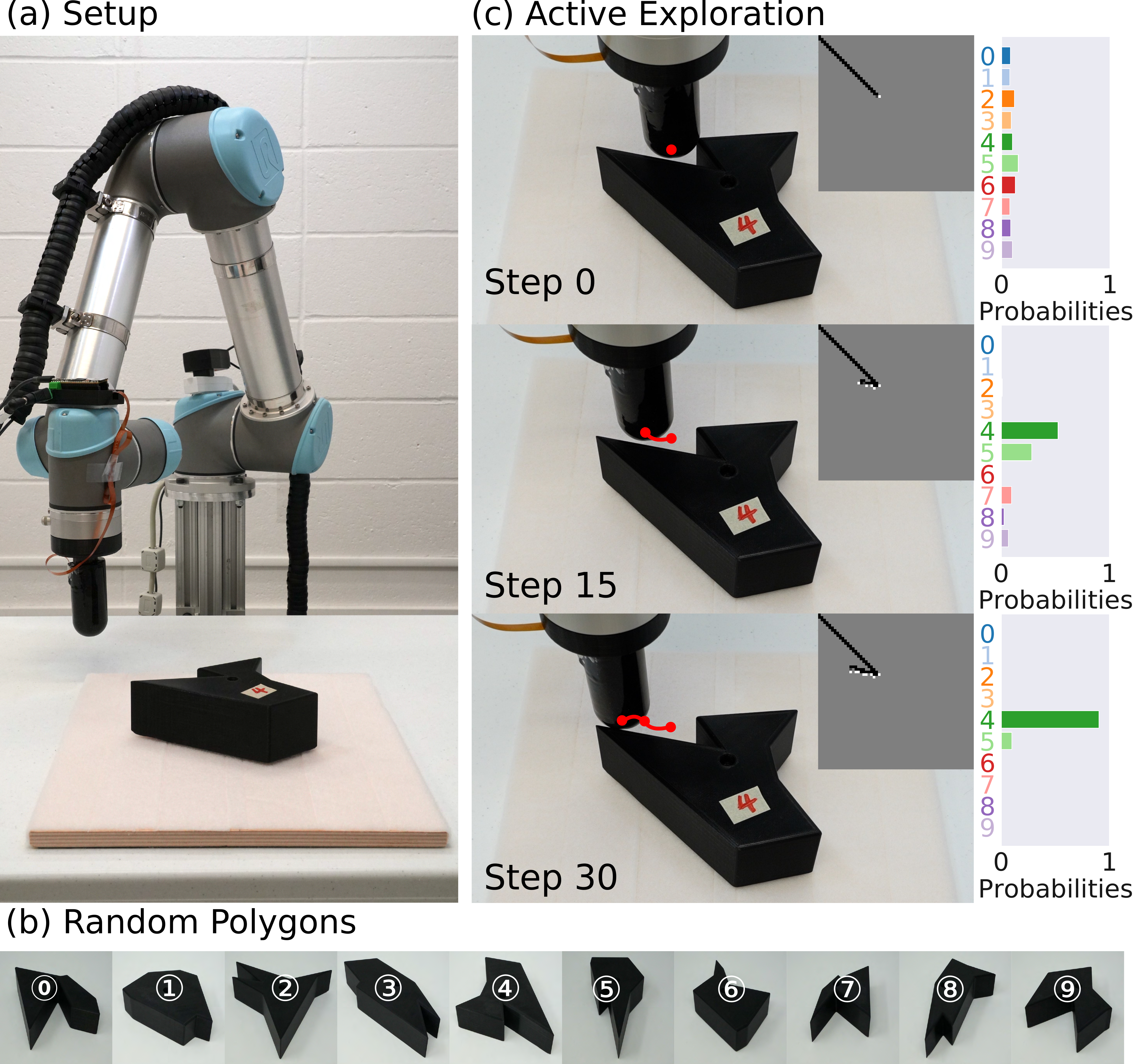}
    \caption{Object recognition based on tactile feedback alone. (a) Real robot setup. Our tactile finger is mounted on a robot arm, and the target object (unknown identity and orientation) is placed roughly around the workspace center. (b) Known object set of 10 randomly-generated polygons. (c) Active exploration. Using our framework, our robot collects data and quickly converges on the correct object identity (object 4 from the set).}
    \label{fig:teaser_tandem}
\end{figure}

In this work, we focus on the process of guiding tactile exploration, and its interplay with task-related decision making. Our goal is to provide a method that can train effective guidance (exploration) strategies. The task we chose to highlight this interplay and to develop our method is tactile object recognition, in which one object must be identified out of a set of known models based only on touch feedback (Fig.~\ref{fig:teaser_tandem}). The goal of our method is to correctly recognize the object with as few actions as possible. 

In order to learn efficient guidance for such tasks, we propose an architecture combining an exploration strategy (i.e. \textit{explorer}) and a discrimination strategy (i.e. \textit{discriminator}). The \textit{explorer} guides the tactile exploration process by providing actions to take; the \textit{discriminator} attempts to identify the target object and determines when to terminate the exploration after enough information has been collected. To convert local and sparse tactile signals into a global representation, we also use an encoding strategy (i.e. \textit{encoder}). In our current version, the \textit{encoder} simply rearranges sparse tactile signals into an occupancy grid, but more complex implementations could be used for future tasks. 


Critically, even though our architecture separates the exploration and decision making, we interleave their training process: we propose a co-training framework that allows batch and repeated training of the discriminator on a set of samples collected by the explorer. We call our method \textbf{TANDEM}, for \textbf{TA}ctile exploration a\textbf{N}d \textbf{DE}cision \textbf{M}aking. 


\section{Method}
Our work aims to develop a framework that combines effective exploration and decision-making when using an active and local sensing modality, such as touch. Our key insight is that exploration and decision-making are distinct, yet deeply intertwined components of such a framework. An ideal exploration strategy will strive to reveal information that the decision-making component can make the best use of. Similarly, a decision-making component will adapt to the constraints of a real-world robot collecting touch data, which can only be obtained sequentially and incrementally. 

The concrete task we develop and test our method on is touch-only object recognition using a robot arm equipped with a tactile finger. We assume a set of known two-dimensional object shapes (randomly-generated polygons). One object is placed in the robot's workspace, in an unknown orientation. The robot must determine the object's identity using only tactile data, and with as little movement as possible. Performance is measured by both identification accuracy and the number of robot movements. 

\begin{figure*}[t!]
\centering
\includegraphics[width=\textwidth]{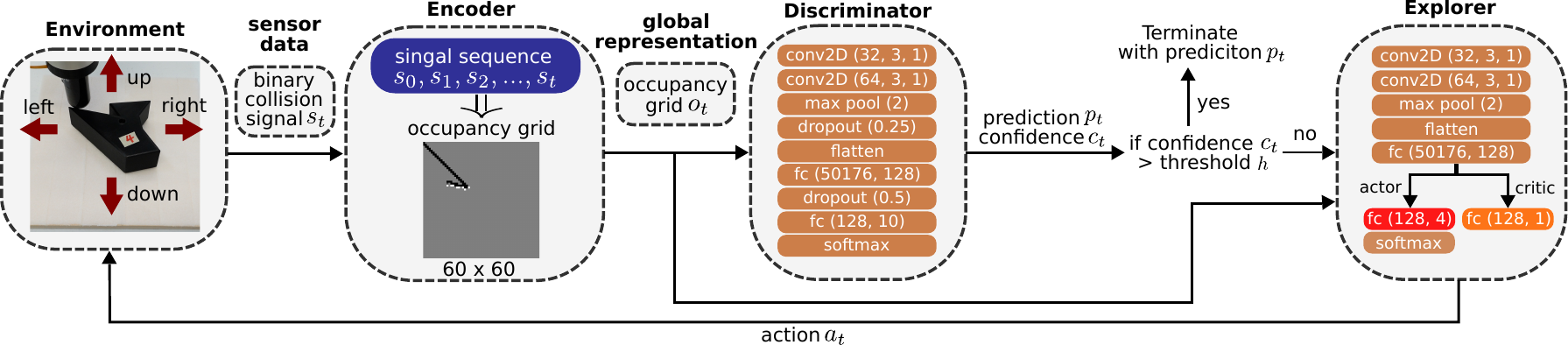}
\caption{An overview of the proposed architecture, and its application to tactile object recognition. The tactile finger interacts with the target object and generates local and sparse sensor data (in this task, binary collision signals). The encoder keeps a history buffer of such sequential signals and converts them into a global representation. Our encoder in this task rearranges them into an occupancy grid image. The discriminator takes in the global representation and attempts to identify the object along with a confidence estimate. If the confidence is higher than a predefined threshold, the exploration is terminated and the final prediction is produced. Otherwise, the explorer reads the representation and generates the next move. The neural networks used by the discriminator and explorer are shown inside their respective block. The parameters of the \texttt{conv2D} layer are the number of filters, kernel size, and stride. The parameters of the \texttt{max pool} layer is stride. The parameters of the \texttt{fc} layer are input dimension and output dimension. The parameter of \texttt{dropout} layer is the probability of an element being zeroed out.}
\label{fig:overview_tandem}
\end{figure*}

Our proposed architecture is illustrated in Fig.~\ref{fig:overview_tandem}. The key components are the following: (1) The \textbf{explorer}, which generates an action for the robot to take in order to collect more data. In our implementation, the explorer consists of a policy trained via deep RL. (2) The \textbf{discriminator}, which predicts the identity of the object, along with a confidence value. This is a supervised learning problem, implemented in this case as a Convolutional Neural Network (CNN). Finally, in addition to the explorer and discriminator, we distinguish one additional component, namely (3) the \textbf{encoder} which converts the sequence of local and sparse tactile signals into a global representation. For our object recognition problem, the encoder simply aggregates binary touch signals into an occupancy grid.

An equally important aspect of the proposed architecture is the training process. While we formulate distinct explorer and discriminator modules, trained via different formalisms (RL vs. supervised learning), we choose to interweave their training processes. This allows us to train the discriminator with data batches gathered by the explorer, which significantly improves data efficiency compared to an all-in-one approach that combines exploration and decision-making into a single component. In the co-training process, the explorer learns to increase the discriminator's confidence as fast as possible, and the discriminator learns to predict object identity based on the type of data generated by the explorer. 

\subsection{Encoder}

The job of the encoder is to maintain a history buffer of the sequence of contact data, convert that history into a global representation, and provide this representation as input to both the explorer and the discriminator. In our current implementation, we use binary signals indicating touch / no-touch. The encoder simply integrates these into an occupancy grid representation of the world, as shown in Fig.~\ref{fig:overview_tandem}. 

All pixels of the occupancy grid are initially grey (unexplored). After each action, if contact is detected, the corresponding pixel is colored white; otherwise, it is colored black. We also use a special value (light grey) to mark the current position of the finger on the grid. Knowing the current location of the finger is useful for the explorer to compute the next action; however, this special color is eliminated when the grid is provided as input to the discriminator because such information is not necessary for predicting the object identity.

For the task addressed here, we believe an occupancy grid works well due to its simple nature, ability to represent geometrical information, and small size in memory. However, when aggregating more complex information (e.g. from tactile sensors providing more than binary touch signals) or for more complex tasks, we expect that different encoding methods will be needed, even while the role in the architecture will be the same. We hope to explore more complex, learning-based encoders for our architecture in future studies.

\subsection{Discriminator}
The discriminator is the component of our pipeline in charge of interpreting sensor data for task-related purposes. Thus, for our problem, its job is to provide a prediction regarding the object identity, along with an associated confidence value. Making a confident prediction also implicitly terminates the exploration. 

In our implementation, underlying the discriminator is a CNN, as shown in Fig.~\ref{fig:overview_tandem}, taking as input the occupancy grid produced by the encoder. The network consists of two convolutional layers followed by a max-pool layer. After the dropout layer, the input is then flattened to go through another two fully-connected layers. A softmax function is applied to the raw 10-dimensional output from the fully-connected layer to generate a probability distribution. The object with the highest probability is chosen as the predicted identity and its corresponding probability is the confidence estimate. If the prediction confidence is greater than a preset threshold, the exploration is terminated and a final prediction is made. Otherwise, the occupancy grid is passed to the explorer to generate the next move.


As part of the co-training process, the discriminator is trained on partially complete occupancy grids, which can be ambiguous over objects, especially when very few pixels have been explored. This ambiguity is in fact the supervision needed to learn a confidence estimate. For instance, if the discriminator data buffer contains multiple duplicates of a highly incomplete grid, each with a different object label, then, in order to minimize the loss, the discriminator network will assign equal probabilities to all candidate objects, thus decreasing the confidence in each individual prediction.

\subsection{Explorer}
The job of the explorer is to generate the next action for the robot, actively collecting additional information. For our task, this means selecting the next move (up, down, left, or right). Tactile data is collected automatically during the move and passed to the encoder as described above.

We implement the explorer as a Proximal Policy Optimization (PPO)~\cite{schulman2017proximal} agent taking the occupancy grid provided by the encoder as input. It has a similar architecture as the discriminator but the last fully-connected layer is replaced by a separate fully-connected layer for both the actor and critic, as shown in Fig.~\ref{fig:overview_tandem}. Even though the discriminator and explorer share part of the same architecture, we found through experiments that keeping the weights separate has a much better performance. This is likely because the discriminator and explorer focus on different aspects of the grid and should learn separate intermediate embeddings. As mentioned earlier, the grid input to the explorer has an extra bit of information providing the current location of the agent.

The reward structure warrants additional discussion. The explorer receives a reward if the discriminator reaches a confidence level that exceeds a preset threshold and thus terminates the exploration. However, the reward for the explorer is not conditioned on the correctness of the prediction. This is in keeping with our tenet of separating the exploration from decision making: the explorer is not aware of prediction correctness and it is rewarded as long as the discriminator is confident enough to make a prediction.

\subsection{Co-training}


While our architecture is constructed around separate discriminator and explorer modules, we find that the interplay and inter-dependencies between the two components make independent training infeasible and suggest a co-training framework. On one hand, training the discriminator requires a labeled dataset with partial observations of object geometry, but the distribution of partial observability highly depends on the exploration policy. On the other hand, training the explorer needs termination signals provided by the discriminator. This termination signal can highly affect the explorer's learning efficiency. Co-training is also important because any pre-trained discriminator will not generalize well as the explorer evolves and implicitly changes the distribution of the data presented to the discriminator. To handle this shift, the discriminator needs to co-evolve with the explorer.

Our co-training process is shown in Alg.~\ref{algo:ours}. Initially, both discriminator and explorer are initialized randomly. We collect an initial data buffer of labeled samples for the discriminator with a randomly initialized explorer. In the co-training loop, we first train the discriminator using the data buffer. Then we fix the discriminator, train the explorer, and, at the same time, push the partially observed occupancy grids collected by the explorer along with their ground truth identities into the data buffer. The updated data buffer is used for discriminator training in the next iteration. 

\begin{algorithm}
\SetAlgoLined
 Initialize discriminator randomly\;
 Initialize explorer randomly\;
 Collect an initial data buffer $\mathcal{D}$ using the explorer\;
 \While{steps $<$ maximum step}{
  Train the discriminator for $N_d$ epochs\;
  Fix the discriminator, train the explorer for $N_e$ steps, and push all occupancy grids (with object identity labels) collected by the explorer into data buffer $\mathcal{D}$\;
 }
 \caption{Co-training Discriminator and Explorer}
 \label{algo:ours}
\end{algorithm}

In this process, the discriminator affects episode termination and the explorer affects partial observability of the labeled training data.
The explorer is rewarded when the discriminator becomes certain and terminates the episode; thus, it learns to make the discriminator confident as quickly as possible. Batch training of the discriminator with samples collected by the explorer also facilitates data reuse and efficiency. Every time one component gets improved, the other component adapts to the distributional shift. Because updates happen with each iteration, this shift is manageable. As a result, the discriminator and the explorer co-evolve, gradually pushing the other to improve and eventually converge.

\section{Experiments}

In this section, we describe our experimental setup, in both simulation and the real world\footnote{For real-world video demonstrations or more information, please visit our project website at \url{https://jxu.ai/tandem}.}. Our method is trained entirely in simulation; it can then be tested either in simulation or on a real robot. We present an extensive set of comparisons against a number of baselines in simulation, then validate the performance of our method on real hardware. 

\subsection{Setup}

Our experiments assume a tactile finger that moves on a 30cm by 30cm plane and is always perpendicular to the plane (Fig.~\ref{fig:teaser_tandem}). The target object is placed roughly at the center of the workspace in any random orientation. The object is fixed and does not move after interaction with the finger. At each time step $t$, the robot can execute an action $a_t \in \mathcal{A} = \{\text{up, right, down, left}\}$ which corresponds to a 5mm translation in the 4 directions on the plane. After each action, the robot receives a binary collision signal $s_t \in \{0, 1\}$, where 0 indicates collision and 1 indicates collision-free. As described above, this information is encoded in an occupancy grid with a 5mm cell size.

In real-world experiments, we use the DISCO finger~\cite{piacenza2020sensorized} as our tactile sensor (Fig.~\ref{fig:teaser_tandem}), but discard additional tactile information (such as contact force magnitude) and only rely on touch/no-touch data. We mount the finger on a UR5 robot arm. For simulation, we use the PyBullet engine and assume a floating finger with similar tactile capabilities. 

Sensor noise is an important consideration since most real-world tactile sensors exhibit some level of noise in their readings, and ours is no exception. It is important for any tactile-based methods to be able to handle erroneous readings without compromising efficiency or accuracy. In particular, we found through empirical observations of our sensors that the chance of an incorrect touch signal being reported is around 0.3\% - 0.5\%. We thus compared all the methods presented below for relevant levels of tactile sensor noise. For learning-based methods, we also have the option of simulating noise during the training process in order to increase robustness; in our case, we simulate a 0.5\% sensor failure rate in the co-training process for our method. 

We generate 10 polygons with random shapes as our test objects, as shown in Fig.~\ref{fig:teaser_tandem}. These polygons are generated by walking around the circle, taking a random angular step each time, and at each step putting a point at a random radius. The maximum number of edges is 8 and the maximum radius for each sampled point is 10cm. We 3D-print these polygons for real-world experiments or use their triangular meshes for the simulated versions. For simulation, we decompose each polygon into a set of convex parts for collision checking.

Each episode is terminated when the confidence of the discriminator is greater than the preset threshold of 0.98 or the number of actions has exceeded 2,000. At termination, the prediction of the discriminator is compared to the ground truth identification of that object to check success. 




\subsection{Baselines}
\label{sec:methods}

In order to evaluate the effectiveness of our learned exploration policy on the tactile object recognition task, we choose to compare our approach to learned all-in-one (without separating exploration and discrimination) and non-learned (heuristic-based) baselines. The metrics that we are most interested in are the number of actions and the success rate in accurately identifying the objects. The methods we evaluate are as follows: 

\begin{itemize}
    \item \textit{Random-walk}. This method generates a random move at each step. A discriminator is trained with this exploration policy for object identification and terminating exploration. We apply a 0.5\% sensor failure rate during training. 
    \item \textit{Not-go-back}. Similar to \textit{Random-walk}, except that the random move generated at each time step is always to an unexplored neighboring pixel.
    \item \textit{Info-gain}. This method uses the info-gain heuristics: it also picks an action that leads to an unexplored pixel, but, unlike \textit{Not-go-back} which picks it randomly, it picks the action that provides the most salient information. At time step $t$, let $\mathbf{p}$ denote the probability distribution over 10 objects predicted by the discriminator on the current grid. Let $\mathbf{p_w}$ and $\mathbf{p_b}$ denote the new probability distributions if the newly explored pixel turns out to be white and black respectively, after applying a particular action. Then the action $a_t$ is chosen by:
    \begin{equation*}
        a_t = \argmax_{a \in \mathcal{A}} \: \bigg\{ \mathcal{H} (\mathbf{p}) - \left(\frac{1}{2} \mathcal{H} (\mathbf{p_w}) + \frac{1}{2} \mathcal{H} (\mathbf{p_b})\right) \bigg\}
    \end{equation*}
    where $\mathcal{H}$ denotes the entropy of a probability distribution. It uses entropy as a measure of uncertainty and picks an action that provides the most information gain (reduces the most uncertainty). A discriminator is trained and we apply a 0.5\% sensor failure rate during training. 
    \item \textit{Edge-follower}. This method uses the popular contour-following heuristic as the exploration policy. A discriminator is trained in this method but we do not apply sensor noise during training. We notice that when applying sensor noise during training, the performance of the \textit{Edge-follower} drops significantly. This is because \textit{Edge-follower} can sometimes get trapped at locations where a collision-free pixel is identified as collision and starts circling that pixel. In such a case, unlike other methods such as \textit{Random} and \textit{Not-go-back}, the \textit{Edge-follower} can not keep exploring with random actions. Thus, the discriminator trained in \textit{Edge-follower} becomes unnecessarily cautious but its exploration policy is not able to increase its confidence. 
    \textit{Edge-ICP}. This method uses the same exploration policy as \textit{Edge-follower}. However, instead of training a learning-based discriminator, it uses the Iterative Closest Point (ICP) algorithm. The occupancy grid is converted to a point cloud using the center location of each pixel. The discriminator runs ICP to match the point cloud to each object using 36 different initial orientations evenly spaced between [0\degree, 360\degree]. For each object, the minimum error among all orientations represents the matching quality. If the error is smaller than 0.0025cm then the object is marked as a match. The output probability distribution assigns equal probabilities to the matched objects and zeroes to not-matched ones. There is no training required for this method.
    \item \textit{PPO-ICP}. This method trains a PPO explorer using the ICP discriminator as in \textit{Edge-ICP}. A 0.5\% sensor failure rate is applied during training. 
    \item \textit{All-in-one}. This method does not separate explorer and discriminator. It has the same structure as the PPO explorer proposed in our approach except that the action space has been expanded to 14 actions. The first 4 actions correspond to a move and the remaining 10 actions correspond to a prediction. If a prediction is made, the episode is terminated. A reward of 1 is given only when the episode is terminated and the prediction is correct. A 0.5\% sensor failure rate is applied during training. 
    \textit{TANDEM}. This is our proposed method. 
\end{itemize}

\begin{figure}[h!]
\centering
\includegraphics[width=0.65\textwidth]{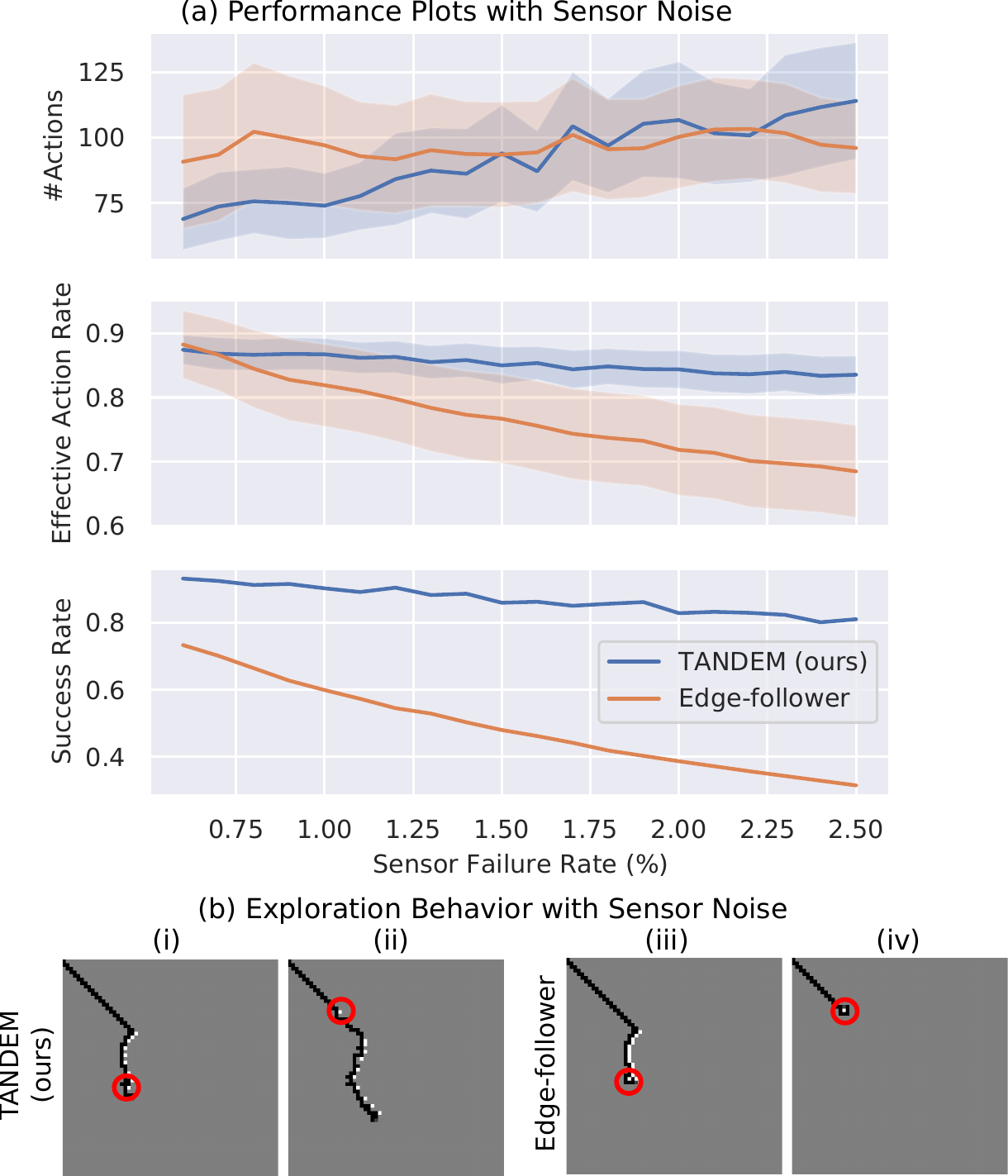}
\caption{(a) Performance of \textit{TANDEM} and \textit{Edge-follower} as the sensor failure rate increases from 0.6\% to 2.5\%. For \#Actions, $\pm$ 0.1 standard deviation is shaded. For Effective Action Rate, $\pm$ 0.2 standard deviation is shaded. With higher sensor noise, both methods need more actions. However, \textit{TANDEM} retains a high success rate and action efficiency while those of \textit{Edge-follower} deteriorate continuously. (b) Exploration behavior of \textit{TANDEM} and \textit{Edge-follower} when sensor failure happens. The location of the sensor failure is circled in red (in the simulation we can ensure it occurs at the same location for both methods). (i)(iii) show a sensor failure after contacting object 1, and (ii)(iv) show a sensor failure before contacting object 5. For these two examples, \textit{Edge-follower} makes the wrong prediction with 39 and 6 actions while \textit{TANDEM} correctly identifies the objects with 38 and 79 actions respectively.}
\label{fig:noise_demo}
\end{figure}

\begin{figure*}[t!]
\centering
\includegraphics[width=\textwidth]{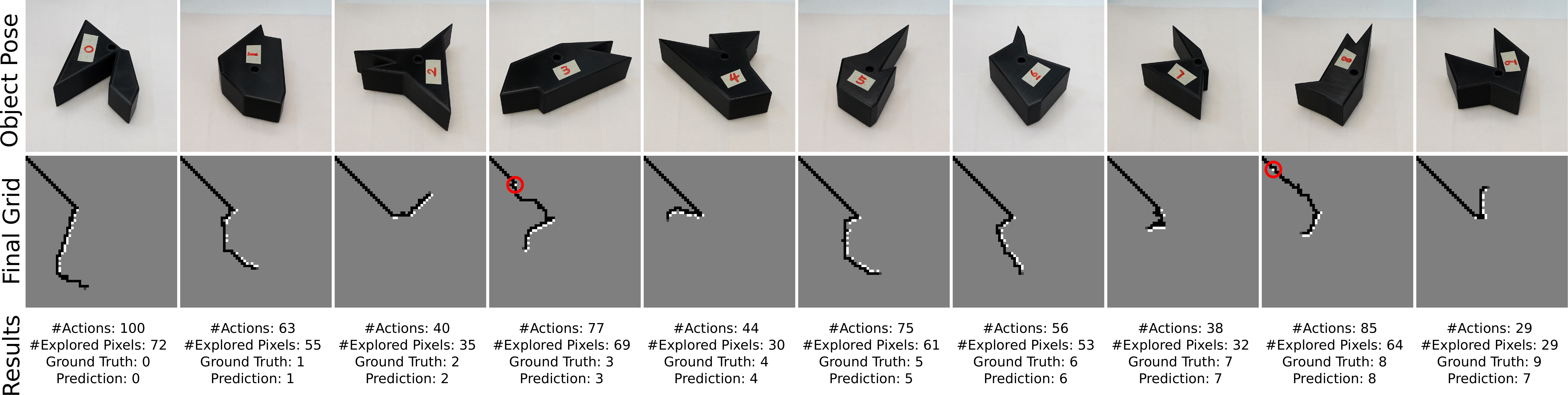}
\caption{10 examples of our method on real robot experiments. The top row shows the object poses, the medium row shows the occupancy grids at termination, and the last row shows the results for each trial. The first 9 examples are successful and the last one is a failure case. While sensor noise can happen anywhere in a trial, it is easier to identify when it occurs before the contact. We highlight in red circles such sensor noise for objects 3 and 8. Our method is able to bypass the noisy pixel,  continue exploring and make the correct prediction.}
\label{fig:real}
\end{figure*}


\section{Results and Discussion}

\begin{table*}
    \footnotesize
    \setlength\tabcolsep{2.5pt}
    \label{tab:results}
    \centering
    \begin{tabular}{c|ccc|ccc}
    \toprule
    \multirow{2}{*}{\textbf{Methods}} & \multicolumn{3}{c|}{\textit{0.1\% Sensor Failure}} & \multicolumn{3}{c}{\textit{0.5\% Sensor Failure}} \\
    & \#\textbf{Actions} & \textbf{\#Explored Pixels} & \textbf{Success Rate} & \#\textbf{Actions} & \textbf{\#Explored Pixels} & \textbf{Success Rate} \\
    \midrule
    Random-walk & 1427 $\pm$ 654.8 & 354.8 $\pm$ 148.9 & 0.31 & 1350 $\pm$ 667.4 & 338.3 $\pm$ 148.5 & 0.27 \\
    Not-go-back & 684.5 $\pm$ 565.9 & 466.6 $\pm$ 320.4 & 0.49 & 621.4 $\pm$ 524.7 & 427.9 $\pm$ 293.8 & 0.43 \\
    Info-gain & 435.1 $\pm$ 397.5 & 341.7 $\pm$ 250.3 & 0.45 & 365.1 $\pm$ 360.6 & 291.2 $\pm$ 232.2 & 0.42 \\   
    Edge-follower & 60.05 $\pm$ 218.6 & 33.01 $\pm$ 15.95 & 0.91 & 95.24 $\pm$ 282.5 & 32.48 $\pm$ 32.81 & 0.75 \\
    Edge-ICP & 136.1 $\pm$ 339.1 & 72.29 $\pm$ 16.78 & 0.94 & 400.6 $\pm$ 719.4 & 75.63 $\pm$ 41.35 & 0.81 \\
    PPO-ICP& 921.2 $\pm$ 679.1 & 286.2 $\pm$ 189.6 & 0.35 & 860.4 $\pm$ 698.3 & 231.7 $\pm$ 172.4 & 0.31 \\
    All-in-one & 28.63 $\pm$ 207.8 & 3.827 $\pm$ 6.735 & 0.23 & 66.05 $\pm$ 328.0 & 6.229 $\pm$ 15.15 & 0.22 \\
    TANDEM (ours) & 54.97 $\pm$ 106.5 & 44.74 $\pm$ 37.32 & 0.96 & 64.76 $\pm$ 109.3 & 49.71 $\pm$ 36.27 & 0.95 \\  
    \bottomrule
    \end{tabular}
    \caption{Comparative performance of various methods in simulation under 0.1\% and 0.5\% sensor failure rate. For each method, we present the number of actions taken (\#Actions) and the number of pixels explored  (\#Explored Pixels) before making a prediction, as well as the success rate in identifying the correct object (Success Rate). Mean and standard deviation over 1,000 trials are shown. A detailed description of each method can be found in Sec.~\ref{sec:methods}.}
\end{table*}

\subsection{Comparative Performance Analysis} We compare all methods described above over a large set of simulated experiments, as shown in Table~\ref{tab:results}. 

For both sensor noise levels we consider, \textit{TANDEM} outperforms the baselines in terms of both success rate and the number of actions required. Only \textit{All-in-one} uses fewer actions at 0.1\% sensor noise but at the price of an extremely low success rate. 

We attribute this gap in performance to multiple factors. For example, while \textit{Random-walk} or \textit{Not-go-back} are clearly inefficient exploration strategies, \textit{Info-gain} is a popular heuristic-based method and has been shown to be efficient in other contexts by previous works. However, we found it to not work well in conjunction with a CNN discriminator. Compared to other methods, the \textit{Info-gain} explorer is more dependent on the discriminator because the discriminator affects not only the termination of each episode but also the action selection at each time step. For the \textit{Info-gain} explorer to be effective, it likely requires a discriminator with high accuracy to begin with, which our method does not. The \textit{All-in-one} method, which is not equipped with a dedicated discriminator, cannot train decision-making directly using the labeled samples collected by the explorer, leading to inefficient training and much worse performance if given the same amount of training time as \textit{TANDEM}.

Edge-following, unsurprisingly, is an efficient exploration heuristics for our task, given its 2D nature. \textit{Edge-follower} and \textit{Edge-ICP} have the best performance among all baselines. However, they are shown to be very sensitive to sensor noise, in terms of both accuracy and efficiency. To further investigate this aspect, we compared \textit{TANDEM} and \textit{Edge-follower} for sensor failure chance further increased up to 2.5\%. As shown in Fig.~\ref{fig:noise_demo}, despite being trained with a fixed 0.5\% sensor noise, \textit{TANDEM} maintains a high success rate even in the presence of more noise. We also report the Effective Action Rate (EAR) in this experiment, where EAR is computed as \#Explored Pixels / \#Actions per episode, a metric reflecting the effectiveness of the move in exploring new locations. We can see that the actions generated by our method maintain high exploration efficiency as shown by the EAR plot. In comparison, both EAR and success rate drop as the sensor failure rate increases for \textit{Edge-follower}. Both methods need longer episode lengths to handle larger sensor noise. Two examples of exploration behavior under noise are shown in Fig.~\ref{fig:noise_demo}. \textit{Edge-follower} makes the wrong prediction for both examples while \textit{TANDEM} successfully handles both. This is due to \textit{Edge-follower}'s discrimination policy overfitting to the edge-following behavior and not being able to explore further after being trapped at an incorrect collision signal.

Unlike \textit{Edge-ICP}, \textit{PPO-ICP} struggles to achieve similar performance. ICP needs a sufficient number of points to achieve decent recognition accuracy and terminate the exploration because it is not able to utilize non-collision pixels. While the edge-following policy is good at collecting points through constantly touching the object, the PPO explorer struggles at learning similar behavior because of the extremely sparse termination reward provided by ICP.


\subsection{Real-World Performance}

\begin{table}
    \footnotesize
    \setlength\tabcolsep{4.5pt}
    \label{tab:real}
    \centering
    \begin{tabular}{c|ccc}
    \toprule
    \textbf{Method} & \#\textbf{Actions} & \textbf{\#Explored Pixels} &  \textbf{Success Rate} \\
    \midrule
    TANDEM & 67.33 $\pm$ 23.47 & 53.95 $\pm$ 18.16 & 0.90 (27/30) \\  
    \bottomrule
    \end{tabular}
    \caption{Real robot experiment results (mean and standard deviation over 30 trials).}
\end{table}

We validate the performance of \textit{TANDEM} on a real robot. We run 3 trials for each of 10 objects with random orientations (30 trials total), with results shown in Table~\ref{tab:real}.

Our method still achieves a high identification accuracy, even if slightly lower when compared to simulation results at a 0.5\% sensor failure rate. Exploration efficiency, as illustrated by the number of actions, is at similar levels. We attribute the sim-to-real gap to imperfections in our noise models, shape printing, and robot control.

Fig.~\ref{fig:real} shows ten examples of \textit{TANDEM} in operation, one for each object in a random orientation, also showing the occupancy grid at the moment that a decision is made. This decision is correct 90\% of the time despite the limited nature of the information collected by that point. We also note that our method is robust enough to handle sensor noise, even before making first contact (objects 3 and 8). We also show a failure case where our method incorrectly recognizes object 9 as object 7: both these polygons have a large opening triangle, which makes them hard to distinguish when this area is under contact. Our learned exploration policy is often similar to edge-following, but has the added ability to handle sensor noise, and also learns to take shortcuts when appropriate and take advantage of non-collision pixels for discrimination: the discriminator terminates the episode at a non-collision location for object 0.

\section{Chapter Summary}

In this work, we address the data sparsity problem through methodically choosing the next action to guide tactile exploration. We co-train such an exploration policy in conjunction with a decision-making policy, and we demonstrate this framework on tactile object identification tasks. Through co-training, both components affect each other, improve each other, co-evolve, and eventually converge. 

Identifying objects through poking, with sparse tactile data such as sparse contact locations, is challenging even for humans, when visual input is not provided at all. It provides a great platform to study the fundamental challenge of data sparsity in tactile manipulation. With the proposed co-training framework, the exploration policy trained entirely in simulation can zero-shot transfer to the real robot and can identify objects with very few actions.

%% file: chapters/tandem3d.tex
\chapter{Extending the Co-training Framework to 3D Objects}
\label{chap:tandem3d}

In this chapter, we discuss TANDEM3D that extends the co-training framework proposed in TANDEM to 3D object recognition. 

\section{Motivation}

While TANDEM co-learns an exploration policy with decision making for recognizing 2D objects and is shown to be able to outperform state-of-the-art baselines despite the local and sparse tactile signals, it does not scale to 3D problems. Firstly, TANDEM uses binary collision signals, rearranges them into an occupancy grid, and then encodes the grid with convolutional neural networks (CNNs). The direct extension to 3D would be voxel grids and 3D convolution; however, the size of voxel grids grows cubically with the size of the workspace, which is not memory efficient. Secondly, the tactile sensor moves in a 2D horizontal plane, limiting the ability of the robotic manipulators to move flexibly and freely on a 3D surface.

\begin{figure}[h!]
    \centering
    \includegraphics[width=0.65\textwidth]{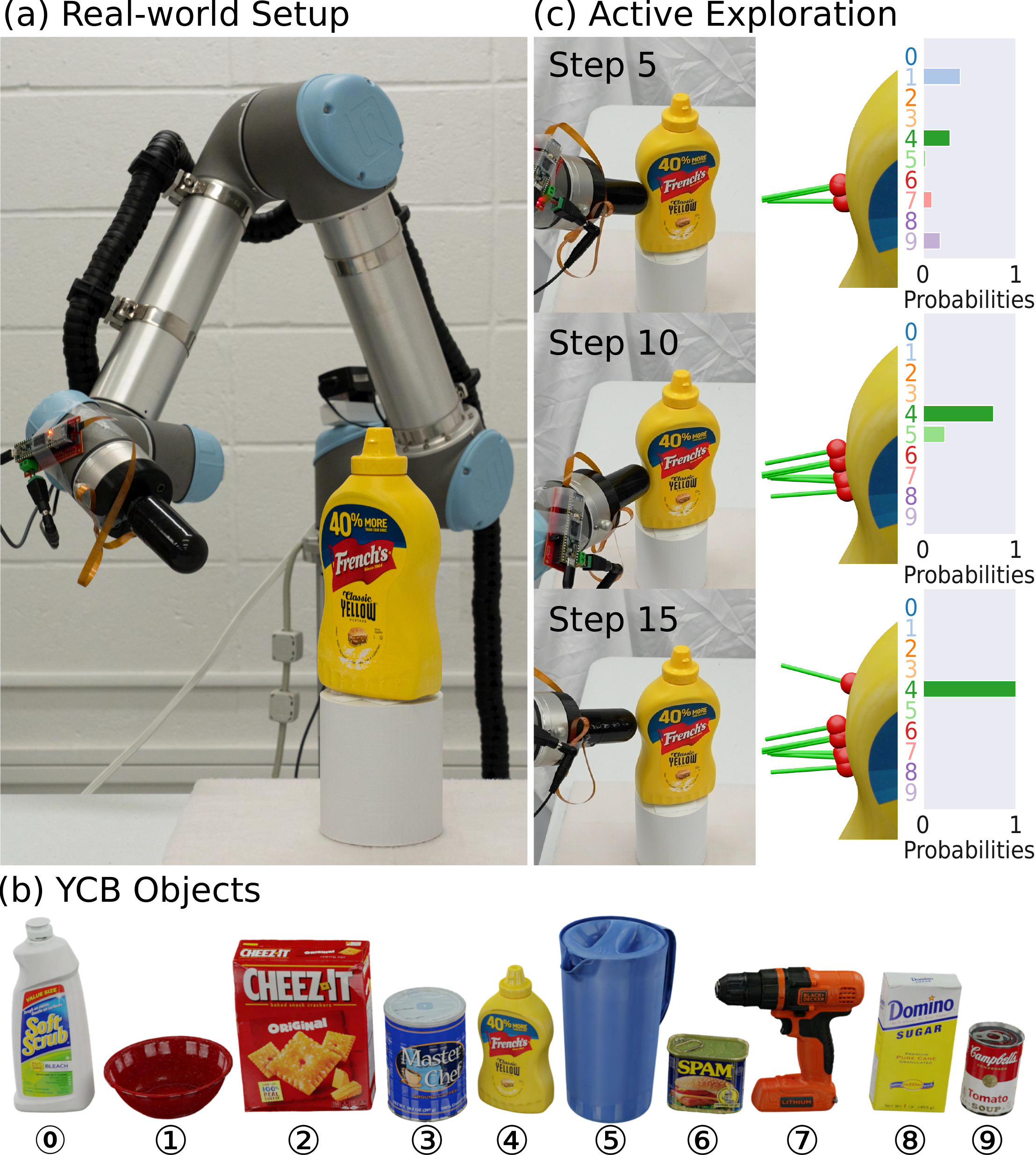}
    \caption{Tactile recognition of 3D objects. (a) Real-world setup. Our tactile finger is mounted on a robot arm. The object is placed on a flat surface with its Z-axis facing up, but the exact position on the horizontal plane and the orientation around the vertical axis are all unknown. (b) A known set of 10 YCB objects, selected to cover a variety of concave and convex 3D shapes and sizes. (c) Using our method, the robot actively collects data and quickly recognizes this object (mustard bottle). It first moves down the vertical edge and makes an initial hypothesis between objects 4 (mustard bottle) and 5 (pitcher base). It then moves up to obtain a final contact on a distinguishable geometry before making the correct identification.}
    \label{fig:teaser}
\end{figure}

In this work, we focus on learning an active exploration policy for the tactile recognition of 3D objects. We show that a co-training approach can be used for 3D object recognition, where it outperforms state-of-the-art alternatives. To achieve this performance level, we rely on the following advances: (1) We design our policies to make use of a richer action space by enabling 6 degrees of freedom (DOF) movements of the tactile sensor. With 6DOF movements, critical discriminative points can be obtained through small-angle adjustments, especially for our multi-curved tactile finger with all-around sensing coverage. (2) We use a richer representation of contact, going beyond binary cell occupancy. In particular, we use contact locations and surface normals, which provide important information about 3D shapes while still being simple enough to simulate, which enables a large amount of training in simulation and zero-shot transfer onto real robots. (3) We propose a more advanced, learning-based encoding of the tactile sensor data. We store tactile points into an unordered point set and encode them with PointNet++~\cite{qi2017pointnet++}. We show that this method provides effective representation for discrimination and exploration, and allows us to scale from binary 2D grids to 3D shape representation.
 
To the best of our knowledge, we are the first to co-train a 6DOF exploration and decision-making strategy for 3D object tactile recognition, a method that we dub TANDEM3D. We show that TANDEM3D outperforms a variety of baselines for recognizing a set of 3D objects placed under partially unknown positions and orientations. Our baselines include a learning-based all-in-one policy that does not distinguish between discriminator and explorer, heuristic-based exploration policies (such as info-gain, contour following, etc.), and an ICP discrimination policy. Compared to these methods, TANDEM3D recognizes objects with a higher success rate and is also more robust to different types and amounts of sensor noise.  We also validate our method on real robot experiments using a subset of the YCB objects~\cite{calli2017yale}.

\begin{figure*}[t!]
\centering
\includegraphics[width=\textwidth]{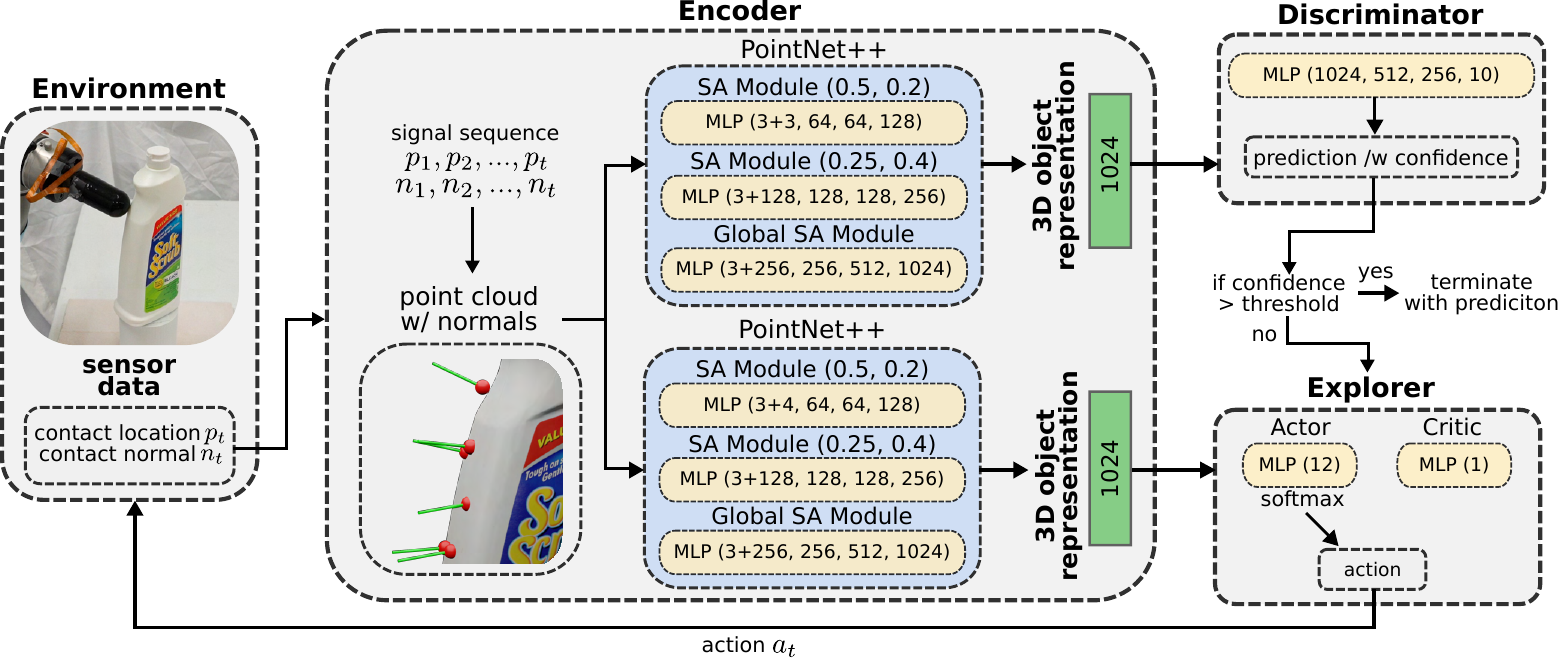}
\caption{An overview of our co-training architecture. Our proposed encoder encodes a sequence of tactile signals into 3D object representation using PointNet++. The parameters of the set abstraction (SA) module are the sampling ratio of the sampling layer and the grouping radius of the grouping layer. The multilayer perceptron (MLP) networks are highlighted in yellow rectangles and their parameters are the number of nodes in each fully-connected layer.}
\label{fig:overview}
\end{figure*}

\section{Method}

Our method builds on the co-training framework that we previously introduced and validated for 2D environments~\cite{xu2022tandem}. Referred to as TANDEM, it jointly learns exploration and decision making: an exploration policy (i.e., \textit{explorer}) determines the next action to collect more information, while a decision-making component (i.e., \textit{discriminator}) outputs a prediction for object identity along with a confidence value. If the confidence is high enough, it terminates the exploration. A third component, the \textit{encoder}, encodes the sparse and local tactile signals into a global object representation that is used by the explorer and discriminator. 

While TANDEM is effective for recognizing 2D polygons, it does not scale to 3D problems. Reducing tactile data to occupancy grids discards valuable information that could be used to discriminate between complex surfaces. Furthermore, the grid representation increases exponentially in size with the number of dimensions, which prevents learning. Finally, the limited action set is unable to quickly collect discriminative data on 3D geometry. We address these aspects below.

\subsection{6DOF Action Space}
\label{action}
In order to efficiently explore the complex geometry of 3D objects, we allow the finger to move in 6DOF action space $\mathcal{A}$. Using the top of the fingertip hemisphere as the reference point, we discretize the 6DOF action into small translation (x, y, z) and rotation (roll, pitch, yaw) steps, all with respect to the workspace frame. The robot picks one of x, y, z, roll, pitch, yaw, and can either increase or decrease for a small step (1cm for translation and 15 degrees for rotation). Thus, there are $2 \times 6 = 12$ actions in total. Within these small movements, the tactile finger constantly checks collision and produces contact locations and normals when  collision.

\subsection{3D Object Representation}
TANDEM uses binary collision signals and consequently rearranges them into an occupancy grid. They encode occupancy grids with CNNs. A straightforward extension into 3D space is voxel grids and 3D convolution. However, we use point clouds to store contact information for two reasons. (1) Contact locations and normals provide richer information for a single contact about the 3D object and yet are still simple enough to simulate. (2) The object is allowed to be anywhere in the workspace as long as the workspace center is occupied, resulting in a large workspace. Voxel grids are voluminous and scale cubically with the size of the workspace, taking up an unnecessarily large amount of memory. 

Our point cloud contains contact positions with contact normals as extra features. Compared to regular input data formats such as 3D voxels, point cloud data are unordered and can contain a  variable length of points. We choose PointNet++ as our network architecture to encode point clouds due to its permutation invariance of points and ability to handle variable input length. PointNet++ is a hierarchical extension of PointNet~\cite{qi2017pointnet} and uses PointNet as its basic building block for local pattern learning. 

In our encoder, there is a separate PointNet++ for both the discriminator and the explorer. Weights are not shared because they need to encode different aspects of the contact information, which leads to better performance. The two PointNet++'s have similar architecture but different inputs, as shown in Fig.~\ref{fig:overview}. For the discriminator, each point has a dimension of 3 (contact locations) + 3 (contact normals). For the explorer, the point cloud includes one extra point representing the current finger pose and each point has an extra bit of binary information indicating whether the point is the current finger pose or a regular contact pose. Without the important information about the current finger pose, the finger frequently gets stuck at workspace corners and is not able to navigate back toward the object.

The PointNet++ in our method has 2 set abstraction (SA) modules and 1 global SA module. Each SA module contains a sampling layer, a grouping layer, and a PointNet layer. In the sampling layer, a ratio of points is chosen as the centroid using the farthest point sampling (FPS) algorithm. In the grouping layer, points are grouped using the centroids with respect to a radius. In the PointNet layer, the local region of each group is abstracted by its centroid and local feature that encodes the centroid’s
neighborhood. The global SA module only has one PointNet layer that consumes the final abstracted centroids and their features.

Our 3D object representation is a feature vector of dimension 1024 produced by PointNet++. In the discriminator, the feature is passed through another MLP to generate the probability distribution over the 10 objects. The object with the highest probability is our prediction and the probability becomes the confidence. In the explorer, the feature is passed to the actor MLP to generate the action distribution and also the critic MLP to generate the state value for PPO training.



\section{Experiments}

In this section, we describe our experimental setup in both simulation and the real world\footnote{For real-world experiment videos or more information, please visit our project website at \url{https://jxu.ai/tandem3d}}. Our method is trained entirely in simulation and can be evaluated in simulation or with real hardware. We compare our method with a comprehensive set of baselines and then validate its performance on the real robot with a tactile sensor. 

\subsection{Experimental Setup}

We pick 10 objects from the YCB dataset as shown in Fig.~\ref{fig:teaser}. These objects are chosen to cover both convex and concave shapes and a variety of sizes intentionally. In simulation, we decompose each object into a set of convex parts using V-HACD~\cite{mamou2016volumetric} for collision checking.

The object is placed on a horizontal surface with its Z-axis pointing up. However, the rotation of the object around the vertical axis, as well as its translation along the axes of the horizontal plane are all unknown and randomized inside a 30cm by 30cm workspace. The only constraint is that some part of the object must occupy the center of the workspace so that the robot can make initial contact. In order to compensate for the calibration error in the real-world experiments, we also add a 2cm translation variance on the height of the objects during training in simulation. 

Real-world tactile sensors inevitably apply some level of force to the object before the contact is detected; for sensors with low sensitivity or light objects, the sensing act itself can thus induce movement. Handling such object movement during exploration is beyond the scope of this work, and we thus assume a sensor that can detect contact before causing movement. While such behavior is easy to simulate, in order to also achieve it in real life, we attach the object to the surface using Velcro. However, as can be seen in our accompanying video, some level of object tilt is inevitable after finger poking. Nevertheless, our algorithm is robust enough to such changes in object orientation. 

In real-world experiments, we use the DISCO~\cite{piacenza2020sensorized} finger (Fig.~\ref{fig:teaser}) to provide contact locations and normals. This multi-curved tactile finger has sensing abilities covering the hemisphere top and cylinder. We only keep contact information with a force magnitude larger than 0.5N and discard the rest because weaker contacts tend to have larger noise. We mount this finger on a UR5 robot arm to achieve 6DOF finger movement. In simulation, we use PyBullet to simulate a floating finger with the same tactile sensing capabilities.

Each episode is terminated if the number of steps surpasses 2,000 or the discriminator has a confidence value greater than the preset threshold of 0.98. When the episode is terminated, the prediction from the discriminator is compared to the true identity of the object to determine success. 

\subsection{Sensor Noise}
Being able to handle erroneous tactile readings without compromising efficiency and accuracy is critical for tactile-based applications. 
There are two types of sensor noise observed on our tactile finger.

\begin{itemize}
    \item \textit{Contact Noise}. This is the noise when a fake contact is reported when there is no contact. When it occurs, the output contact location can be anywhere on the finger surface. We quantify this noise as the percentage of time steps where the finger reports a non-existing contact. For our real sensors, we have empirically observed this to be approximately 0.1\%.
    \item \textit{Localization Noise}. This is the error of the predicted contact location when a real contact happens. When contact occurs, both the contact locations and the contact normals deduced from them can be noisy. We quantify this as the distance in mm between the real contact location and the reported one. For our DISCO finger, the reported level for this noise is between 1 and 2 mm on average.
\end{itemize}

In our simulation experiments, we test all methods described below under two noise conditions. The first one is intended to emulate our real sensor: 0.1\% contact noise and 2 mm localization noise. The second condition attempts to emulate a better sensor, assuming future advances in sensing technology: 0.025\% contact noise and 0.5 mm localization noise. All real-world experiments are obviously subject to the noise level of the real tactile finger.

An important advantage of learning-based methods is that noise can be introduced during the training process, enabling the method to adapt. Thus, for both noise conditions described above, all the learning-based methods discussed below are trained under the respective noise conditions.

\subsection{Baselines}
\label{sec:methods}

We compare our approach to a variety of learning-based and heuristic-based baselines explained below. The metrics that we are most interested in are the number of actions and the success rate in accurately identifying the objects.

\begin{itemize}
    \item \textit{Not-go-back}. This exploration policy picks a random move that leads to an unexplored finger pose at each step. A discriminator is trained with this exploration policy.
    \item \textit{Info-gain}. This method uses the info-gain heuristic. It chooses an action that provides the most salient information and leads to an unexplored finger pose. At the current time step $t$, let $\mathbf{p}$ denote the probability distribution over 10 objects produced by the discriminator. Let $\mathbf{p_{c}}$ and $\mathbf{p_{n}}$ denote the new probability distributions after applying a particular action when a new contact happens or not respectively. Clearly, $\mathbf{p_n} = \mathbf{p}$ because the predicted distribution does not change without new contacts. When computing $\mathbf{p_c}$, we assume the new contact location is on the top of the fingertip hemisphere. The action $a_t$ is chosen by: 
    \begin{align*}
        a_t &= \argmax_{a \in \mathcal{A}} \: \bigg\{ \mathcal{H} (\mathbf{p}) - \left(\frac{1}{2} \mathcal{H} (\mathbf{p_c}) + \frac{1}{2} \mathcal{H} (\mathbf{p_n})\right) \bigg\} \\
        &= \argmax_{a \in \mathcal{A}} \: \left\{ \mathcal{H} (\mathbf{p}) - \mathcal{H} (\mathbf{p_c}) \right\}
    \end{align*} 
    where $\mathcal{H}$ denotes the entropy of a probability distribution. It uses entropy as a measure of uncertainty and picks the action that reduces the most uncertainty. A discriminator is trained along with the explorer. 
    \item \textit{Edge-follower}. This method uses the contour-following heuristic. However, contour-following on a 3D object is not well-defined. At any point on a 3D surface, there is an infinite number of candidate edges to follow. In this implementation, the finger starts following a horizontal edge parallel to the workspace plane at the initial contact height. The finger angle is adjusted to avoid collision between the finger bottom and the object. We also implement a variant that follows the vertical edge parallel to the workspace XZ-plane but it performs much worse than our horizontal version. The reason is that depending on the initial position of the object and the initial contact location, the intersection of the object with the vertical edge plane can vary, potentially leading to a very small contour. Such a contour contains little information and can result in recognition failure. \textit{Edge-follower} is the only baseline that is not trained with contact noise. The performance drops significantly when applying contact noise during training because \textit{Edge-follower} can get trapped at fake contact locations and starts circling that location. In such a case, it can not continue exploring and the discriminator becomes unnecessarily cautious but the exploration policy is not able to increase its confidence. 
    \item \textit{Edge-ICP}. This method uses the same edge-following exploration heuristic but instead of training a discriminator, it uses the Iterative Closest Point (ICP) algorithm for recognition. We randomly sample 1,000 points on object mesh surfaces that are approximately evenly spaced as the ground truth point clouds. The discriminator uses ICP to match the point cloud with 36 different initial orientations linearly spaced between [0\degree, 360\degree]. For each object, the minimum error among all orientations represents the overall matching error. If the error is smaller than a threshold, it is marked as a match. Our threshold decays linearly from $5e^{-3}$ with $5e^{-5}$ step size. The output probability distribution assigns equal probabilities to the matched objects and zeroes to not-matched ones. This method requires no training. 
    \item \textit{PPO-ICP}. This method trains a PPO explorer as in ours with the ICP discriminator as in \textit{Edge-ICP}.
    \item \textit{All-in-one} This method does not separate explorer and discriminator. It has the same structure as the PPO explorer in our method but the action space is expanded to 24 actions. The first 12 actions indicate a move and the remaining 10 actions indicate a prediction. If the explorer picks a prediction instead of a move, the episode is terminated. A reward of 1 is given when an accurate prediction is made. 
    \item \textit{TANDEM3D}. This is our proposed method. Training takes 120 hours for the high-noise condition on an Nvidia RTX 3090 GPU and an Intel i9-12900K CPU. 
\end{itemize}

\section{Results and Discussion}

\begin{table*}
    \footnotesize
    \setlength\tabcolsep{2.5pt}
    \label{tab:results}
    \centering
    \begin{tabular}{c|ccc|ccc}
    \toprule
    \multirow{2}{*}{\textbf{Method}} & \multicolumn{3}{c|}{\textit{0.025\% Contact Noise, 0.5mm Localization Noise}} & \multicolumn{3}{c}{\textit{0.1\% Contact Noise, 2mm Localization Noise}} \\
    & \#\textbf{Actions} & \textbf{\#Points} & \textbf{Success Rate} & \#\textbf{Actions} & \textbf{\#Points} & \textbf{Success Rate} \\
    \midrule
    Not-go-back &$360.6\pm315.9$ & $4.666\pm2.847$ & $0.68$ & $294.2\pm289.9$ & $4.677\pm2.579$ & $0.52$  \\
    Info-gain & $678.2\pm318.1$ & $2.486\pm1.725$ & $0.29$ & $536.7\pm336.2$ & $2.988\pm1.875$ & $0.23$ \\   
    Edge-follower & $57.80\pm133.4$ & $12.21\pm27.10$ & $0.89$ & $59.22\pm142.2$ & $10.43\pm20.56$ & $0.83$ \\
    Edge-ICP & $85.30\pm126.7$ & $20.92\pm16.75$ & $0.26$ & $93.77\pm162.2$ & $19.84\pm14.52$ & $0.26$ \\
    PPO-ICP & $117.4\pm141.8$ & $16.95\pm11.09$ & $0.13$  & $109.0\pm126.4$ & $16.29\pm9.550$ & $0.14$ \\
    All-in-one & $118.3\pm268.8$ & $6.613\pm2.277$ & $0.13$ & $94.65\pm162.9$ & $6.669\pm2.114$ & $0.12$  \\  
    TANDEM3D (ours) & $45.73\pm96.88$ & $9.416\pm9.831$ & $0.98$ & $43.02\pm71.08$ & $9.806\pm10.61$ & $0.93$ \\
    \bottomrule
    \end{tabular}
    \caption{Comparative performance of various methods in simulation. Methods are trained and evaluated under two noise conditions. For each method, we present the number of actions taken (\#Actions) and the number of points explored  (\#Points) at termination, as well as the success rate in identifying the correct object (Success Rate). Mean and standard deviation over 1,000 trials are shown. A detailed description of each method can be found in Sec.~\ref{sec:methods}.}
\end{table*}


\begin{figure}
\centering
\includegraphics[width=0.65\textwidth]{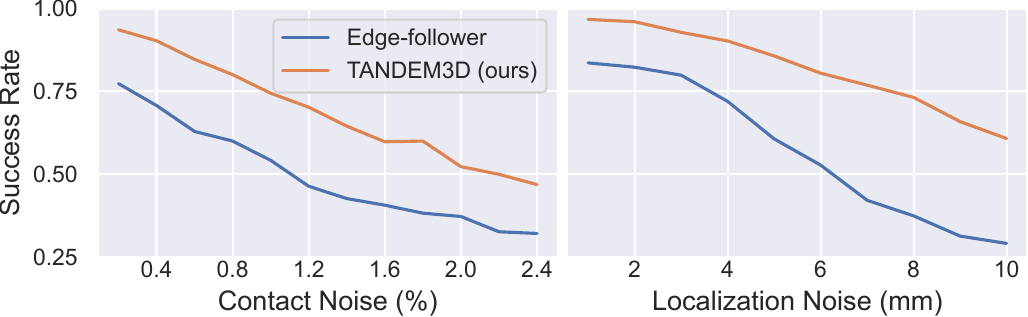}
\caption{Robustness and generalization to larger sensor noise. Despite being trained on lower noise levels, \textit{TANDEM3D} retains a high success rate as the contact noise increases to 2.4\% or the localization noise increases to 10mm, while \textit{Edge-follower}'s performance worsens more significantly.}
\label{fig:noise}
\end{figure}

\begin{figure}
\centering
\includegraphics[width=\textwidth]{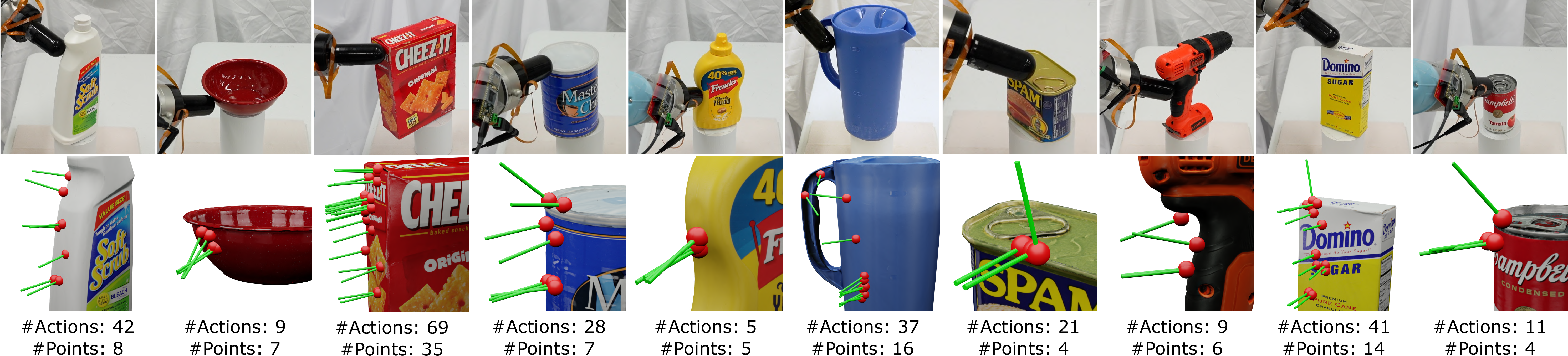}
\caption{10 successful examples of our method in real robot experiments. The top row shows the object poses and the final finger poses. The medium row shows the collected contact positions and normals at termination. The last row shows the results for each trial. Our method reacts to previous observations along the trajectory and gradually moves to the most distinguishable areas for efficient recognition. Each example is accompanied by a real-world video on our project website at \url{https://jxu.ai/tandem3d}.}
\label{fig:real}
\end{figure} 

We evaluate all methods with 1,000 simulated trials and the results are shown in Table~\ref{tab:results}. 

\textit{TANDEM3D} outperforms all other baselines. It learns an exploration behavior that combines following approximate vertical edges (either upwards or downwards) and swinging finger by adjusting angles, as illustrated in the real-world examples in Fig.~\ref{fig:real}. The angle adjustment enables the robot to collect contacts diverging from the vertical trajectory quickly which can provide critical information about object geometry. The swing motion also suits very well with our multi-curved tactile finger which has large sensing coverage. 

\textit{Edge-follower} has the closest performance to ours. However, without the flexible swing motions, it has to follow a large portion of the edge in order to gather the same amount of information. In addition, \textit{Edge-follower} is much more sensitive to both contact noise and localization noise. With contact noise, the finger can get stuck in following a fake contact. With larger localization noise, the points diverge from the contour trajectory which \textit{Edge-follower} overfits to, leading to a poor success rate.

To further compare the robustness and generalization of \textit{TANDEM3D} and \textit{Edge-follower} to larger sensor noise than what they are trained on, we evaluate both methods with contact noise up to 2.4\% and localization noise up to 10mm. Note that, in this condition, noise levels are only increased for testing, and not for training. As shown in Fig.~\ref{fig:noise}, the success rate of \textit{Edge-follower} deteriorates more dramatically as sensor noise increases beyond the levels seen in training.

Despite being proven to be a useful exploration heuristic in many previous works, \textit{Info-gain} performs surprisingly badly. We attribute its unsatisfying performance to the fact that we are training a discriminator along with it from scratch. Unlike other explorers, the \textit{Info-gain} explorer relies on the output from the discriminator at each time step. It needs a good discriminator to start with. We also notice that the finger tends to leave the object instead of approaching it during exploration. We think that the \textit{Info-gain} explorer attempts to make a contact that is far from the center as those points tend to provide the most salient information for discrimination. 

The methods with classic ICP recognition also demonstrate low accuracy due to the limited number of points in the partial point clouds. Taking \textit{Edge-ICP} as an example, even with a complete horizontal edge-following trajectory, the point clouds gathered for each object are mostly circles and ellipses with a few points. It is difficult for the ICP algorithm to match such simple shapes to a full 3D point cloud with 1,000 points. The \textit{All-in-one} baseline does not have a dedicated discriminator that can be co-trained with batches of labeled samples collected by the explorer, leading to inefficient training and low performance.

\subsection{Real-world Performance}

We test the performance of TANDEM3D in real-world experiments. For each object, we run 2 trials with the object randomly placed on the support in its upright orientation. The results are shown in Table~\ref{tab:real}. Our method still achieves a high success rate (although slightly lower than in simulation), thanks to its ability to generalize in the presence of noise. The sim-to-real gap when transferring our trained models to the real robot is largely due to workspace calibration, change of object pose under contact, mesh discrepancy caused by convex decomposition, etc. For example, real object 2 (cracker box) has distorted faces which introduce a large mismatch.  We select one successful example per object from our real-world experiments and visualize the object poses, final finger poses, collected contact positions, and contact normals before a final prediction is made in Fig.~\ref{fig:real}.

\begin{table}[h!]
    \footnotesize
    \label{tab:real}
    \centering
    \begin{tabular}{c|ccc}
    \toprule
    \textbf{Method} & \#\textbf{Actions} & \textbf{\#Points} &  \textbf{Success Rate} \\
    \midrule
    TANDEM3D & 38.21 $\pm$ 25.16 & 10.31 $\pm$ 10.93 & 0.85 (17/20) \\  
    \bottomrule
    \end{tabular}
    \caption{Real robot experiment results (mean and standard deviation over 20 trials).}
\end{table}

As we can see, \textit{TANDEM3D} reacts to previous observations and takes moves to the most discriminative areas. For example, on object 0 (bleach cleanser), the finger moves to the back of the object and contacts the recess, which is a distinguishable geometry special to object 0. On object 5 (pitcher base), the finger moves up towards the pitcher handle and then makes a decision by making a small angle adjustment to contact the handle using the side of the finger. On object 7 (power drill), the finger swings up to contact from beneath the drill and immediately makes the correct prediction. Please see our accompanying video for a better demonstration of these behaviors.

\section{Chapter Summary}

The data sparsity problem is even more pronounced in 3D environments, where objects present more complicated geometry, and we need a better encoder to compute 3D representations from a sequence of sparse tactile data. Built on top of the co-training framework, TANDEM3D further addresses the data sparsity challenge by enabling 6DOF movements of the tactile sensor and is able to discover discriminative points through small-angle adjustments, taking advantage of a tactile finger with all-around sensing coverage. It is based on an encoder that is co-learned with the discriminator and explorer, and builds the 3D object representation from the contact positions and surface normals acquired via tactile sensing. Compared to state-of-the-art alternatives, TANDEM3D can correctly identify 3D objects with fewer actions and a higher success rate.  It also demonstrates better generalization ability to different types and amounts of sensor noise.

%% file: chapters/geotact.tex
\chapter{Tactile Object Retrieval from Granular Media}
\label{chap:geotact}

\def\OURS{GEOTACT\xspace}
\hyphenation{GEOTACT}

In this chapter, we discuss GEOTACT, the first robotic system that can retrieve unseen objects buried completely under granular media using only tactile sensing. 

\section{Motivation}

While TANDEM~\cite{xu2022tandem} and TANDEM3D~\cite{xu2023tandem3d} can recognize objects with very few actions, a step towards human dexterity with tactile sensors, it operates in an open-air and tabletop environment. In GEOTACT, we want to push the limit further by moving into granular media, a highly sensor-deprived environment where vision is completely occluded. Interacting with granular media, for example, in order to retrieve buried objects, is a foundational skill for numerous important robotics applications. Examples include mining and demining, search and rescue, excavation and construction, archaeology, and exploration of terrestrial, seabed, and even extraterrestrial environments, all requiring robots that can extract valuable minerals, metals, or other geological materials from the ground. Humans are experts at performing such tasks, for example, using tactile sensing to compensate when visual feedback is absent; however, transferring such behaviors to robots poses multiple challenges. 

First, uncertainty and sensing noise greatly affect operation inside sensor-deprived environments such as granular media. With vision unavailable, touch becomes the prominent sensing modality. However, all tactile sensors exhibit some level of noise in their readings, which is typically exacerbated when submerged under granular media. Tactile sensing is fundamentally an active sensing modality, needing an intelligent exploration policy to guide movement and compensate for the sparse nature of the data. 


Second, manipulation within granular media places specific requirements on the hardware. In a granular environment, contacts come from every angle and direction, thus requiring extensive sensing coverage. Despite many breakthroughs~\cite{patel2021digger,lambeta2020digit,johnson2009retrographic,yuan2017gelsight,dong2017improved,chang2024investigation} in sensor developments, many sensors are either flat or do not have sensing coverage around edges and corners.  
In addition, we want an intruder shape that can render low resistance and smooth movement inside granular media. For example, circular and elliptical shapes typically experience slightly less resistance during horizontal dragging and significantly less resistance during vertical lifting compared to rectangular shapes~\cite{tripura2022role}.
Due to such compounding limitations, robotic manipulation in granular media remains poorly explored compared to tactile-based work with objects in open-air environments.

\begin{figure}[!t]
\centering
\includegraphics[width=0.65\textwidth]{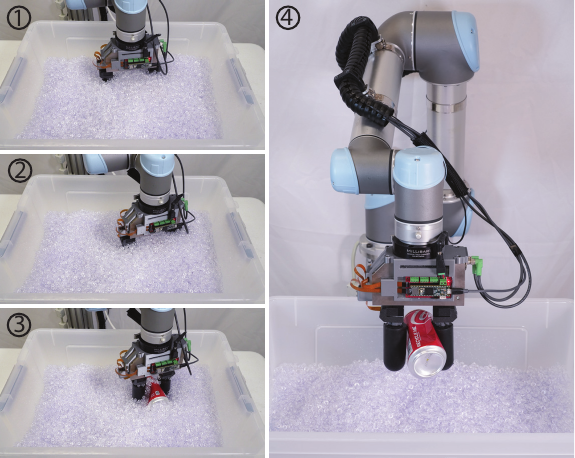}
\caption{Retrieving buried object from granular media. In this task, an object is buried under granular beads. We mount a parallel gripper with two tactile fingers on a robot arm to retrieve the buried object using only tactile feedback. Our learned policy uses a series of pushing actions to funnel the buried object into a stable grasp and then successfully retrieve the target object.}
\label{fig:teaser}
\end{figure}

In this work, we propose \textbf{\OURS} (\textbf{G}ranular \textbf{E}nvironment \textbf{O}bject \textbf{T}racing and \textbf{AC}quisition via \textbf{T}ouch), a novel method for retrieving objects buried under granular media. \OURS relies exclusively on tactile sensing, as we assume the object can be completely hidden from any vision sensors. In particular, we use the DISCO tactile finger~\cite{piacenza2020sensorized}, which provides both the streamlined shape and sensing coverage needed to enable the task. 

Still, sensor noise remains a significant challenge in addressing our proposed task. To address this, we take a multi-pronged approach. \OURS is learning-based, trained with reinforcement learning entirely in simulation, and then zero-shot transferred onto the real hardware. Our method is trained end-to-end, without separating perception and control policies, and is model-free, making no assumptions about the shapes or dynamics of objects. Compared to heuristics-based methods, which are more sensitive to sensor noise, we find that learning-based methods can incorporate noise into the training process and thus become more robust to the considerable amount of noise in granular media.

Critically, \OURS relies on a pushing behavior which further reduces uncertainty and increases robustness against noisy tactile readings. Despite pushing having been shown in the literature to be a key element of many manipulator operations~\cite{brost1988automatic,mason1986mechanics,dogar2010push,dogar2012planning}, no previous work has utilized pushing action to grasp objects within granular media. The \OURS formulation of the learning problem action space leads to the emergence of a series of pushing behaviors that funnels the target object into a stable grasp. 

To enable this learning process, we propose a curriculum strategy that enables learning in simulation, followed by zero-shot transfer to real hardware. We first pre-train our policy in an open-air environment and then fine-tune it inside simulated granular media. Despite the recent effort from the community to improve granular material simulation accuracy and efficiency~\cite{kloss2012models,millard2023granular,thompson2022lammps,xian2023fluidlab,fang2021chrono,servin2021multiscale,ansys,haeri2020efficient,hu2019difftaichi}, granular media simulation are still significantly slower than open-air simulation. However, our curriculum strategy helps to bootstrap this training process, allowing us to reach a better policy within a shorter training time by transferring the skills learned from the open air to granular media. 

In summary, GEOTACT's contributions are as follows:
\begin{itemize}
    \item 
    To the best of our knowledge, we are the first to demonstrate robotic grasping of objects buried under a specific type of granular media using only touch sensing.
    \item In order to handle the tactile noise inherent to this problem, we use a learning-based method trained end-to-end in the presence of noise, and introduce a curriculum that enables training in simulation and zero-shot transfer to real hardware for this task.
    \item We show that our formulation leads to the emergence of pushing behaviors that reduce uncertainty and enable robust performance in real scenarios.
\end{itemize}

Our quantitative real-robot experiments show that \OURS achieves success rates of 54\% and 63\%  on its training set (seven objects) and testing set (six additional objects not seen during training), respectively.
In additional testing, we also show promising early results on an extended set of 22 objects, including rigid, deformable, and articulated objects with various complex shapes. Our extensive experiments in both simulation and on the real robot show that our method is more robust to sensor noise and achieves a higher success rate compared to baselines.

\section{Method} 

We formulate our problem as follows. A robotic manipulator is presented with a workspace filled with granular media. 
An object is buried completely under the granular media, placed with unknown orientation or exact position, such that the object is in the way of the gripper and a contact is guaranteed if the gripper moves straight forward.
The goal of the manipulator is to retrieve the buried object, by lifting it clear of the granular environment. We assume that the object can start out fully buried with no part visible from the surface. Our goal is to perform the task without relying on vision and using only touch sensing.

Tactile sensing in granular media requires the ability to distinguish between readings due to ubiquitous contacts with the granules themselves, and readings indicating contact with the target object. As explained later in this section, we use a simple filtering technique to make this distinction. However, the nature of operation in such environments increases the noise inherent in any tactile sensor. We thus need an exploration policy that is well-suited for handling tactile noise and spurious or erroneous readings. 

Rather than attempting to handle sensor noise via heuristics or further signal processing, we achieve a high level of robustness by formulating the object retrieval task as a learning problem, trained in the presence of noise. We describe this formulation next, and show how it leads to the natural emergence of a pushing behavior that has the role of reducing uncertainty before a grasp is attempted. Finally, we will discuss our training curriculum, which enables training of this policy in simulation followed by zero-shot transfer to hardware.

\subsection{Action Space Definition and Pushing Behavior} 
\label{sec:action}

We formulate our problem as a model-free reinforcement learning task. 
Object exploration and retrieval are achieved through a series of discrete translation/rotation actions of the gripper, where the translation action is two-dimensional on a 2D $xy$ plane, and the rotation action is one-dimensional along the $z$ axis.
At each step, our policy takes in a sequence of contact locations and forces from the past steps and then outputs the next action for the gripper to take for exploration. Our policy also determines when to stop pushing and execute the grasp. 


In defining the action space of our learning task, we build on the extensive body of work (reviewed earlier) showing that a planar-pushing primitive can be used to reduce uncertainty and funnel an object into a state that lends itself to robust grasping. However, a heuristic-based pushing primitive is difficult to implement in very noisy conditions. Instead, we aim to formulate the learning problem such that the agent naturally learns to make use of pushing in the context of end-to-end task learning.

Specifically, our action primitives consist of a planar pushing action and a grasping action. The gripper moves in a 2D plane and is always perpendicular to the table, as shown in Fig.~\ref{fig:teaser}. 
Given the current gripper pose (position $[x, y]$ and rotation $\theta$), the action is a 4-dimensional vector $[dx, dy, d\theta, G]$.
A value of $G=0$ indicates a pushing action, in which case $dx, dy, d\theta$ indicate the change to $x$, $y$, and $\theta$, respectively. For $x, y$, the change is 1cm and for $\theta$, the change is $15^\circ$; namely, $dx, dy \in \{-1\text{cm}, 0, +1\text{cm}\}$ and $d\theta \in \{-15^\circ, 0, +15^\circ\}$. 
Notice that in this action space formulation, it is possible to change $x, y$, and $\theta$ simultaneously in one action, by setting the corresponding elements at the same time.
$G=1$ corresponds to a grasping action. In this case, the values of $dx, dy, d\theta$ are ignored, the gripper is closed, and the object is lifted. The episode is terminated when a grasping action is executed or a predefined maximum number of steps is reached. 

As discussed in the results section, this formulation results in the learned policy effectively making use of sequences of pushing behaviors of various lengths. However, a critical aspect of learning such behaviors is training in the presence of noise, which we discuss later in Sec.~\ref{sec:noise}.

\subsection{Observation Space and Tactile Data Processing} 
\label{sec:noise_gap}

\begin{figure}[t]
\centering
\includegraphics[width=0.65\textwidth]{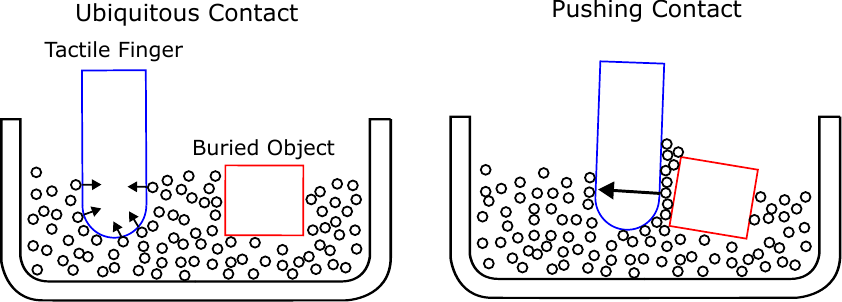}
\caption{Two types of contact inside granular media: ubiquitous contact and pushing contact. Ubiquitous contacts are between the finger and the granular media, but pushing contacts are sensed only when the finger has started pushing the buried object.}
\label{fig:pushing_contact}
\end{figure}

\begin{figure*}[t]
\centering
\includegraphics[width=\textwidth]{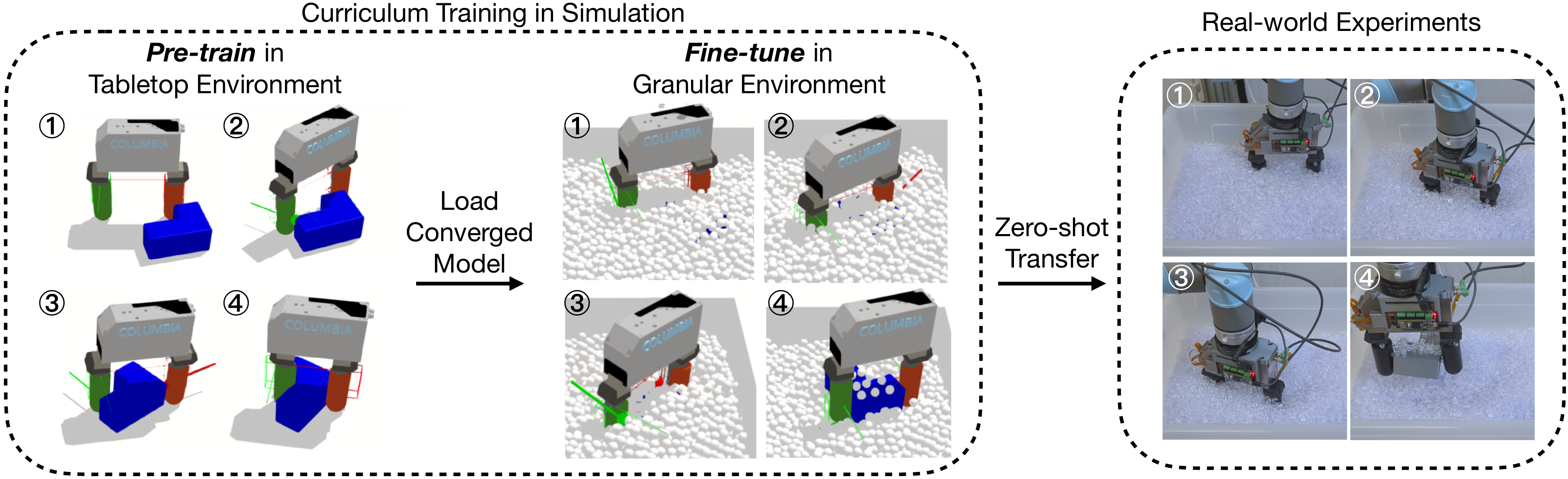}
\caption{Curriculum training strategy.
The numbering of the frames corresponds to their order within a single episode.
We first pre-train our policy on a tabletop environment and then fine-tune it inside granular media. We then zero-shot deploy our policy on the real hardware for evaluation. This curriculum allows the policy to converge to a higher performance within a shorter training time.}
\label{fig:curriculum}
\end{figure*}

At each time step $t$, the observation of our policy is a sequence of contact locations $p_t, p_{t-1},$ $\dots,$ $p_{t-h}$, net forces $f_t, f_{t-1}, \dots, f_{t-h}$, and previous actions $a_{t-1}, a_{t-2}, \dots, a_{t-h-1}$ from the last $h = 10$ steps. Observations from both fingers are concatenated to form the final observation of the policy. Both the contact location $p_t$ and the net force $f_t$ are 3-dimensional vectors under the finger's frame of reference (using the center of the finger base as the origin). The previous action $a_{t-1}$ is a 4-dimensional vector, as discussed in Sec.~\ref{sec:action}.
When there is no contact recorded after a particular step, we simply fill in zeros in the location and force vectors.
Compared to high-dimensional or multimodal tactile data, contact locations and forces are easy to simulate, enabling a large amount of training in simulation and zero-shot transfer to real-world experiments.

Unlike open-air environments, where the contacts are simply between the tactile finger and the target object, inside the granular media there are two types of contacts our policy needs to distinguish between: ubiquitous contact and pushing contact (Fig.~\ref{fig:pushing_contact}). Ubiquitous contact is the contact with the granular media, and the contact forces are small whether the finger is moving inside the granular media or not. Pushing contact is the contact sensed by the finger when the object has started moving due to the pushing, and because of the larger dimensions of the target object compared with granular media, the contact forces are much larger. We assume only pushing contacts carry important information for our policy, and we want to discard ubiquitous contacts with granular media. We thus use a force filter to keep only the contact positions with a force magnitude larger than a threshold. 
The net force vector is the combined force applied to our finger, which also contains the direction of the force.

Tactile observation in granular media is much more noisy compared to open air. First, there might not be direct contact between the finger and the target object when pushing contacts are detected. Instead, there could be a layer of granules between the finger and the target object. This layer can potentially introduce more noise to the force directions. Second, for small and light objects, pushing contact can be hard to distinguish from ubiquitous contacts, and, depending on how fast the finger accelerates and moves inside the medium, pushing contacts can be falsely reported even though the finger has not started pushing the target object. 

Ideally, object- and granular media-specific force filtering thresholds might be needed for better filtering accuracy. 
However, in this work, the filtering threshold only differs between simulation and real hardware but remains the same for all objects, and we observe that our policy learns to be robust to spurious pushing contacts.
The filtering threshold is found experimentally through calibration. During the calibration phase, we drag the submerged finger through the granular media without any buried objects. The dragging follows a predefined trajectory with multiple changes in direction. We record the maximum force sensed during the whole trajectory. We then find the minimum integer value larger than the maximum force to be the chosen force threshold. We do the calibration process separately in simulation and on the real hardware, giving us different filtering thresholds.

\subsection{Reward Design and Policy Architecture}

The reward to our policy is a hybrid of dense and sparse rewards. We constantly reward the policy when the center of the gripper gets closer to the center of the object, and we provide a large sparse reward when the policy decides to grasp and successfully lift the target object from the granular materials. At step $t$, let $d_t$ denote the Euclidean distance between the gripper center position $(x_g, y_g)$ and the object center position $(x_o, y_o)$, then the reward at each step $t$ is computed as 
\begin{gather*}
    r_t = r_t^{dense} + r_t^{sparse} \\
    r_t^{dense} = \alpha \times (\frac{1}{d_t + 0.1} - \frac{1}{d_{t-1} + 0.1}) \\
    r_t^{sparse} = \begin{cases}
        \beta, \text{if successfully retrieves the object} \\
        0, \text{otherwise}
    \end{cases}
\end{gather*}
where $\alpha$ and $\beta$ are hyperparameters used to balance the dense reward versus the sparse reward. In our experiments, we set $\alpha = 20$ and $\beta = 800$.

Our policy has one actor network and one critic network, trained with PPO~\cite{schulman2017proximal}. 
Both networks are a 5-layer multilayer perceptron (MLP), and each hidden layer has 256 neurons.
Since our task is not image-based, the actor and critic networks do not share any weights or layers, which is a common practice and shown to have better performance for control tasks~\cite{andrychowicz2020matters}.
We use \texttt{arctan} as our activation function between layers. Notice that we also tried a Transformer~\cite{vaswani2017attention} and an LSTM~\cite{hochreiter1997long} encoder to learn the relationship among observations from different steps in our history buffer; however, we found that the fully connected MLP learns faster and converges to a higher reward.

\subsection{Curriculum Training Strategy}

We train our policy completely in a simulated environment, then attempt zero-shot transfer to real hardware. However, this approach raises the important issue of simulating granular media, which, as noted earlier, is still an active research problem. Current methods for simulating granular media are still slow compared to open-air simulation, sometimes even slower than real-time. However, training inside simulated granular media is still necessary for us due to the gap between the open-air environment and the granular media environment. Apart from extra sensor noise, the object motion differs significantly between the two environments. First, the object can be perturbed by forces propagated through granular media even though the gripper is still some distance away from the object. Second, the object can move in 6DOF in granular media rather than in a 2D plane when sitting on a surface in open air. As a result, policy trained purely in the open-air environment will transfer badly to a granular media environment. 

To not only bridge the gap from open-air to granular media but also train fast, our method uses a curriculum training strategy, as shown in Fig.~\ref{fig:curriculum}. We first train a policy on an open-air tabletop environment and then fine-tune the tabletop policy inside a granular media environment. The tabletop task follows the same RL formulation except that the object sits on a tabletop in the open air. With this curriculum, we are able to bootstrap the training inside granular media by transferring skills learned from the tabletop environments. This curriculum helps our method to converge to a better performance within a shorter training time. 

In order to increase the simulation speed in granular media further, we sacrifice some simulation accuracy for speed. Compared to the pony beads that we use in the real world, we use spheres to approximate their shape. We also make the particles larger in size and then simulate a smaller number of them compared to the real world. Despite the mismatch in the simulated particles, fine-tuning still helps to adapt the tabletop policy to the granular media environments and, consequently, a better sim-to-real transfer. 

\section{Experimental Setup}

\subsection{Real-world Setup} 

\begin{figure}
\centering
\includegraphics[width=0.5\textwidth]{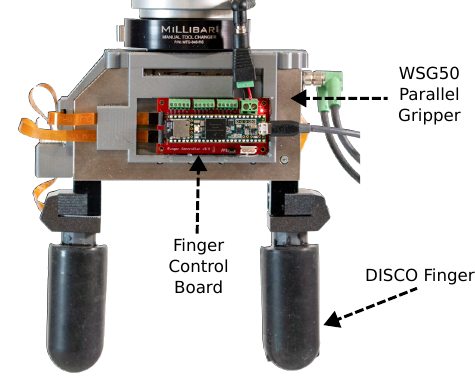}
\caption{Parallel gripper with multi-curved tactile fingers. Our gripper is effective at exploring under granular media due to the DISCO tactile fingers' streamlined shapes and all-around sensing coverage.}
\label{fig:hardware}
\end{figure}

We use a WSG50 parallel gripper with two DISCO~\cite{piacenza2020sensorized} tactile fingers, as shown in Fig.~\ref{fig:hardware}. We mount this gripper to a UR5 robotic arm to control its movement. This setup is effective for interacting with objects under granular materials, as the hemispherical top allows easy penetration into granular media and the cylindrical shape allows smooth movement with small drag. This multi-curved tactile finger has sensing abilities covering the hemisphere top and cylinder. This all-around sensing coverage allows it to sense contact from many directions, which is essential for exploration under granular media. When fully opened, the distance between the two finger centers is 143mm.

\begin{figure*}[h]
\centering
\includegraphics[width=\textwidth]{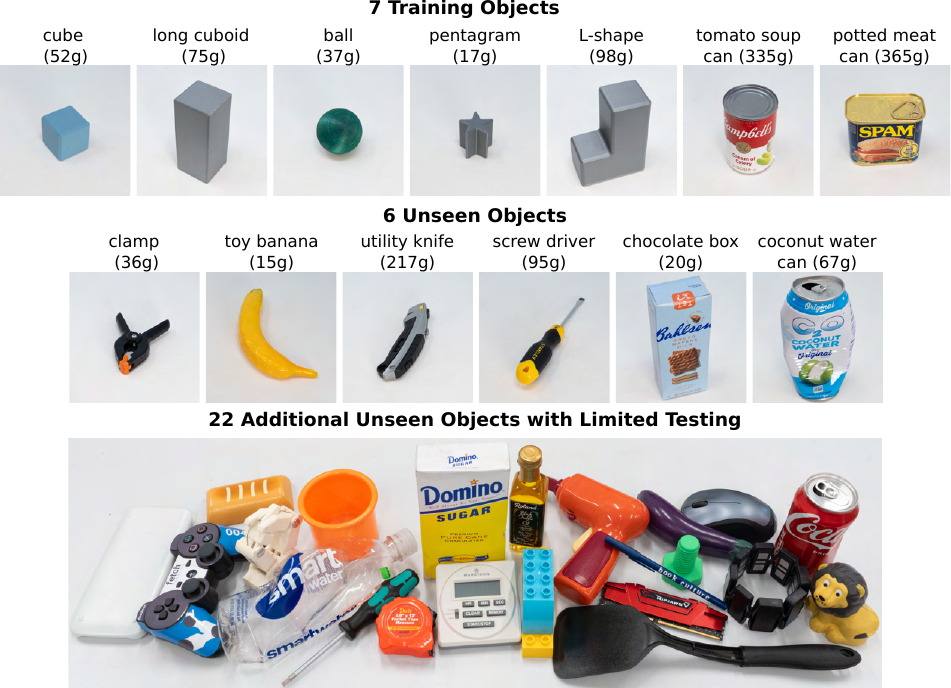}
\caption{35 objects used in our real-robot experiments. Our method is trained on a set of 7 objects. It is then evaluated quantitatively (10 attempts on each object) on both the set of 7 training objects and a set of 6 previously unseen objects. In additional testing, we also show that our method can work on an extended set of 22 unseen objects with at most three attempts per object. The images of training and unseen objects do not correctly reflect their relative sizes. See the demonstration videos for each object on our website for accurate size information.}
\label{fig:objects}
\end{figure*}

The DISCO finger returns a single contact containing the contact location vector and force vector. The tactile finger uses a pre-trained machine learning model to predict a single contact from the high-dimensional raw signals. When there are multiple contacts, the output contact location roughly matches the location of the contact with the largest force, and the output force roughly matches the combined net force from all contact forces.

To filter out ubiquitous contacts with the granular media and keep only the pushing contacts with the target object, we discard those contacts with a force magnitude smaller than 4N. We fill in zeros into the location and force vectors to our policy when there are no pushing contacts detected at a particular step. We find the 4N filtering threshold empirically through calibration, and we use the same 4N threshold for all objects. The optimal filtering threshold should vary across different objects and different types of granular media. But our policy learns to be robust to spurious pushing contacts so we do not need object and granular media-specific calibration. The granular media in our experiments are pony beads with a diameter of 9mm. Our gripper with DISCO fingers is able to move smoothly within such media without jamming. 

In our real-world experiments, we evaluate our method on 35 objects in total, as shown in Fig.~\ref{fig:objects}. Our object set covers a variety of complex shapes, including rigid, deformable, and articulated objects. The object weight in our set also varies a lot, with the lightest object (pen) being 4.5g and the heaviest object (sugar box) being 470g. We evaluate quantitatively on 13 objects with 10 trials each and among them, 7 objects are used for training in simulation and 6 objects are unseen. We then qualitatively show that our method can work on another 22 unseen objects randomly selected from our lab. We quickly go through this set of objects, and we move on to the next object as soon as a successful trial is achieved on the current object. 

The gripper starts initially in the air, and it follows a predefined heuristic trajectory to move into the granular media, makes a shaking motion to shake off jammed beads, and then moves straight forward until the first pushing contact. We start applying force filtering after the finger starts moving forward inside the granular media to filter out ubiquitous contacts, and the learned policy only takes over after the first pushing contact is obtained.

We bury the object so that part of the object occupies the initial forward trajectory of the finger. This is to guarantee an initial first contact as the finger moves forward. Between each trial, we randomize the location of the object, alternate the location of the object in front of the left and right fingers, and we also rotate the object to cover a diverse set of orientations. The episode starts with the first pushing contact and then ends when the maximum steps of 100 are reached or the policy executes a grasp. 


\subsection{Simulation Setup} 
Different from the real-world experiments where we only evaluate inside granular media, we have two different setups for the curriculum training strategy in simulation: tabletop and granular media. We build the two environments using the IsaacGym physics simulator~\cite{makoviychuk2021isaac}. For both environments, we simulate a floating parallel gripper that moves in a 2D plane, according to its real hardware dimensions. 

Despite the fact that we attempt to bury the object such that its center is in front of one of the fingers, it is very hard to guarantee it in practice on the real setup. As a result, in order to be robust to the object location variance, we add more randomization on the object location and orientation during training in simulation. In each episode in the simulation, the object is loaded at a fixed distance away from the gripper. The exact distance does not matter since the finger always moves straight forward until the first contact. However, we vary the location of the object along the direction perpendicular to the gripper's approaching direction by $\pm2$cm. We also randomize the orientation of the object. It is possible that for small objects like pentagrams, the finger can pass the object without making contact under the location randomization, in which case we simply move on to the next episode. 

For simulating granular media, we use spheres with a diameter of 14mm as our particles, and we simulate 4000 particles in total. These spheres do not match the shape of the pony beads in real-world experiments and their sizes are much larger. However, simulating the exact shape and size makes the granular media simulation extremely slow and impractical for training. Despite such a mismatch, fine-tuning in simulated granular media still helps a lot in transferring the tabletop policy to the real robot. 

Unlike the real DISCO finger, which reports a single contact, in simulation we can obtain the contact between the finger and any other object, including the granular beads. We then use a force filter of 3N to discard the ubiquitous contacts. Since there can be multiple pushing contacts even after applying the force filter, we feed the contact with the largest force magnitude into our policy. 

\subsection{Sensor Noise}
\label{sec:noise}
Real-world tactile sensors all show some level of erroneous readings, especially inside granular media. One advantage of the learning-based method is that we can incorporate noise into our training process in simulation for more robust transfer onto the real robot. For a single contact on our DISCO finger, the reported level for the contact location noise is between 1mm and 2mm on average, and the reported level for the contact force noise is 0.1N on average for a force level of $3 \sim 4$N. As a result, due to increased noise level in granular media, we apply a larger location noise uniformly distributed from $[-1\text{cm}, 1\text{cm}]$ and a larger force noise uniformly distributed from $[-0.2\text{N}, 0.2\text{N}]$ to each element in the contact location and net force vectors during training in simulation for all learning-based baselines. 

\subsection{Experimental Conditions and Baselines}

We perform an extensive evaluation of \OURS on the object sets introduced earlier, in both simulation and on real hardware. Since the task we are tackling here has not been previously demonstrated in the literature, we do not have a direct state-of-art method to compare against. However, we compare against a number of alternative approaches and ablations aiming to quantify the importance of each component of our approach. We present all experimental conditions below:

\vspace{3mm}
\noindent \textbf{Align-to-grasp (A2G)}.
This method rotates the gripper to align with the direction of the contact force after the first pushing contact is detected. This is a heuristics-based method and does not require training, intended to evaluate the importance of a learning-based method that trains in the presence of noise. 

\vspace{3mm}
\noindent \textbf{A2G-push}.
This variant of A2G starts rotating and aligning only after five consecutive contacts on the same finger have been detected. This is intended to filter out spurious pushing contacts in granular media, and to evaluate the performance of pushing alone as a heuristic for reducing uncertainty.

\vspace{3mm}
\noindent \textbf{\OURS-tabletop}.
This is the policy from the first stage of our curriculum training pipeline, trained on the open-air tabletop environment. 

\vspace{3mm}
\noindent \textbf{\OURS-granular}.
This is the policy defined in our method, but trained exclusively in granular media, thus bypassing our proposed curriculum.

\vspace{3mm}
\noindent \textbf{\OURS}.
This is our full proposed method, combining all the aspects discussed in this study.

\section{Results and Discussion}

\begin{figure*}[h!]
\centering
\includegraphics[width=\textwidth]{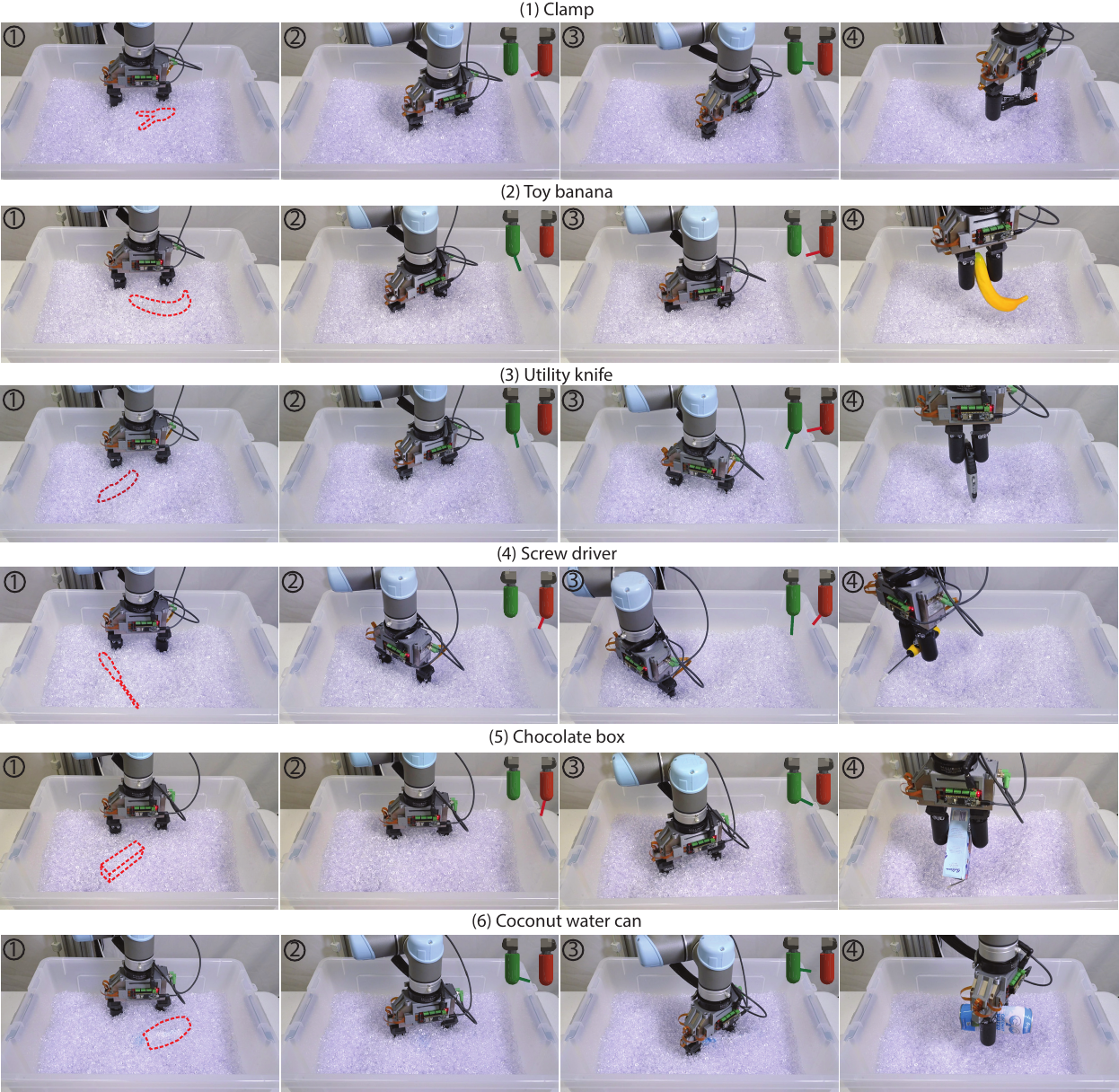}
\caption{\OURS successful demonstrations on the real robot. We show one example each for the six unseen objects evaluated quantitatively. The first image in each example outlines the initial position of the buried object. In the second and third images, our policy interacts with the object using the emergent pushing actions. We visualize the contact locations and normals on the rendered fingers. Our policy moves the object to a graspable location, and in the fourth image, the objects are successfully retrieved. 
Video demonstrations can be found on our project website at \url{https://jxu.ai/geotact}}.
\label{fig:real_demo}
\end{figure*}

In this section, we present and discuss the results of our real-robot and simulation experiments. Visit our project website at \url{https://jxu.ai/geotact} for video demonstrations and additional information. 

\subsection{Real-robot Results}

\begin{table*}
    \addtolength{\tabcolsep}{-3pt}
    \centering
    \begin{tabular}{c|ccccccc|c}
    \toprule
    \multirow[b]{2}{*}{\textbf{Method}} & \multicolumn{7}{c}{\textbf{Training Objects}} & \multirow[b]{2}{*}{\textbf{Average}}\\
    & Cube & \makecell{Long\\Cuboid} & Ball & Pentagram & L-shape & \makecell{Tomato\\Soup Can} & \makecell{Potted\\Meat Can} \\
    \midrule
    A2G & 3/10 & 3/10 & 0/10 & 2/10 & 4/10 & 1/10 & 4/10 & 0.24 \\
    \OURS & \textbf{7/10} & \textbf{8/10} & \textbf{1/10} & \textbf{6/10} & \textbf{6/10} & \textbf{3/10} & \textbf{7/10} & \textbf{0.54} \\
    \bottomrule
    \end{tabular}
    \caption{Real-world experiment results on 7 training objects. We show the number of successes out of 10 trials.}
    \label{tab:real-train}
\end{table*}

\begin{table*}
    \addtolength{\tabcolsep}{-3pt}
    \centering
    \begin{tabular}{c|cccccc|c}
    \toprule
    \multirow[b]{2}{*}{\textbf{Method}} & \multicolumn{6}{c}{\textbf{Unseen Objects}} & \multirow[b]{2}{*}{\textbf{Average}} \\
    & Clamp & \makecell{Toy\\Banana} & \makecell{Utility\\Knife} & \makecell{Screw\\Driver} & \makecell{Chocolate\\Box} & \makecell{Coconut\\Water Can} \\
    \midrule
    A2G & 3/10 & 3/10 & 3/10 & 2/10 & 4/10 & 4/10 & 0.32 \\
    \OURS & \textbf{8/10} & \textbf{6/10} & \textbf{6/10} & \textbf{5/10} & \textbf{7/10} & \textbf{6/10} & \textbf{0.63} \\
    \bottomrule
    \end{tabular}
    \caption{Real-world experiment results on 6 unseen objects. We show the number of successes out of 10 trials.}
    \label{tab:real-unseen}
\end{table*}

We evaluate two representative methods on the real robot --- our method, \OURS, which is learning-based and uses pushing, and A2G, which is heuristics-based and does not use pushing. We first evaluate both methods on the training set (seven objects) and unseen set (six objects) shown in Fig.~\ref{fig:objects}, performing ten trials on each object. The results on the training set and the unseen set are shown in Table~\ref{tab:real-train} and Table~\ref{tab:real-unseen}, respectively. 

\OURS achieves success rates of 54\%  and 63\% on the training set and unseen set, respectively, demonstrating strong generalization to unseen objects without any performance drop.
It outperforms A2G, which suffers from the significant amount of sensor noise introduced by granular media. In comparison, \OURS is more robust to sensing noise by continuously interacting with the object through emergent pushing actions and reacting to new contact observations. In addition, fine-tuned in granular media, \OURS is more adapted to the object motion discrepancy between tabletop and granular media. We also notice that the success rates of \OURS on two predominantly rounded shapes (ball and tomato soup can) are significantly lower compared to other objects. We discuss these failure cases, along with failure cases of A2G, later in Sec.~\ref{sec:failure}.

An example of \OURS's noise-robustness behavior can be seen in the chocolate box demonstration in Fig.~\ref{fig:real_demo}. There is a spurious pushing contact reading from the right (red) tactile finger even though the object is still some distance away. As our policy reacts to that spurious reading, the left (green) finger obtains a true contact. Our policy then pushes the object for a successful grasp. 

In additional testing on an extensive set of objects, we test our method on the set of 22 novel objects (not used in training) shown in Fig.~\ref{fig:objects}. For each object, we try \OURS until we obtain a successful retrieval. All 22 objects are successfully retrieved within at most three attempts. 

\subsection{Simulation Results}

\begin{table*}
    \addtolength{\tabcolsep}{-3pt}
    \renewcommand\cellset{\renewcommand\arraystretch{0.6}}
    \centering
    \begin{tabular}{c|ccccccc|c}
    \toprule
    \multirow[b]{2}{*}{\textbf{Method}} & \multicolumn{7}{c|}{\textbf{Training Objects}} & \multirow[b]{2}{*}{\textbf{Average}} \\
    & Cube & \makecell{Long\\Cuboid} & Ball & Pentagram & L-shape & \makecell{Tomato\\Soup Can} & \makecell{Potted\\Meat Can} \\
    \midrule
    A2G & 0.61 & 0.53 & 0.86 & 0.46 & 0.45 & 0.90 & \textbf{0.77} & 0.65 \\
    \OURS-tabletop & \textbf{0.96} & \textbf{0.80} & \textbf{1.00} & \textbf{0.96} & \textbf{0.89} & \textbf{0.97} & \textbf{0.77} & \textbf{0.91} \\
    \bottomrule
    \end{tabular}
    \caption{Simulation experiment results on the tabletop. Each success rate is computed over 1000 trials.}
    \label{tab:sim-tabletop}
\end{table*}

\begin{table*}
    \addtolength{\tabcolsep}{-3pt}
    \renewcommand\cellset{\renewcommand\arraystretch{0.6}}
    \centering
    \begin{tabular}{c|ccccccc|c}
    \toprule
    \multirow[b]{2}{*}{\textbf{Method}} & \multicolumn{7}{c|}{\textbf{Training Objects}} & \multirow[b]{2}{*}{\textbf{Average}}\\
    & Cube & \makecell{Long\\Cuboid} & Ball & Pentagram & L-shape & \makecell{Tomato\\Soup Can} & \makecell{Potted\\Meat Can} \\
    \midrule
    A2G & 0.22 & 0.54 & 0.49 & 0.21 & 0.45 & 0.48 & 0.35 & 0.39 \\
    A2G-push & 0.50 & 0.60 & 0.48 & 0.19 & 0.51 & 0.44 & 0.36 & 0.44 \\
    \OURS-tabletop & 0.18 & 0.41 & 0.13 & 0.08 & 0.73 & 0.06 & 0.33 & 0.27 \\
    \OURS-granular & 0.73 & 0.74 & 0.67 & 0.56 & \textbf{0.77} & 0.43 & 0.31 & 0.60 \\
    \OURS & \textbf{0.79} & \textbf{0.84} & \textbf{0.73} & \textbf{0.60} & 0.71 & \textbf{0.53} & \textbf{0.50} & \textbf{0.67} \\
    \bottomrule
    \end{tabular}
    \caption{Simulation experiment results inside granular materials. Each success rate is computed over 1000 trials.}
    \label{tab:sim-granular}
\end{table*}

We present more extensive experiments and ablations to analyze our method and the baselines in simulation, which lends itself to experiments and analysis on a larger scale.

First, in an open-air tabletop environment, we evaluate A2G and \OURS-tabletop using the same amount of noise as used in training (shown in Table~\ref{tab:sim-tabletop}). Note that we do not evaluate A2G-push on the tabletop, as A2G-push is designed to be robust to spurious contact readings, which do not happen in an open-air environment. Without the extra sensor noise and complex object motion introduced by granular media, \OURS-tabletop is achieving over 90\% success rate. A2G also has a higher success rate (65\%) compared to its evaluation in granular media on the real robot. However, A2G struggles with a few objects, such as L-shape, long cuboid, and pentagram, even in the tabletop environment, and we discuss these failure cases in detail in the next section. 

Inside granular media simulation, we evaluate all five methods using the same amount of tactile noise as training (Table~\ref{tab:sim-granular}). Working within granular media is much more challenging than the tabletop setup. As a result, the performance of both \OURS-tabletop and A2G drops significantly compared to tabletop simulation, from 91\% to 27\% and from 65\% to 39\%, respectively. A2G-push performs slightly better (44\%) than A2G, showing that adding heuristic push can reduce uncertainty and increase robustness.

The importance of fine-tuning inside simulated granular media can be shown by the performance improvement from \OURS-tabletop (27\%) to \OURS (67\%). Fine-tuning helps our policy adapt to the larger sensor noise and object motion discrepancy in granular media.

\begin{figure}[h]
\centering
\includegraphics[width=0.55\textwidth]{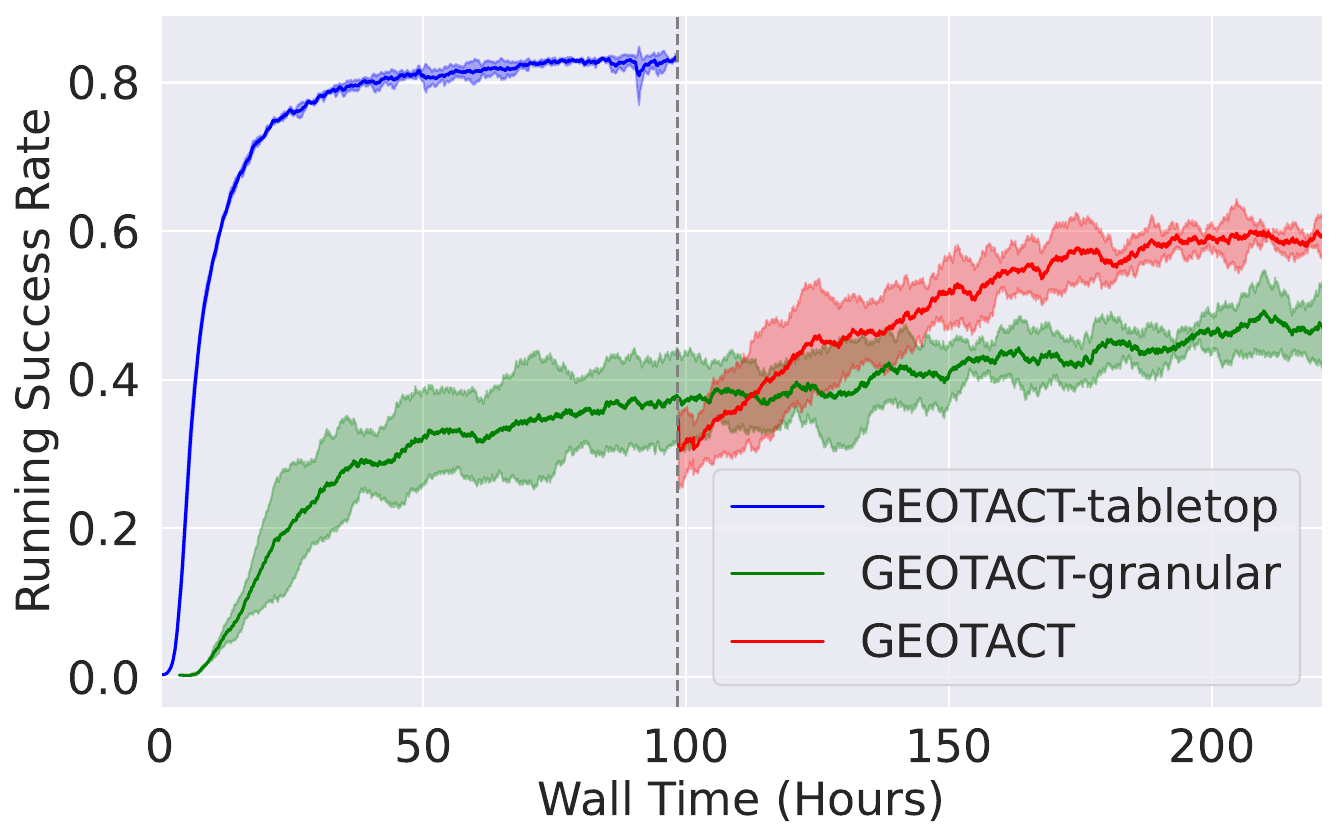}
\caption{Training plots of the curriculum strategy vs no-curriculum baseline. We show the running success rate (mean $\pm$ standard deviation) of the past 1,000 episodes during training, generated over five random seeds. Despite the initial performance drop when \OURS-tabletop is loaded into the granular media environment (indicated by the dashed line), \OURS quickly surpasses \OURS-granular and converges to a better performance.}
\label{fig:curriculum_training_plot}
\end{figure}

\begin{figure}[h]
\centering
\includegraphics[width=0.55\textwidth]{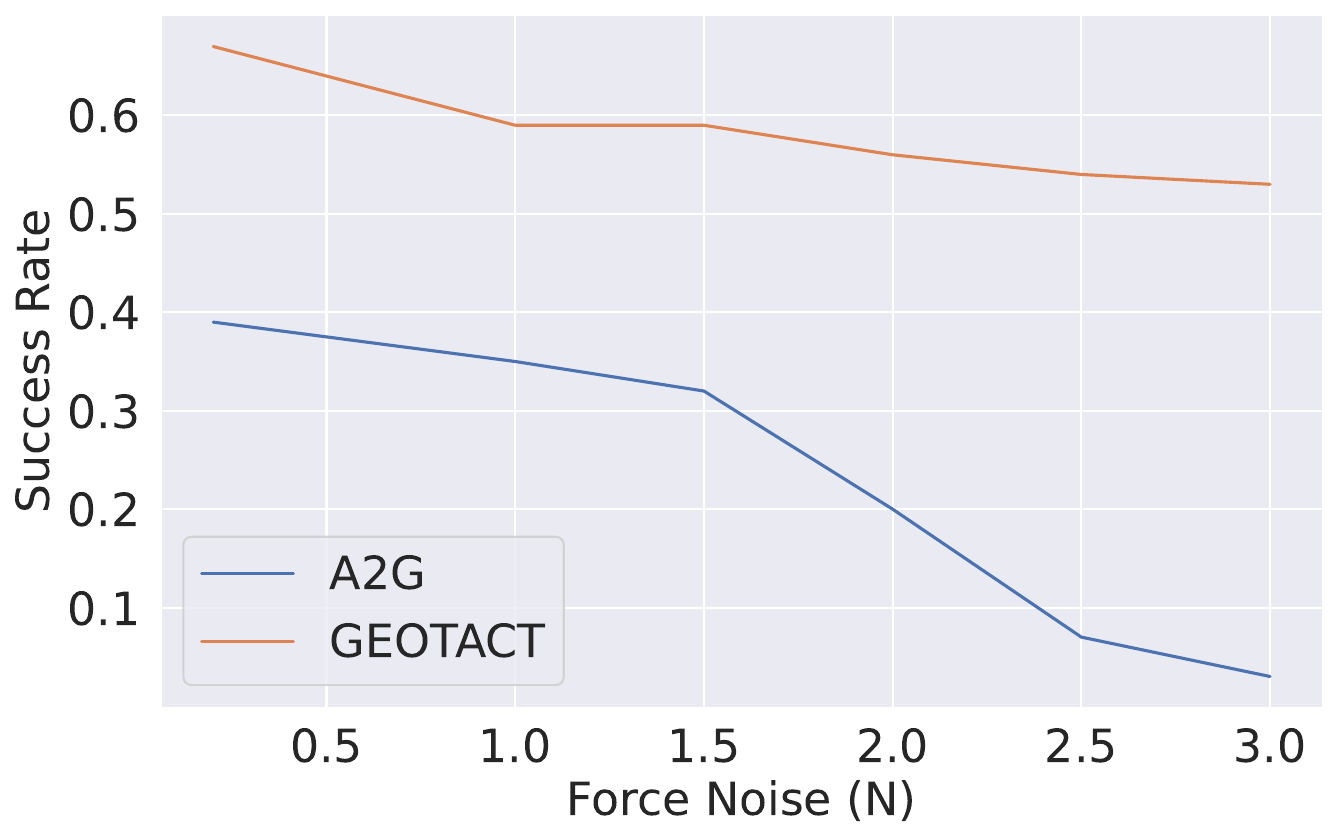}
\caption{Noise robustness analysis in granular media simulation. When evaluated on force noise larger than training, \OURS maintains above 50\% success rate even with 3N force noise, while A2G's performance drops to 3\%.}
\label{fig:noise}
\end{figure}

The performance gap between \OURS (67\%) and \OURS-granular (60\%) demonstrates the necessity of the proposed curriculum training strategy. Without pre-training on the tabletop, \OURS-granular performs worse than \OURS. We further show in Fig.~\ref{fig:curriculum_training_plot} the training plots of \OURS and \OURS-granular. After the initial performance drop when the tabletop policy is loaded into the granular media, the success rate of \OURS quickly surpasses \OURS-granular and eventually converges to higher performance. This experiment shows that pre-training on the tabletop is able to boost the training speed and improve the final performance of our policy. 

Finally, we then evaluate the noise robustness of \OURS, and compare it against A2G when faced with sensor noises larger than seen in training. 
We gradually increase the force noise range up to [-3N, 3N], and the success rate is shown in Fig.~\ref{fig:noise}. At a force noise level of [-3N, 3N], the success rate of A2G drops to 3\% while \OURS remains over 50\%. As an open-loop policy, A2G relies heavily on an accurate first contact. However, when the previous contact information is largely incorrect, \OURS is reactive and can continue interacting with the target object through emergent pushing behaviors.

\subsection{Failure Cases and Discussion}
\label{sec:failure}

\begin{figure*}[h]
\centering
\includegraphics[width=\textwidth]{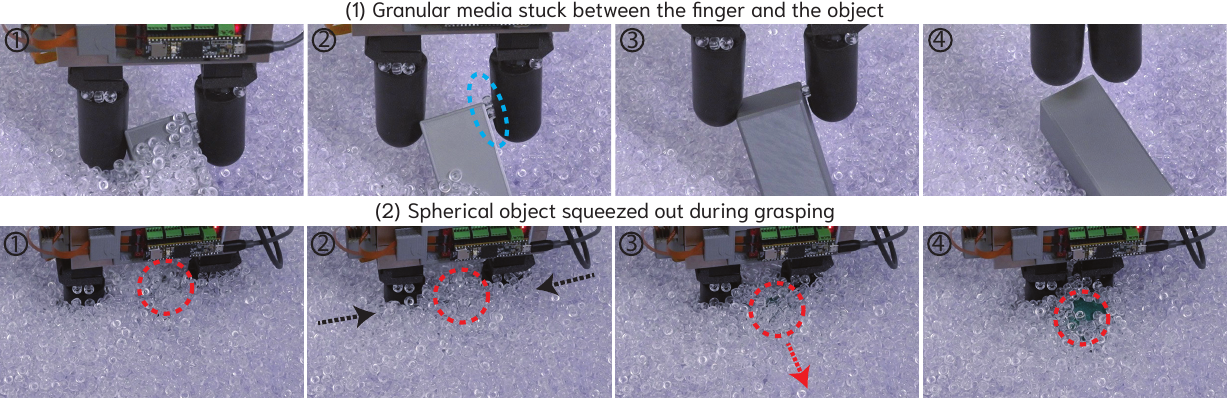}
\vspace{-3mm}
\caption{\OURS failure case demonstrations on the real robot. (1) There is a thin layer of granular beads (circled in blue) stuck between the finger and the object during grasping. Both the plastic surface and the beads are slippery, so the grasp fails. (2) Spherical and cylindrical objects are very hard to grasp with our parallel gripper with cylindrical fingers. They are always squeezed out even though a good grasp is formed. The location of the sphere is marked in red circles. The movement of the fingers and the sphere is marked with dashed lines with arrowheads.}
\label{fig:failure_cases}
\end{figure*}

\begin{figure}[!h]
\centering
\includegraphics[width=0.65\textwidth]{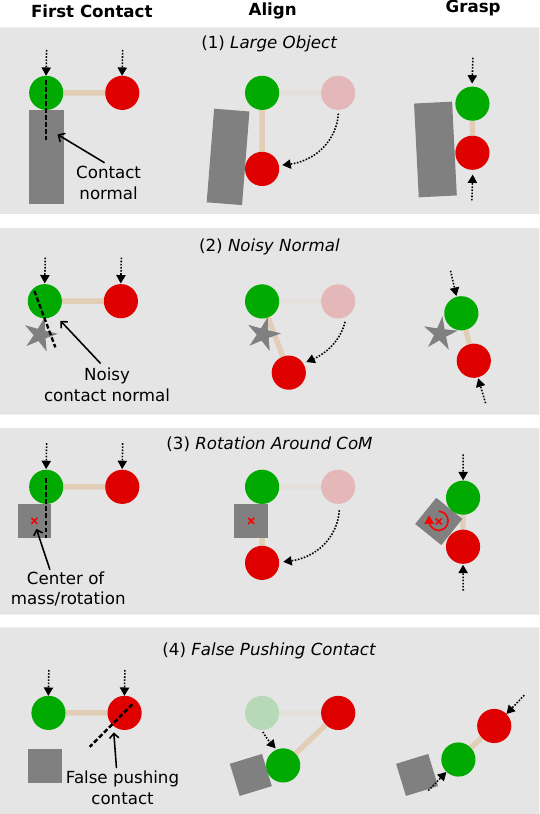}
\vspace{2mm}
\caption{Bird's-eye view of four failure scenarios for A2G. A2G consists of three stages: obtaining the first contact, aligning the gripper with the contact normal, and then closing the gripper. See Sec.~\ref{sec:failure} for a detailed explanation of these failure cases. 
}
\label{fig:a2g_failure}
\end{figure}

We notice that our method often fails on rounded shapes (e.g., ball, tomato soup can) when deployed on the real robot, despite a high success rate on the same objects in simulation. A successful grasp of these two objects is extremely sensitive to sensor error, due to their rounded shape. On the real robot, our policy locates both objects well, but the gripper simply often squeezes them out during closing, as shown in Fig.~\ref{fig:failure_cases}(2). A very precise grasp (centered on the object) is necessary, which is easier to achieve in a simulated environment with perfect control and low sensor noise. This is also partly an inherent limitation of a parallel gripper with cylindrical fingers attempting to grasp a round object. A multi-fingered hand might work better for such objects, and we leave it for future work.

The two objects, ball and tomato soup can, is exactly the reason why our method has a higher success rate on unseen objects than training objects. Despite the fact that these two objects are seen during training, they are inherently hard to grasp due to the limitation of the parallel gripper with cylindrical fingers. If we exclude these two objects, the remaining objects in the training set actually have a higher average success rate than unseen objects. Most of our unseen objects have irregular shapes and are not cylindrical or spherical, which is not challenging from a hardware perspective. The coconut water can was originally cylindrical, but it was squeezed and lost its perfect cylindrical shape.

Another common failure case of our policy on the real robot is due to a thin layer of granules stuck between the target object and the finger, as shown in Fig.~\ref{fig:failure_cases}(1). When this happens, due to the low friction coefficient with the granules, the object will slip out of the formed grasp. Again, this is partly a limitation of our parallel gripper, and a multi-fingered gripper might be able to relieve this problem. We will leave this to future work. 

A2G also fails for the same two reasons above, but in addition, it fails in another four cases, as shown in Fig.~\ref{fig:a2g_failure}. (1) \textit{Large Object}: For large objects such as L-shape or long cuboid, because the maximum length of the object exceeds the span of the gripper, the gripper can only grasp successfully along the short dimension. Simply aligning the gripper with the contact normal is likely to push the object away. (2) \textit{Noisy Normal}: The contact normal can be noisy, especially when contacting a pointy/sharp edge of an object, such as that of the pentagram, making the aligned grasp unstable. This failure case happens much more often in granular media than on the tabletop. (3) \textit{Rotation Around CoM}: During the grasping phase of A2G, the finger contacting the object will start pushing the object, and then the object will start rotating around its center of mass. This rotation can result in the object sliding outside the grasping closure. (4) \textit{False Pushing Contact}: When there is a spurious pushing contact being reported by the finger, A2G does not have any recovery behavior and will align to the force direction of the false contact, leading to an improper grasp. This failure case is specific to granular media since there is no ubiquitous contact to filter on the tabletop.

In comparison, \OURS is resilient to all these failure types. We attribute this to the combination of its reactive nature, obtained via end-to-end training in the presence of noise, and the emergent pushing actions that reduce the uncertainty of the object location and funnel the object into a stable grasp.

\section{Chapter Summary}

Compared to tactile object identification, object retrieval from granular media is undoubtedly a more challenging task, and granular media is such a sensor-deprived environment that vision is completely occluded. In addition to the data sparsity problem that is fundamental to tactile manipulation, this task presents additional challenges such as large uncertainty and significant sensor noise, introduced by the ubiquitous contacts inside granular media. 

GEOTACT formulates this task as a model-free reinforcement learning problem and trains it end-to-end in simulation with a curriculum strategy. Our formulation of the learning task action space leads to emergent pushing behaviors that efficiently explore under granular media with sparse tactile signals and help reduce uncertainty. Compared to heuristic-based methods, our learned policy can incorporate simulated sensor noise during training and is thus more robust to spurious and noisy tactile readings. Despite being trained only on 7 objects with simple shapes in simulation, we show that our method can zero-shot transfer to the real hardware and successfully pick up 28 unseen objects, including rigid, deformable, and articulated objects with various complex shapes. 

\newpage

%% file: chapters/part2.tex


\part{Addressing Data Scarcity in Rehabilitation Robots}
\label{part:scarcity}

In the second part of this thesis (Chapters~\ref{chap:semiemg}, \ref{chap:metaemg}, \ref{chap:reciprocal_learning}, \ref{chap:chatemg}), we switch gears and discuss our effort in addressing data scarcity in rehabilitation robots. While the data sparsity challenge (as discussed in Part~\ref{part:sparsity}) focuses on the quality (representation) of data, highlighting the challenge from the empty space or non-relevant information presented in robotic signals, data scarcity focuses on the quantity of the data. The data scarcity challenge is about the limited amount of data we can collect from physical hardware and humans. The rehabilitation robot domain is a great platform to study the data scarcity problem, as among all applications that require data collection, collecting data from people with disabilities is the most difficult.

We specifically focus on the application of intent inferral for stroke in the rehabilitation robot domain, as an effective intent inferral mechanism is considered to be an intuitive way to control an assistive and rehabilitative device. Our first work SemiEMG (Chapter~\ref{chap:semiemg}) explores semi-supervised learning that exploits unlabeled data. It is an effective way to reduce the amount of labeled data needed and adapt to intrasession concept drift. In order to also handle intersession drift, we propose MetaEMG (Chapter~\ref{chap:metaemg}), and it models the intent inferral problem as a multi-task learning problem and uses meta-learning to train a classifier that adapts fast to a new session or subject. 

We then realized that intent inferral should be bi-directional --- the intent classifier should learn from the patient, and the patient should also adapt to the classifier to generate higher-quality control signals. Thus, we propose reciprocal learning (Chapter~\ref{chap:reciprocal_learning}). Last but not least, we want to take advantage of the power of high-capacity models such as Transformers, which heavily rely on large training sets, so we propose ChatEMG (Chapter~\ref{chap:chatemg}). ChatEMG mediates the data scarcity challenge by generating context-specific synthetic data, which can work with any type of classifier and is particularly beneficial to large-capacity models. 


%% file: chapters/semiemg.tex
\chapter{Semi-supervised Learning for Utilizing Unlabeled Data}
\label{chap:semiemg}

In this chapter, we discuss SemiEMG, which adopts semi-supervised learning to address the data scarcity problem in intent inferral by learning from unlabeled data.

\section{Motivation} 

\begin{figure}[t]
    \centering
    \includegraphics[width=0.65\textwidth]{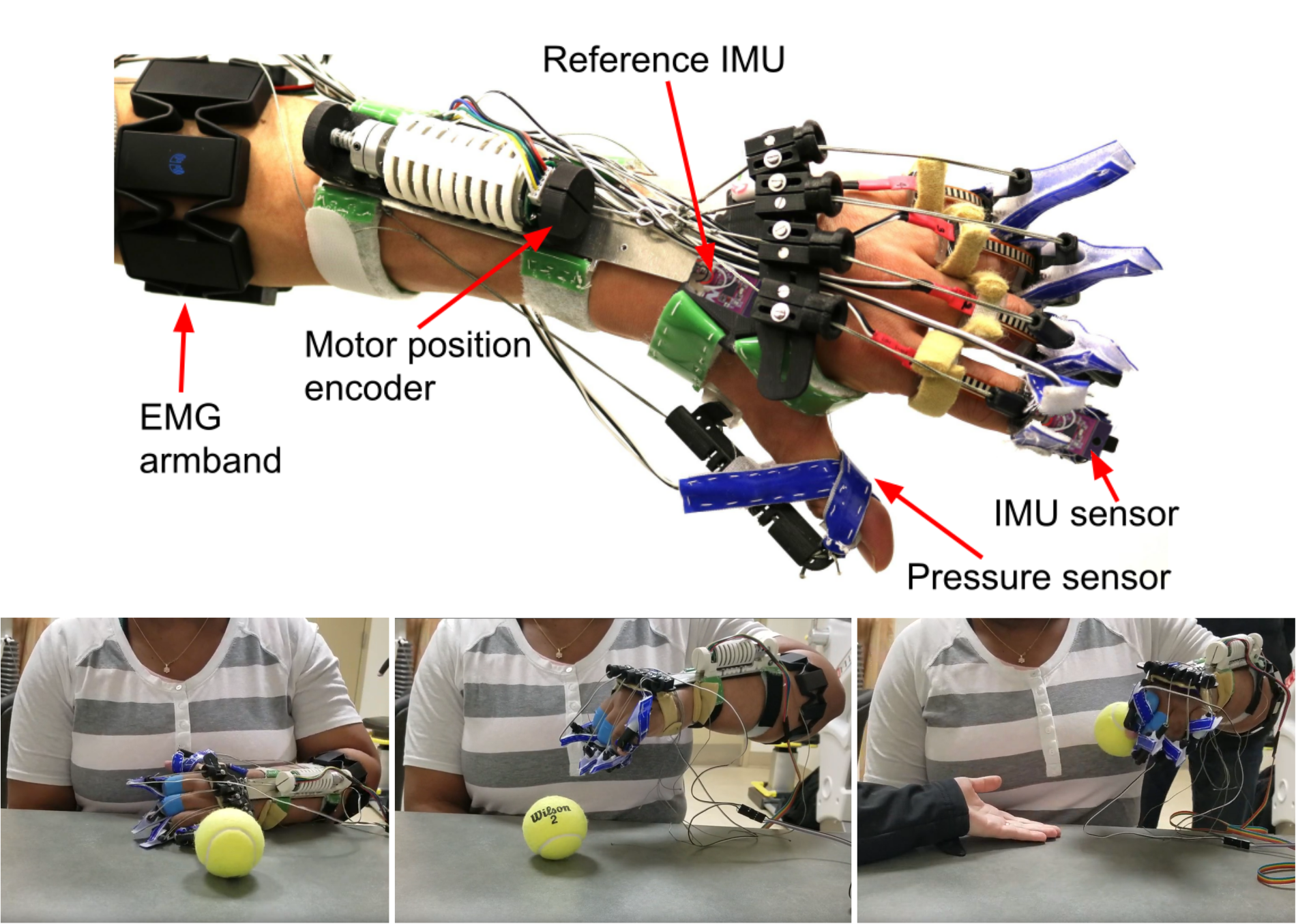}
    \caption{Top: hand orthosis with multimodal sensing suite. Bottom: stroke subject performing an assisted grasp. Due to abnormal synergies, muscle activation
    signals change significantly compared to collected training data, a type of concept drift that must be accounted for during intent inferral.}
    \label{fig:updated_orthosis}
\end{figure}

Data scarcity presents a fundamental challenge in inferring intent in stroke patients, and this issue is further compounded by the presence of concept drift, i.e., the phenomenon where the measured biosignals change over time. Concept drift can occur between different sessions (intersession drift) or gradually within a single session (intrasession drift). In healthy subjects and amputees, concept drift is caused by fatigue and repositioning of the sensors, among other factors~\cite{jain2012improving}. In stroke subjects, concept drift is additionally aggravated by abnormal muscle coactivation~\cite{miller2012} and by the interaction of the hand and the robotic device (Figure~\ref{fig:updated_orthosis}).

While we have found concept drift to be a significant problem for the stroke population, it is also rarely addressed in the literature. Supervised learners compensate for concept drift by using a training set that incorporates data with as much signal variation as possible. In our own previous work cited above, we trained with data for many arm poses and orthosis states. However, such training comes at a high cost, since it is manually labeled, requiring the user to generate motions and labels specifically for training. This places a substantial burden on the user, especially in the case of stroke subjects, who fatigue quickly~\cite{riley2002changes}. As a result, concept drift exacerbates the issue of data scarcity.

This paper focuses specifically on the intrasession drift with semi-supervised learning. Semi-supervised learning provides a potential solution, as it uses a small labeled dataset and then exploits additional unlabeled data to improve classifier performance. When intrasession drift happens, semi-supervised learners can adapt themselves to the changes in the input data. We hypothesize that these controls will require less training data than supervised controls, while maintaining high accuracies.

The main contributions of this paper are as follows. We propose a disagreement-based semi-supervised learning algorithm to help our orthosis adapt to intrasession concept drift when the device migrates, the subject gets fatigued or changes their arm poses, etc. To the best of our knowledge, \textit{we are the first to propose a semi-supervised control for a hand orthosis and validate its feasibility with functional tasks}. We evaluate the performance of our algorithm on data collected from five stroke subjects using multimodal sensing. We show that semi-supervised controls can adapt to intrasession drift with new unlabeled data and reduce the burden placed on the subject during training for ipsilateral hand controls. In the functional task experiments, two subjects successfully picked up and handed over a block multiple times in a minute with our proposed algorithm.

\section{Method}

In this section, we discuss how to use a semi-supervised learning algorithm to improve our intent detection accuracies. Semi-supervised learning exploits unlabeled data to update the intent detection classifier. 
We hypothesize that this can make the intent detection robust to intrasession concept drift caused by fatigue, arm movement, device migration, etc., by constantly updating itself with new information. 

For any semi-supervised algorithm, we require an oracle. The oracle determines which new data sample should be used to update the classifier and labels the data. The data samples labeled by the oracle assemble into a training dataset. Since we use an ensemble classifier, we generate a training dataset $X^i$ for every base learner $i$ in the ensemble. When the aggregated dataset $X = \bigcup_{i=1}^{\eta}X_{i}$ contains a prespecified amount of data, we use the training data and labels from the oracle to update the classifier.

We would like our semi-supervised learning algorithm to address intrasession concept drift. Intrasession concept drift can be a sudden and large redistribution of the data in feature space (principally caused by subjects moving the arm to an unseen pose during training). It can also be a gradual shift in the data over time (primarily caused by device migration or subject fatigue). The proposed semi-supervised learning algorithm needs to be robust to these scenarios.

\subsection{Disagreement-based Oracle}
The semi-supervised learning algorithm presented here uses a disagreement-based oracle. When intrasession concept drift occurs, we hypothesize that the oracle can leverage the disagreement between multiple learners. Some of the base learners with a particular set of modalities will be more robust to changes in the data during the drift and will remain confident. Other learners which are not robust to the drift are corrected by the confident learners.

Our proposed disagreement-based semi-supervision is enacted as follows. At time $T$, we calculate the entropy $E^i_T$ of each base learner $i$:
\begin{gather}
E^i_T = - {\boldsymbol{\hat{\Phi}}^i_T}^\top log_k(\boldsymbol{\hat{\Phi}}^i_T)
\end{gather} 
where $k = 3$ is the number of possible intent classes. We use entropy as a measure of the learner's confidence. Lower entropy indicates higher confidence. 

Confident base learners are used to correct less confident learners. We define confident learners as those whose \mbox{$E^i_T < 0.2$.} Learners which are not confident have $E^i_T > 0.8$. We select our confidence thresholds empirically. 

If all confident learners agree on the subject's intent (i.e., if $\argmax(\boldsymbol{\hat{\Phi}}^i_T)$ is the same for all confident learners), for each unconfident learner $i$, we add $\boldsymbol{f}^i_T$ to $X^i$, along with the intent label agreed upon by the confident learners.

Once the combined number of training samples across all $X^i$s is a sufficiently large value (we choose 200 as the threshold), the data and labels are used to update the classifier, the supervision process starts, and all $X^i$s are reset.

Oracles can also correct the classifier prediction (before the base learners are updated). With disagreement-based semi-supervision, we only want confident learners to contribute to the final output of the ensemble. However, only including confident learners whose $E^i_T < 0.2$ is too restrictive, so we have an additional empirically-selected threshold for the correction. We calculate the final probability for the ensemble as an average of all the probabilities from learners whose entropy is less than $0.6$:
\begin{equation}
\boldsymbol{\bar{\Phi}}^{ens}_{T}=\frac{1}{\eta} \times \sum\limits_{\substack{i=1 \\ E^i_T < 0.6}}^{\eta} \boldsymbol{\hat{\Phi}}^i_{T}
\end{equation}
If there are no base learners whose $E^i_T < 0.6$, then $\boldsymbol{\bar{\Phi}}^{ens}_{T}$ is calculated using all base learners in the ensemble.
 
We are the first to use disagreement-based semi-supervision for an assistive robot. We can successfully leverage this paradigm for a novel application because our orthosis includes a multimodal sensing suite with independent sensing modalities~\cite{park2019}. Disagreement-based semi-supervision works best if the base learners include multiple independent views~\cite{blum1998combining}, or include a large number of base learners~\cite{zhou2005tri}. Therefore, we use ensembles with at least five base learners whose features are sampled randomly from all the sensors in the multimodal sensing suite.

\subsection{Updating the Classifier}

Once the aggregated dataset $X$ across all $X^i$s has a sufficient number of data samples, we update our classifier. We use linear discriminant analysis (LDA) for our base learners because it does not need past training data for updates. An LDA base learner $i$ with $k$ classes has the following parameters: a mean vector for each class $\boldsymbol \mu^i_k$ and a covariance matrix $\boldsymbol \Sigma^i$ (LDA assumes that the covariance matrices are identical across all classes). 

To update our ensemble, we update each base learner $i$ independently using the dataset $X^i$ collected for that learner. To update the parameters of an LDA base learner $i$, let $\boldsymbol z_k$ be a sample from $X^i$ whose label is class $k$. The updated mean vector for class $k$, namely $\widetilde{\boldsymbol \mu}_k^i$, and the updated covariance matrix $\widetilde{\boldsymbol \Sigma}^i$ for the learner are calculated as follows:
\begin{gather}
\widetilde{\boldsymbol \mu}^i_k = \frac{n^i_k \times \boldsymbol \mu^i_k + \boldsymbol z_k}{n^i_k + 1} \\ 
\widetilde{\boldsymbol \Sigma}^i = \frac{N^i}{N^i+1}\boldsymbol \Sigma^i + \frac{1}{N^i+1} \times \frac{n^i_k}{n^i_k + 1}(\boldsymbol z_k-\boldsymbol \mu^i_k)(\boldsymbol z_k-\boldsymbol \mu^i_k)^\top
\end{gather}
where $N^i$ is the number of training samples for the base learner $i$ so far, and $n^i_k$ is the number of training samples for the base learner $i$, labeled as class $k$ so far.

\section{Experiments}

\subsection{Data Collection} 
\label{par:data_collection}

We have two data collection protocols: a complete protocol and an abbreviated protocol. For the complete protocol, we collect data for all three intents under different conditions: 1) with the arm resting on a table and the orthosis motor off \{\textit{arm on table motor off}\}, 2) with the arm raised above the table and the orthosis motor off \{\textit{arm off table motor off}\}, and 3) with the arm raised above the table and the orthosis motor on, actively moving the hand \{\textit{arm off table motor on}\}. Specifically for the third case, Table~\ref{tab:modified_train} shows the commands given to the subject and the ground truth intent collected while the motor retracts and extends the tendons.  In the abbreviated protocol, we only collect data from condition three in the complete protocol. For each condition, we ask the subject to open and close their hand three to four times. 
 
\begin{table}
\centering
\footnotesize 
\begin{tabular}{c|ccc}
Device State & \multicolumn{3}{c}{Subject Instruction}\\
&Open&Relax&Close\\\hline\hline
Tendon extended 	& O & R & C \\
Tendon retracting 	& O &   & C \\
Tendon retracted 	& O & R & C 
\end{tabular}
\caption{Training protocol: for each combination of instruction and exotendon state, the table shows the assigned label. We begin with the tendon extended and the subject relaxing (top row, middle column) and proceed 
 counter-clockwise.}
\label{tab:modified_train}
\end{table}

During data collection, the experimenter collects the true subject intent, or ground truth, while providing verbal commands to the subject. For conditions where the motor is on, we move the motor approximately one second after the verbal command is given.

\subsection{Baselines} 

In our experiments, we test whether a classifier can be trained using less data and still achieve high accuracies during intent detection. We collect one dataset using the abbreviated protocol. We collect four datasets using the complete protocol --- one trains the \textit{SE-full} baseline control, and the others are testing datasets. Specifically, we evaluate four methods:

\begin{itemize}
    \item \textit{Supervised EMG, full training data (SE-full)}. The only classifier trained on labeled data from the complete collection protocol, including data from all three conditions. Classifiers trained under the same conditions as the test data are expected to have high accuracies. We consider this classifier as a baseline, as the assumption of labeled data from the complete protocol is impractical due to high burden on the patient. $\eta=1$ and $\boldsymbol f=[e^1, ... , e^8]^\top$. Classifier parameters do not change after training.
    \item \textit{Supervised EMG, partial training data (SE-partial)}. Trained on the abbreviated protocol. $\eta=1$, and $\boldsymbol f=[e^1, ... , e^8]^\top$. Classifier parameters do not change.
    \item \textit{Supervised Multimodal, partial training data (SM-partial)}. Trained on the abbreviated protocol but uses multimodal sensing. $\eta$ is a random number between 5 and 10. The features for each base learner are selected randomly from $\boldsymbol x$. The parameters of this classifier do not change.
    \item \textit{Disagreement Semi-Supervised Multimodal, partial training data (DSSM-partial)}. This is our approach, only requiring labels on the abbreviated protocol. Initially, this classifier is the same as \textit{SM-partial}. However, as new data arrives, it is labeled by the disagreement-based oracle and used to update the classifier. 
\end{itemize}

\begin{table}
    \footnotesize
    \label{tab:intrasession_results}
    \centering
    \begin{tabular}{c|ccccc|c|c}
    \toprule
    \textbf{Controls} & 
    \begin{tabular}[c]{@{}c@{}}\textbf{Subject S1}\\Female, 83 \end{tabular} & 
    \begin{tabular}[c]{@{}c@{}}\textbf{Subject S2}\\Male, 71 \end{tabular} 
    & 
    \begin{tabular}[c]{@{}c@{}}\textbf{Subject S3}\\Female, 51 \end{tabular} & 
    \begin{tabular}[c]{@{}c@{}}\textbf{Subject S4}\\Male, 29 \end{tabular} &
    \begin{tabular}[c]{@{}c@{}}\textbf{Subject S5}\\Male, 51 \end{tabular} &
    \textbf{Average} &
    \begin{tabular}[c]{@{}c@{}}\textbf{$p$-value w\slash}\\\textbf{DSSM-partial} \end{tabular}\\
    \midrule
    SE-full & 72.8 $\pm$ 8.1 & 85.6 $\pm$ 4.6 & 63.1 $\pm$ 8.05 & 69.8 $\pm$ 0.81 & 73.8 $\pm$ 3.05 & 73.0 & $3\mathrm{e}{-4}$ \\
    SE-partial & 72.1 $\pm$  4.6 & 81.9 $\pm$ 10.3 & 62.9 $\pm$ 7.92 & 69.8 $\pm$ 7.82 & 70.7 $\pm$ 5.08 & 71.5 & $1\mathrm{e}{-4}$ \\
    SM-partial & 72.7 $\pm$ 5.7 & 80.3 $\pm$ 5.4 & 71.0 $\pm$ 3.76 & 68.3 $\pm$ 0.35 & 73.4 $\pm$ 2.23 & 73.2 & $6\mathrm{e}{-4}$ \\   
    DSSM-partial & \textbf{79.2 $\pm$ 4.4} & \textbf{85.8 $\pm$ 4.0} & \textbf{71.8 $\pm$ 4.10} & \textbf{76.5 $\pm$ 2.65} & \textbf{82.9 $\pm$ 1.34} & \textbf{79.3} & --- \\
    \bottomrule
    \end{tabular}
    \caption{Classification accuracy and standard deviation in percentage (\%) for 5 stroke subjects. For each subject, we also provide their genders and ages. We also report the average accuracy across all subjects. The best result is in bold-text. We perform a one-sided Wilcoxon rank sum test on the aggregated results from all subjects and show the computed $p$-values for pairwise differences between \textit{DSSM-partial} and  the  other methods.}
\end{table}

We use the above classifiers to predict the subjects' intent for the three testing datasets collected with the complete protocol. We select the multimodal features randomly, but in our experience, this does not notably affect performance. 

\section{Results and Discussion}

For each subject and method, we report the results as mean and standard deviation across the three testing datasets. 
We report the resulting motor-command accuracy: the number of time points at which the command to the motor is correct, divided by the number of total time points. We report motor-command accuracy instead of global accuracy (how often the classifier gets each intent class correct) because we care less about the correct intent than about moving the orthosis as the subject intends.

As shown in Table~\ref{tab:intrasession_results}, our proposed method outperforms all other methods, even over the baseline classifier using complete protocol data. This suggests that semi-supervised learning could make our control more robust to intrasession drift while reducing the training burden. To examine the statistical significance in the difference between our proposed method and the other methods, we perform a one-sided Wilcoxon rank-sum test \cite{wilcoxon1992individual} on the results aggregated across all subjects, using a hypothesis threshold $\alpha = 0.01$. We choose a non-parametric statistical test because we do not assume an underlying normal distribution. Table~\ref{tab:intrasession_results} shows the computed $p$-values for pairwise differences between our proposed method and the other methods. We find all $p$-values to be $< 0.01$; thus, we conclude that the difference between \textit{DSSM-partial} and others indicates a statistically significant improvement in prediction accuracy. 

In the experiments, contrary to our expectations, \textit{SE-full} is not drastically better than \textit{SE-partial}. This is particularly true for subjects S1, S3, and S4. For some stroke subjects, even though we collect datasets following the same protocol, there could be some drift between datasets, caused by subject fatigue, device migration, etc. Given a large drift, the benefit of having more training data is limited. \textit{SE-full} is trained with complete protocol on three conditions (\{\textit{arm on table motor off}\}, \{\textit{arm off table motor off}\}, and \{\textit{arm off table motor on}\}), and the \textit{SE-partial} is trained with only \{\textit{arm off table motor on}\}. It is possible that \textit{SE-partial} is generalizing well enough, even without data from the other two conditions.

The comparison of \textit{SM-partial} and \textit{SE-partial}, which are both trained on a dataset collected with abbreviated protocol, highlights the importance of multimodal sensing and ensemble methods. Ensemble methods have been shown to improve the robustness of machine learning algorithms when large uncertainty is presented and having multiple sensing modalities makes its advantages more pronounced. Even without the disagreement-based updates, the classifier accuracy is improved with multimodal data and an ensemble of base learners. The comparison of \textit{DSSM-partial} and \textit{SM-partial} demonstrates the importance of disagreement-based semi-supervised updates. Despite being trained on partial data, \textit{DSSM-partial} learns from new unlabeled data using semi-supervision and further improves its performance.


Overall, our experiments show that semi-supervised learning has the potential to reduce the burden placed on users by controls that require an extensive labeled dataset to detect intent. Our proposed algorithm provides an improvement in accuracy when the classifier is trained on a more limited dataset and even outperforms the classifier trained on a larger dataset.


\begin{figure}
\centering
\includegraphics[width=\textwidth]{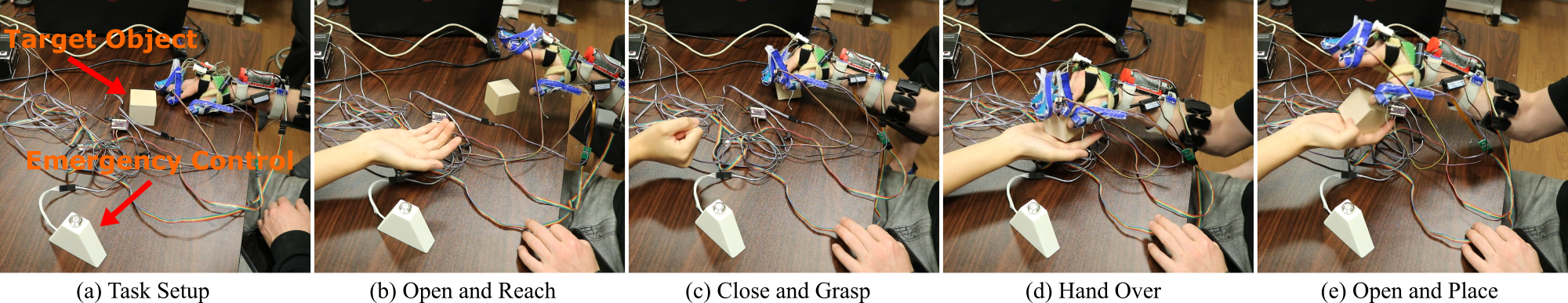}
\caption{Example of subject S5 performing pick-and-handover functional task. (a) The subject is instructed to pick up the wooden block and hand it over to an experimenter. We also provide a button for emergency override of classifier control. The orthosis tendon retracts (hand opens) when the button is pressed and the tendon extends (hand closes) when the button is released. (b) An open signal is detected, the tendon retracts and the subject tries to reach the target object. (c) A close signal is detected, the tendon extends and the subject grasps the block. (d) The subject moves the hand carrying the block to the experimenter. (e) An open signal is detected, the tendon retracts and the subject places the block.}
\label{fig:functional}
\end{figure}

\subsection{Functional Tasks}
We conducted a pick-and-handover functional task with subjects S4 and S5 to validate the feasibility of using our proposed algorithm in real time on the hand orthosis. With the orthosis running the disagreement-based semi-supervised algorithm, we instructed each subject to reach and lift a wooden block, then hand over the block to an experimenter who replaced the block back on the table. We continuously repeated this task until one minute elapsed. Both subjects were able to use the orthosis to pick up and hand over the block seven times in a minute. A demonstration of the functional task is shown in Figure~\ref{fig:functional}, and a recording of this experiment can be found in the accompanying video\footnote{The video and more information can be found at \mbox{\url{https://roamlab.github.io/dssm}}}. In the case of subject S4, we observe a number of stops and starts for the first several grasping motions, but this issue is soon alleviated and the control becomes smoother in the later grasping motions, possibly due to adaptive updates. In informal feedback, both subjects found the control to be intuitive and responsive, although we have not quantified this impression using a standardized questionnaire.

\section{Chapter Summary}

Concept drift exacerbates the data scarcity problem in intent inferral, as it requires the training set to incorporate labeled data with as much signal variation as possible. Such data collection comes at a high cost since it is manually labeled, putting a lot of burden on the stroke subjects. SemiEMG addresses intrasession concept drift through exploiting unlabeled data. We propose a disagreement-based semi-supervised algorithm that automatically labels a stream of new unlabeled data in an online fashion. To our knowledge, this is the first time a semi-supervised learning algorithm has been proposed and used for a hand orthosis based on multimodal ipsilateral sensing. We are also the first to use the proposed algorithm for functional tasks.

Our experiment results show that semi-supervised learning is a promising avenue of exploration for orthotic hand controls. They also indicate areas of future research, including handling intersession concept drift that happens between two different sessions, which we will discuss in the next chapter.

%% file: chapters/metaemg.tex
\chapter{Meta-learning for Rapidly Adaptable Representations}
\label{chap:metaemg}

In this chapter, we discuss MetaEMG, a novel paradigm in which we model intent inferral as a multitask-learning problem and we treat each session/subject as a single task. We use meta-learning to learn an intermediate representation that can adapt to a new task with limited training data and epochs. 

\section{Motivation}

In the last chapter, we discussed our work that utilizes unlabeled data through semi-supervised learning. While semi-supervised learning can help adapt to concept drift within the same session, the heuristics to label incoming unlabeled data will generally fail for a new session. In order to address the data scarcity problem by quickly adapting to not only the intrasession but also the intersession drift, we propose a meta-learning (or learning to learn) approach to training intent inferral models on a robotic orthosis. With this approach, a model can be explicitly optimized to adapt to new environments and tasks when labeled training examples are limited. In the context of rehabilitation robots, this means training intent inferral models that can quickly adapt to new subjects and new sessions with as little subject- or session-specific data as possible. Our directional goal is to train models that are as effective as possible in making use of the limited data available for any new patient or session.

Specifically, we leverage real clinical data collected from different subjects and different orthosis use conditions in order to train a base model for intent inferral using Model-Agnostic Meta-Learning (MAML)~\cite{finn2017model}. The MAML algorithm explicitly rewards a base model that is easily adaptable when presented with new data. When the intent inferral model must be used for a new patient or on a new day, we only need a small set of new, session-specific training data to fine-tune the base model to the new condition. We find that our method, which we dub MetaEMG, is particularly effective at learning good intent classifiers with only a few training epochs, and that classifiers trained via MetaEMG are better at adapting to new subjects. In summary, our main contributions are as follows:

\begin{figure}
    \centering
    \includegraphics[width=0.65\linewidth]{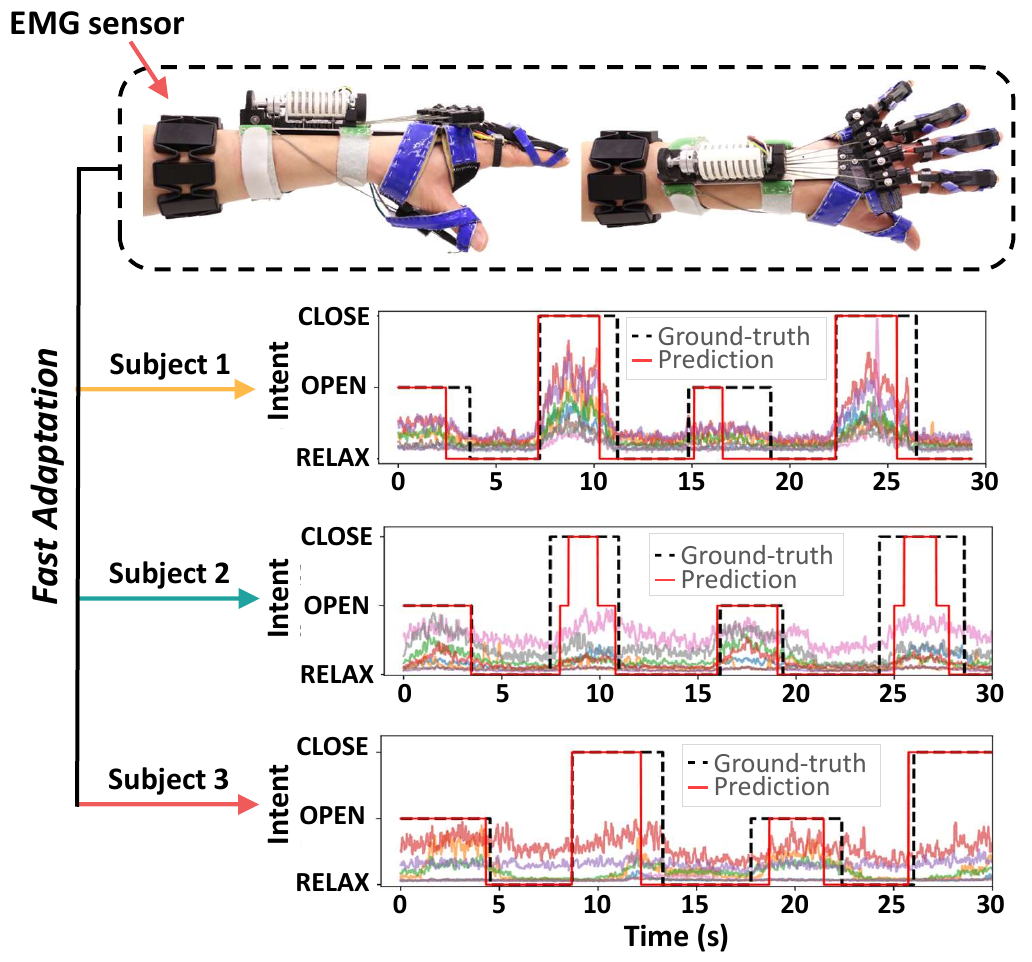}
    \caption{MetaEMG for fast adaptation on three subjects with different EMG patterns. Our method trains EMG-based intent inferral models that can quickly and efficiently adapt to new subjects in the context of a wearable robotic orthosis. Shown are EMG signal recordings of three different stroke survivors, and each colored series in the plots represents the reading of one of the eight EMG electrodes. Training models on new stroke subjects is difficult due to the large subject-to-subject variation in EMG signaling.}
    \label{fig:teaser_metaemg}
\end{figure}

\begin{itemize}
    \item To the best of our knowledge,
    we are the first to formulate intent inferral on stroke subjects as a meta-learning problem. Our proposed method, MetaEMG, achieves fast adaptation to a new session or subject with only a few training samples and epochs. 
    \item We evaluate the performance of our method on data collected from five stroke subjects. We show that our method outperforms baselines whose models are not meta-learned or are only trained on the session or subject-specific data.
\end{itemize}

\section{Method}

\begin{figure}
    \centering
    \includegraphics[width=0.65\linewidth]{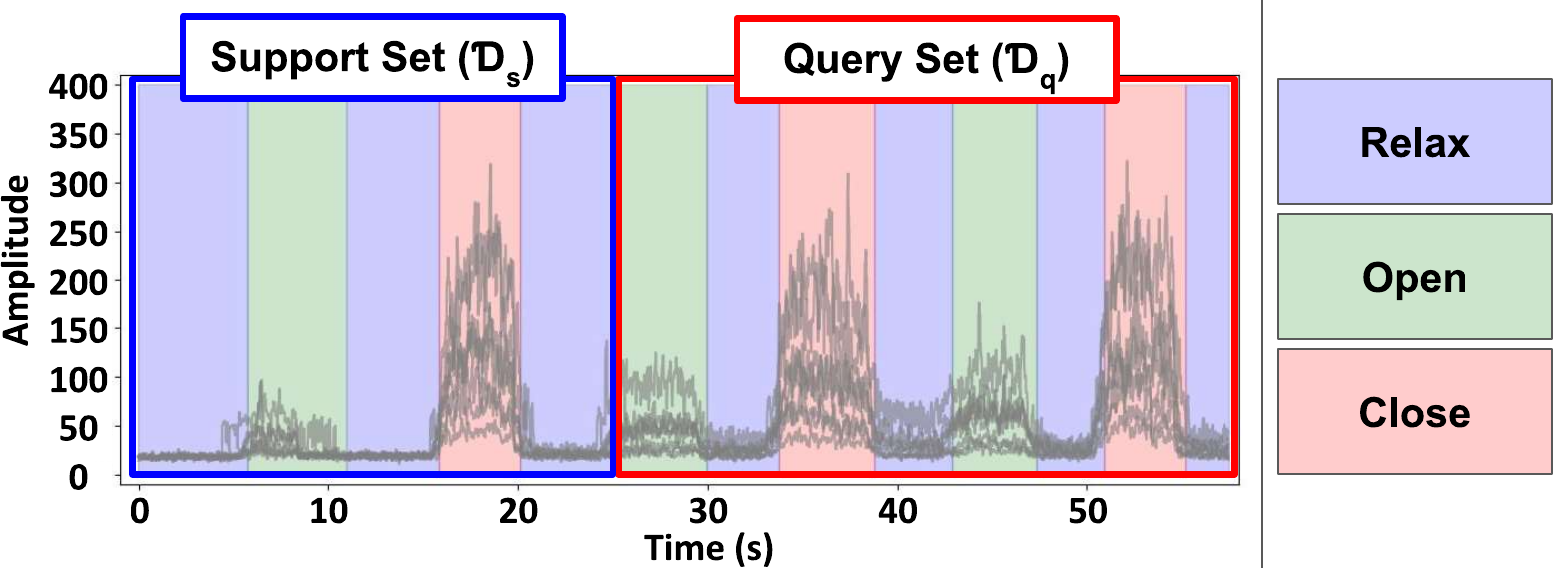}
    \caption{Task visualization in EMG intent inferral. We define an EMG task as a single uninterrupted recording with the 8-channel EMG armband. During each recording, users close and open their hands three times. The first open-relax-close motion (outlined in blue) is the support set. The third and second open-relax-close motions (outlined in red) consist of the query set. The ground-truth intent (verbal cues) is shaded in blue, green, and pink for relax, open, and close, respectively.}
    \label{fig:task}
\end{figure}

\begin{figure*}
    \centering
    \includegraphics[width=1\linewidth]{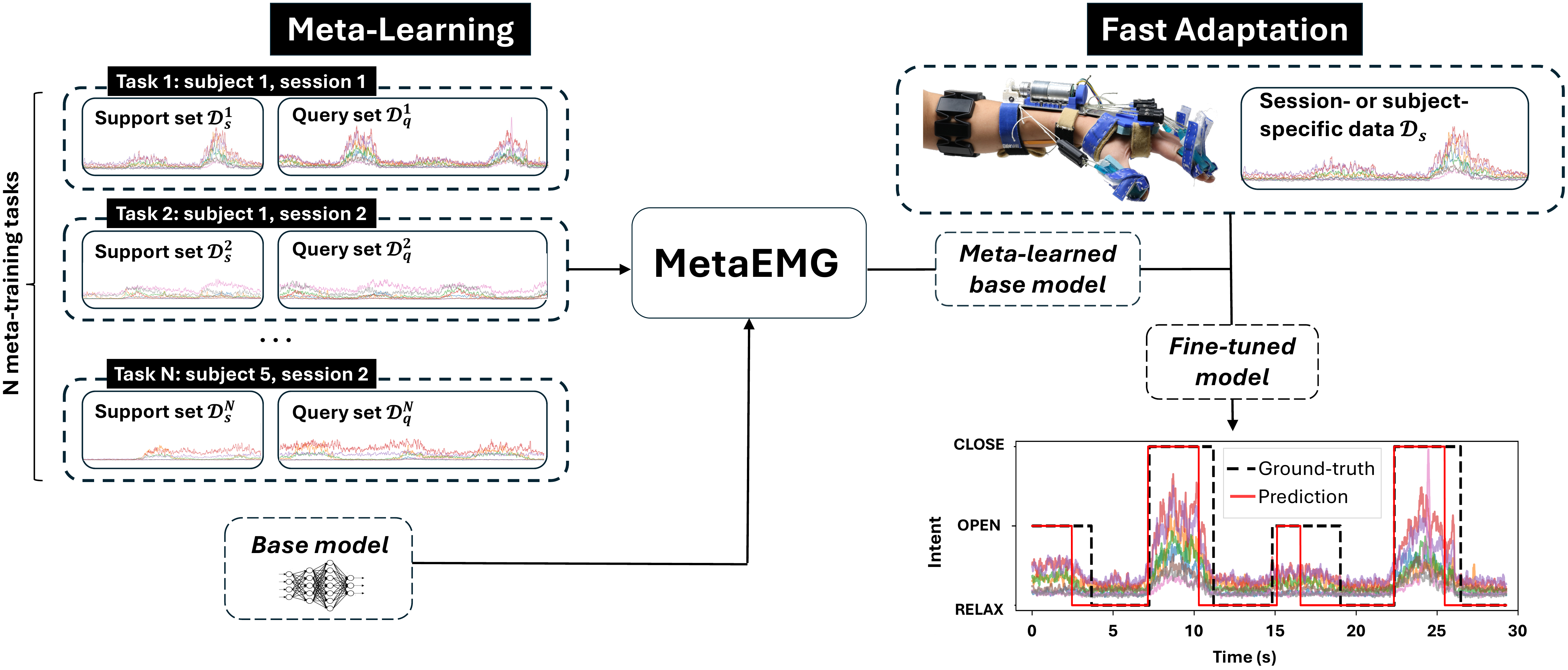}
    \caption{MetaEMG for EMG intent inferral. Our method uses previously collected data from our orthosis to meta-learn models that adapt to new sessions or subjects more quickly and with less data. }
    \label{fig:overview}
\end{figure*}

\subsection{Intent Inferral Formulation}

We formalize intent inferral as the forecasting problem of predicting the intended user movement (in our case, hand opening, closing, or relaxing) from a stream of biosignals collected by a wearable robot (in our case, EMG data collected by an armband). Specifically, our goal is to determine intent from a data vector $x_t$ where at every time $t$, $x_t$ stores a 2-second window of EMG data $e^i$ across all 8 channels of our sensor:
$$x_t=[e^1_t, e^2_t...e^8_t]^T$$
Here, we specifically aim to predict which movement out of $\{open,close,relax\} $ the subject intends to perform with their hand, and we achieve this by calculating a vector of intent probabilities:
$$\hat{y_t}=[p^R_t, p^O_t, p^C_t]^T$$
where $p^R_t$, $p^O_t$, and $p^C_t$ represent the probabilities that the user’s intent is to relax (i.e. maintain current posture), open and close their hand, respectively.

We consider a base intent inferral model as a neural network parameterized by parameters $\theta$ whose output is used to derive the three intent probabilities, or $p^R_t$, $p^O_t$, and $p^C_t$. Traditionally, such a model would be trained on a dataset of labeled data collected by the experimenter at the beginning of each patient session. However, our goal is to reduce the burden of data collection and training for every session of use, so that we can more easily onboard new subjects and new therapeutic activities with our device. 

\subsection{Intent Inferral as a Meta-Learning Problem}

Meta-learning is characterized by the use of meta-training and meta-testing tasks to guide the learning process toward representations best suited for fast adaptation. Meta-training tasks refer to the tasks that are used in the meta-learning procedure to find a good base model initialization, and meta-testing tasks are unseen holdout tasks to evaluate the adaptation ability of the meta-learned base model. A task is defined by a tuple $\mathcal{T}=(\mathcal{D}_s,\mathcal{D}_q) $ composed of a support set $\mathcal{D}_s$ and a query set  $\mathcal{D}_q$, where $\mathcal{D}_s$ is much smaller in size than  $\mathcal{D}_q$ and contains only a small number of labeled data points. The objective of meta-learning in the context of supervised learning is to minimize the cross-entropy loss $\mathcal{L}$ of the meta-learned base model on the query sets of the meta-testing tasks.

In EMG intent inferral, we define a task as a continuous uninterrupted recording of EMG sensor data. For each recording, we instruct the subjects to open and close their hands three times by giving verbal cues of open, close, and relax. Each verbal cue lasts for 5 seconds and there is a relax cue between each open and close cue. The recording contains the EMG signals and their corresponding cues as the ground truth intent labels. We define each completion of the opening and closing hand as one \textit{open-relax-close motion}. 
We define the support set $\mathcal{D}_s$ to be the EMG signals of the first motion. In contrast, the query set $\mathcal{D}_q$ contains the second and third rounds of open-relax-close motions. In other words, each EMG task is given by $\mathcal{T}=(\mathcal{D}_s,\mathcal{D}_q)$ such that $\mathcal{D}_s=\{(x_t,y_t)\}_{t={1}}^{k}$ and $\mathcal{D}_q=\{(x_t,y_t)\}_{t={k+1}}^{n}$, where $k$ denotes the end of the first open-relax-close motion and $n$ denotes the end of the entire recording. An example of our intent inferral task is shown in Fig.~\ref{fig:task}.

With this task structure, the meta-objective can be interpreted as follows. We seek to train a base model using the complete support set $\mathcal{D}_s$ and query set $\mathcal{D}_q$ over a number of meta-training tasks. Then, when presented with a new meta-testing task (e.g., an unseen task), this model should be quickly fine-tuned using only the support set $\mathcal{D}_s$ of the new task, meaning just a single round of open-relax-close motion. 

\subsection{MetaEMG}

Under our task formulation, the proposed MetaEMG is an application of the Model-Agnostic Meta-Learning (MAML)~\cite{finn2017model} algorithm to EMG intent inferral. MAML is a meta learning algorithm that focuses on developing models with enhanced adaptability. Rather than optimizing a model for a specific task, MAML aims to create a model that can quickly adapt to new tasks with minimal additional training essentially training the model to be an efficient learner. By doing so, this algorithm can produce models that perform well across a diverse range of tasks, even when provided with only a small number of examples for each new task~\cite{finn2017model}.

MetaEMG aims to find the optimal initial model parameters for adapting a base model to a new session or subject from a limited dataset in only a few fine-tuning epochs. In MetaEMG, two nested optimization loops are used to fine-tune a base model for different tasks using small amounts of data, as shown in Fig.~\ref{fig:overview} and Alg.~\ref{alg:algo1}.


\begin{algorithm}
\caption{MAML for EMG intent inferral}\label{alg:algo1}
\KwIn{$T$: collection of $N$ tasks $\tau_i$, where each $\tau_i$ is a single session recording}
\KwIn{$\alpha, \beta$: learning rates for inner and outer optimization loops}
\KwIn{$M$, $K$: epochs for inner and outer optimization loops}
Initialize base model parameters $\theta$ randomly\;
$k, m \gets 0$\;
\While{$k \leq K$}{ \tcp{outer loop}
    \For{each task $\tau_i$}{
        $\hat{\theta_i} \gets \theta$\;
        Extract support and query sets $\mathcal{D}_s^i$ and $\mathcal{D}_q^i$\;
        \While{$m \leq M$}{ \tcp{inner loop}
            $\hat{\theta_i} \gets \hat{\theta_i} - \alpha\nabla_{\hat{\theta_i}}\mathcal{L}_{\mathcal{D}_s^i}(f_{\hat{\theta_i}})$\;
        }
        Evaluate query loss $\mathcal{L}_{\mathcal{D}_q^i}(f_{\hat{\theta_i}})$\;
        Evaluate and store gradients $\nabla_{\theta}\mathcal{L}_{\mathcal{D}_q^i}(f_{\hat{\theta_i}})$\;
    }
    Update $\theta \gets \theta - \beta\nabla_{\theta}\sum_{i=0}^N \mathcal{L}_{\mathcal{D}_q^i}(f_{\hat{\theta_i}})$\;
}
\end{algorithm}

One of the advantages of formulating intent inferral in the meta learning framework is that it allows us to train models in a way that directly mirrors their use case. In the inner loop of MetaEMG, a base model is fine-tuned on the support set $\mathcal{D}_s$ of a training task,
and that base model's ability to adapt to the task is measured through the query loss $\mathcal{L}_{\mathcal{D}_q}$ of that task (Alg.~\ref{alg:algo1}, steps 7 - 10). The query loss is simply the cross-entropy loss for the query set of the task computed between the predictions and ground-truth labels. This process simulates the real-life use case of fine-tuning an intent classifier on a small amount of demonstration data and using that model for real-time inference in a session with the device. 

Moreover, the base model update is performed once per iteration of the outer optimization loop. The query loss is used to accumulate the gradients on the base model parameters across all of the training tasks until a backward step is performed once all of those training tasks have been seen (Alg.~\ref{alg:algo1}, steps 11 - 13). 

Importantly, the fine-tuning in the inner loop is performed via gradient descent on $\mathcal{L}_{\mathcal{D}_s}$ (taken with respect to the latest parameter set $\hat{\theta_i}$), while the base model update of the outer loop is performed via gradient descent on $\mathcal{L}_{\mathcal{D}_q}$ (taken with respect to the base model parameter set $\theta$) which requires backpropagation through the inner optimization and thus higher-order gradients. By updating the base model using $\mathcal{L}_{\mathcal{D}_q}$, the training process favors base model updates that allow it to on average generalize to both the support $\mathcal{D}_s$ and query $\mathcal{D}_s$ sets of the training tasks despite only being fine-tuned on the support sets. For EMG intent inferral, we hypothesize that this training structure can produce intent classifiers that are less likely to overfit when fine-tuned to small amounts of demonstration data.

Models are trained on a collection of meta-training tasks and then evaluated on a collection of meta-testing tasks, and the specific allotment of session recordings in each collection depends on the individual use case for the model that we describe in the experimental setup. Generally, larger diverse meta-training task collections are favored, and in this paper, we select recordings to simulate realistic use cases to evaluate our meta-learning framework's fast adaptation ability. 

\section{Data Collection and Session Conditions}
\label{sec:data}


All data used in this study was collected using MyHand, a robotic exotendon device comprised of a system of motors and linkages which assist the user in opening and closing their hand, as shown in Fig.~\ref{fig:teaser_metaemg}. The device rests on the user's forearm and through a series of tendons connected to the user's fingertips assists the hand in opening and closing a grasp. EMG sensor data is recorded from an 8-electrode armband (Myo, Thalmic Labs). Once a command is issued to either open or close the hand (for example, by an intent inferral model as the one that is the focus of this study), the orthosis engages the motor and either retracts or extends the exotendons on the dorsal side of the finger. Finger extension is actively assisted by the device through tendon retraction, while finger flexion is allowed unimpeded via the subject's own strength. Additional hardware details can be found in our previous studies~\cite{park2018multimodal,chen2022thumb}. 

 
\subsection{Data}

Critically, the fact that we collect our data using a complete, functional assistive device means that our collection protocol can span the conditions that are meaningful for real-life operations. Our previous work~\cite{meeker2017emg,xu2022adaptive} has shown that, in our patient population, EMG signals display significant differences when collected in isolation (i.e. using just an armband and no other hardware) vs. when collected in conjunction with an active orthosis assisting finger movement. Thus, in our method, we collect data in four different experimental conditions as follows:
\begin{enumerate}
    \item \textbf{Arm on table, motor off}: the assisted arm of the subject rests on the table, and the motors of the MyHand are disengaged, providing no physical assistance to the subject.
    \item \textbf{Arm on table, motor on}: the assisted arm of the subject rests on the table, and the motors of the MyHand are engaged, allowing the device to provide active grasp assistance to the user.
    \item \textbf{Arm off table, motor off}: the subject keeps their arm raised to shoulder level for the duration of the session, and the motors are disengaged, providing no physical assistance to the subject.
    \item \textbf{Arm off table, motor on}: the subject keeps their arm raised to shoulder level for the duration of the session, and the motors of the MyHand are engaged, allowing the device to provide active grasp assistance to the user.
\end{enumerate}

\subsection{Pre-processing}

The raw output of the EMG sensor is an 8-channel time-series signal (shown in Fig.~\ref{fig:teaser_metaemg}) sampled at 100 Hz which is pre-processed before going into our intent inferral model. The labels attached to each sample are the verbal cues given to the subject during the session, and in a 2-second time interval, there are 200, 8-channel samples. During pre-processing, the 8-channel EMG signal is clipped to the range $[0,1000]$ to discard any outliers, and each channel is re-scaled to the range $[-1,1]$. Finally, the scaled 8-channel signal is binned into 2 second-long 8-channel data windows, following a sliding window stride length of 10ms. While muscle activation is known to occur in a much smaller time window, we hypothesize that 2 seconds encapsulates the total reaction time from interpreting the verbal command to actual movement exertion. 

\section{Experimental Setup}

\begin{table}
\centering
 \begin{tabular}{c c c c} 
 \toprule
 Subject ID & Age & Gender & FM-UE \\
 \midrule
 S1 & 46 & Male & 27 \\ 
 S2 & 48 & Male & 26 \\
 S3 & 53 & Female & 26 \\
 S4 & 31 & Male & 50 \\
 S5 & 53 & Male & 47 \\ 
 \bottomrule
 \end{tabular}
 \caption{Subject information.}
 \label{tab:subjects}
\end{table}

We evaluate MetaEMG in a series of experiments conducted on data collected using the MyHand orthosis and involving five chronic stroke survivors of moderate muscle tone (MAS $\le 2$) and varying degrees of arm impairment (FM-UE scores listed in Table~\ref{tab:subjects}) in sessions approved by the Columbia University Institutional Review Board (IRB-AAAS8104). Subjects are generally able to actively close their hands but not open them, which is one of the motivating factors in both the robotic orthosis design as well as the intent inferral structure.

In each single session, the subjects are asked to follow verbal cues asking them to either relax, close, or open their hand. Each cue lasts a duration of five seconds, and this procedure is followed across the four experimental conditions outlined in Sec.~\ref{sec:data}. Each recording contains three rounds of open-relax-close motions. In total, 14 recordings were gathered for each subject over the course of two separate days, where 8 of those recordings came from the first day of data collection and 6 came from the second day. In order to study intersession concept drift, the two days are at least a week apart, providing enough time for the drift caused by the chronic change in muscle condition to happen. For consistency with our method's terminology, we refer to individual recordings as tasks such that the first open-relax-close motion of that recording is encapsulated in its support set $\mathcal{D}_s$ and the remainder of the recording in the query set $\mathcal{D}_q$.

\subsection{Assessment Scenarios}

All of the experiments are conducted using this same dataset of five subjects. In each experiment, models are first pretrained on the meta-learning task collection and then evaluated on the meta-testing tasks. For both meta-learning and meta-testing tasks, $\mathcal{D}_s$ is always used to fine-tune a base model to a task. The distinction is that during the pretraining phase, the fine-tuned model's prediction error on $\mathcal{D}_q$ is used to guide the optimization of the base model, whereas in the evaluation phase, the classification accuracy on the task's $\mathcal{D}_q$ is used as our primary metric of model performance. We evaluate the performance of our method under two assessment scenarios as follows. 

\begin{itemize}
    \item \textbf{Session adaptation.} This scenario pertains to intersession concept drift, and it seeks to simulate using the orthosis on a subject seen previously in a different session on a different day. A session is defined as a single-use between donning and doffing the device. We run a single experiment, and our meta-learning and meta-training tasks are simply grouped by day. Specifically, the meta-learning task collection contains recordings of all five subjects across all 4 data collection conditions, but it only contains those recordings collected on the first day of data collection. The meta-testing task collection is comprised of the recordings from only the second day of data collection, but it also includes all five subjects and all data collection conditions.
    \item \textbf{Subject adaptation.} This scenario seeks to address the challenge of onboarding new subjects onto our robotic orthosis. To simulate this scenario, we conduct five separate experiments, each one simulating the onboarding of one held-out subject given that we have seen the other four. In each experiment, all 14 recordings of the held-out subject are used as the meta-testing tasks, and the meta-training tasks are comprised of all of the session recordings from the other four subjects.
\end{itemize}


\subsection{Model Architecture}


The base architecture for our models is a simple 3-layer fully connected neural network of 512, 128, and 3 neurons at each layer, respectively. During inference, the final layer output of our network is passed through a softmax activation function to obtain a vector of intent probabilities. Our models and datasets are developed in PyTorch and trained on an NVIDIA GeForce RTX 3090 GPU for 50 epochs on a learning rate scheduler, reducing the outer learning rate by 0.9 every 10 epochs. We use Adam as the optimizer for both optimization loops, and we find the combination of 0.0005 outer learning rate and 0.0001 inner learning rate to yield the most stable training.

\subsection{Baselines}
In order to evaluate MetaEMG as an adaptation method, we compare it to several baselines without pretraining or with vanilla pretraining. The fast adaptability of MetaEMG allows it to be used as a base model for continuous and lifelong online learning with higher-capacity models. We are mostly interested in the performance comparison under limited training epochs as our main goal is to achieve fast adaptation to a new subject or session. However, we also include baselines that are trained to convergence, which normally takes much longer training time. These converged baselines are not suitable to be used as base models for online updates but we still include them for the completeness of results. 

All of our baselines have the same base intent inferral model neural network architecture, and the evaluation metric used across the board is the average classification accuracy on the $\mathcal{D}_q$ of the meta-testing tasks. The methods that we evaluate are as follows.

\subsubsection{No-pretraining (3 epochs)} This baseline does not pretrain on any of the offline corpus of data from the meta-training tasks. We initialize the weights of the model randomly, train it directly on the support set $\mathcal{D}_s$ of the meta-testing tasks for 3 epochs, and evaluate its classification accuracy on the query set $\mathcal{D}_q$ of those same tasks. This helps us distinguish whether the pretraining process (meta-learning or vanilla pretraining) is indeed helping our models learn more effectively or if the training on the limited one open-close-relax motion from the new subject or session suffices to produce a good classifier.

\subsubsection{No-pretraining (converged)} This is the same as the previous baseline, except it is trained to convergence during the fine-tuning process on the support sets of meta-testing tasks, which takes 50 epochs. 


\subsubsection{Conventional-pretraining (3 epochs)} This baseline aggregates both the support sets and the query sets of all meta-training tasks into a single dataset and treats them all as a single support set. The model is trained on the entirety of this lumped dataset, and it is fine-tuned to the support sets of the meta-testing tasks. 

\subsubsection{Conventional-pretraining (converged)} This is the same as the previous baseline, but it is fine-tuned to convergence on the support sets of the meta-testing tasks for 3 epochs. 

\subsubsection{MetaEMG} This is our proposed method, only fine-tuned for 3 epochs on the meta-testing tasks. We note that 3 epochs is convergence for models trained via MetaEMG.


\section{Results \& Discussion}
\begin{table*}
\setlength{\tabcolsep}{2pt}
\footnotesize
\renewcommand{\arraystretch}{0.6}
\centering
    \begin{tabular}[t]{c|ccccc|c}
        \toprule
         \textbf{Method} & \textbf{S1} & \textbf{S2} & \textbf{S3} & \textbf{S4} & \textbf{S5} & \textbf{Average}\\
        \midrule
        No-pretraining (3 epochs) & 52.9 $\pm$ 1.2\% & 41.1 $\pm$ 5.5\% & 53.1 $\pm$ 8.5\% & 55.8 $\pm$ 0.9\% & 57.1 $\pm$ 2.5\% & 52.06\%\\
        No-pretraining (converged) & 67.5 $\pm$ 0.6\% & 62.2 $\pm$ 0.7\%	& 73.7 $\pm$ 1.7\% & \textbf{79.6 $\pm$ 0.2\%} & 66.0 $\pm$ 2.2\% & 69.85\%\\
        Conventional-pretraining (3 epochs) & 70.2 $\pm$ 2.8\% & 63.3 $\pm$ 2.6\% & 76.3 $\pm$ 2.9\% & 71.6 $\pm$ 1.3\% & 71.3 $\pm$ 2.4\% & 70.59\%\\
        Conventional-pretraining (converged) & 70.3 $\pm$ 2.0\% & \textbf{64.7 $\pm$ 1.6\%} & 79.5 $\pm$ 2.1\% & 75.8 $\pm$ 0.8\% & 70.4 $\pm$ 2.9\% & 72.19\%\\
        \textbf{MetaEMG} & \textbf{74.5 $\pm$ 2.5\%} & 63.6 $\pm$ 0.9\% & \textbf{80.5 $\pm$ 1.0\%} & 78.6 $\pm$ 1.3\% & \textbf{74.4 $\pm$ 0.3\%} & \textbf{74.37\%} \\
        \bottomrule
    \end{tabular}
    \caption{Session adaptation results. Classification accuracy and standard deviation across 5 stroke subjects. Results are averaged across 3 separate randomly generated training seeds. The best results for each subject are reported in bold.}
    \label{tab:session}
\end{table*}

\begin{table*}
\setlength{\tabcolsep}{2pt}
\footnotesize
\renewcommand{\arraystretch}{0.6}
\centering
    \begin{tabular}[t]{c|ccccc|c}
        \toprule
         \textbf{Method} & \textbf{S1} & \textbf{S2} & \textbf{S3} & \textbf{S4} & \textbf{S5} & \textbf{Average}\\
        \midrule
        No-pretraining (3 epochs) & 54.6 $\pm$ 1.5\% & 55.3 $\pm$ 1.6\% & 51.6 $\pm$ 5.1\% & 54.8 $\pm$ 2.9\% & 53.9 $\pm$ 1.0\% & 54.1\%\\
        No-pretraining (converged) & 60.1 $\pm$ 0.3\% & 66.8 $\pm$ 1.3\% & 68.3 $\pm$ 0.7\% & 74.7 $\pm$ 0.5\% & 69.4 $\pm$ 0.4\% & 67.9\%\\
        Conventional-pretraining (3 epochs) & 60.9 $\pm$ 0.9\% & 68.8 $\pm$ 1.0\% & 73.7 $\pm$ 0.7\% & 75.4 $\pm$ 1.7\% & 64.7 $\pm$ 1.4\% & 68.7\%\\
        Conventional-pretraining (converged) & 62.9 $\pm$ 0.7\% & \textbf{71.3 $\pm$ 0.4\%} & \textbf{74.2 $\pm$ 0.1\%} & 78.3 $\pm$ 0.6\% & 66.6 $\pm$ 1.3\% & 70.6\%\\
       \textbf{MetaEMG} & \textbf{63.8 $\pm$ 0.8\%} & 69.0 $\pm$ 1.1\% & 74.1 $\pm$ 0.5\% & \textbf{80.5 $\pm$ 1.1\%} & \textbf{70.0 $\pm$ 0.5\%} & \textbf{71.5\%}\\
        \bottomrule
    \end{tabular}
    \caption{Subject adaptation results. Classification accuracy and standard deviation across 5 stroke subjects. Results are averaged across 3 separate randomly generated training seeds. The best results for each subject are reported in bold.}
    \label{tab:subject}
\end{table*}

\begin{figure}
    \centering
    \footnotesize
    \includegraphics[width=0.65\linewidth]{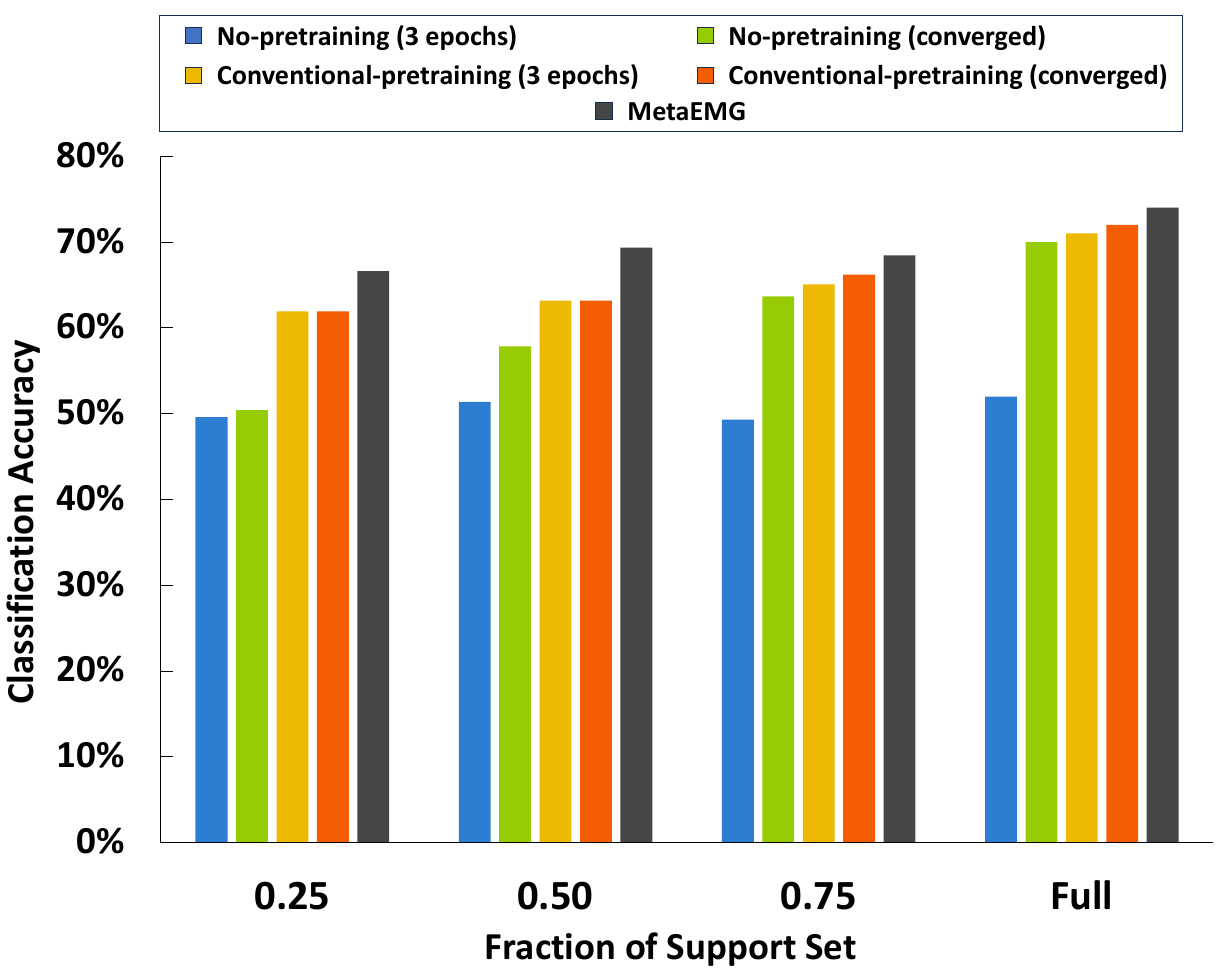}
    \caption{Classification accuracy with fewer session-specific data. In the session adaptation experiments, we further reduce the amount of session-specific fine-tuning data to 0.75, 0.5, and 0.25 of the original amount. }
    \label{fig:fewer_data}
\end{figure}

\begin{figure}
    \centering
    \includegraphics[width=0.65\linewidth]{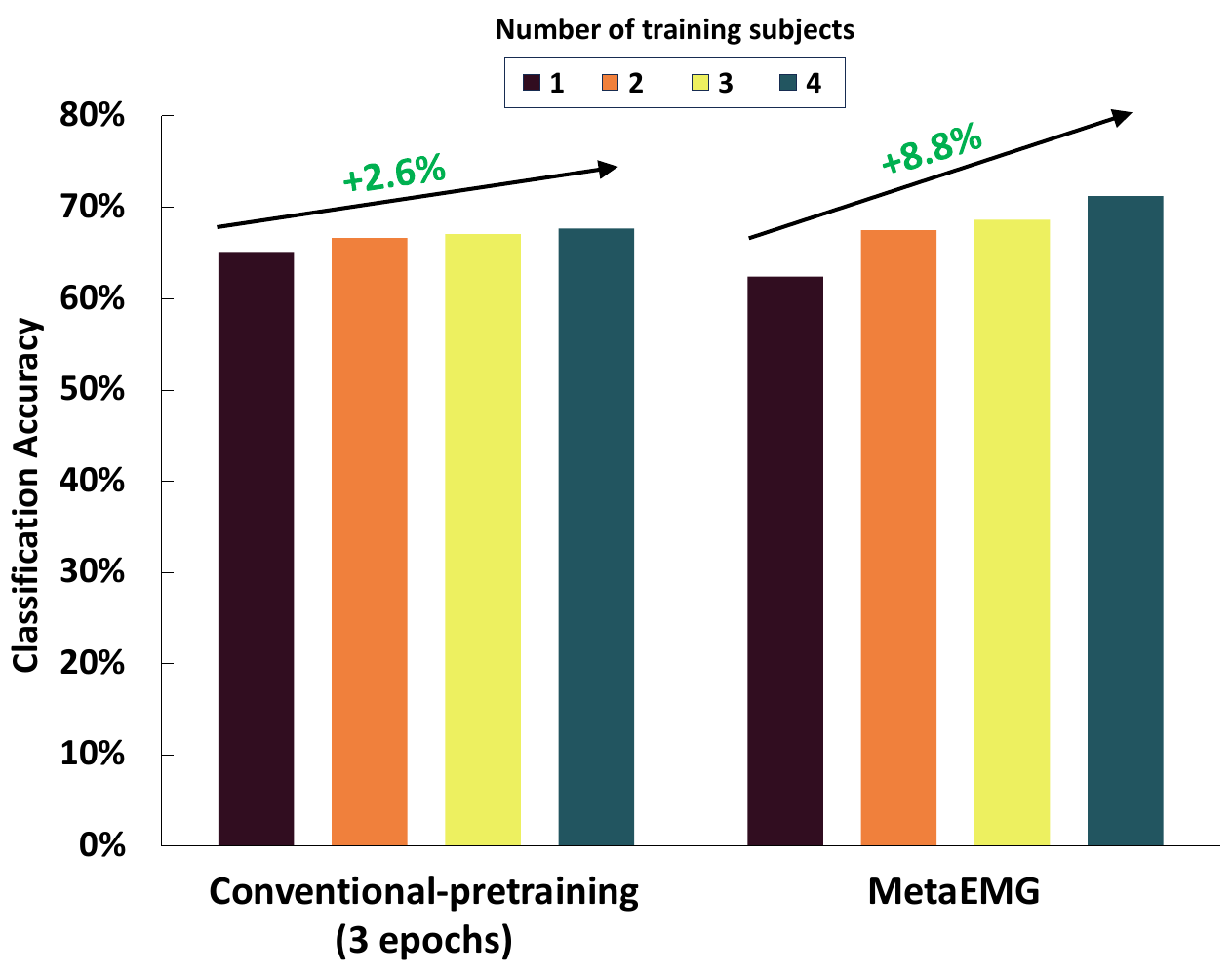}
    \caption{Classification accuracy with different number of pretraining subjects. Models are pretrained on 1, 2, 3, and 4 subjects before being fine-tuned on a subject not seen in pretraining. For each number of pretraining subjects, we run experiments with all possible partitions of the meta-training and meta-testing subjects and report the average classification accuracy.}
    \label{fig:fewer_subjects}
\end{figure}    

Through the experimental results from two realistic adaptation scenarios, we aim to demonstrate the fast adaptation ability of MetaEMG from two perspectives: (1) MetaEMG adapts better with very limited subject- or session-specific data, and (2) MetaEMG adapts better with only a few fine-tuning epochs.

Both perspectives are critical in intent inferral with EMG signals for stroke. Fast adaptation with less data relieves the burden of data collection for a new subject or session. Fast adaptation with fewer training epochs reduces the overhead time for using the device, and more importantly, it has the potential to be used as a base model for on-device online learning, which requires fast continuous model updates.

\subsection{Session Adaptation}

MetaEMG outperforms all baselines on the average intent inferral accuracy in the session adaptation experiments, as shown in Table~\ref{tab:session}. We simulate the scenario of the same subject returning for another use session of our device after at least a week. Only one open-relax-close motion of EMG signals is provided as the support set for the meta-testing tasks. 

Without pretraining on any of the offline corpus of data from previous subjects or sessions, \textit{No-pretraining} baselines perform worse than \textit{Conventional-pretraining} baselines. This shows that our high-capacity neural network classifier benefits from pretraining on a larger corpus of offline EMG data. However, \textit{Conventional-pretraining} baselines still perform worse than MetaEMG. This is due to the intersession concept drift after donning and doffing the device, which is caused by chronic hand function changes and device position discrepancies on the forearm. Through meta-learning, MetaEMG adapts more efficiently to these drifts in EMG signals, achieving a higher classification accuracy (74.37\%) compared to \textit{Conventional-pretraining (3 epochs)} (70.59\%).

We further investigate if MetaEMG adapts better with even smaller fine-tuning datasets. We downsample the support set of each meta-testing task to 25\%, 50\%, and 75\%, and we compare the average classification accuracies across all subjects (shown in Fig.~\ref{fig:fewer_data}). We find that MetaEMG is most resilient to the support set reduction, maintaining a classification accuracy above $66\%$, while other baselines' performance deteriorates more dramatically. This suggests that meta-learned models are potentially also better at learning from smaller session-specific datasets.

Apart from adapting better with limited session-specific data, another major advantage of MetaEMG is that it can adapt effectively with very few epochs. MetaEMG shows a large improvement in classification accuracy over the baselines when limited to only 3 epochs of fine-tuning for the meta-testing tasks. While at convergence, the difference between the baselines and MetaEMG is small. We note that on our hardware, it takes 50 epochs for both baselines to converge when fine-tuning on the session-specific data of the meta-testing tasks. The training time for convergence of the baselines is a magnitude longer than training MetaEMG for 3 epochs, yet MetaEMG still outperforms the converged baselines. Being able to adapt fast with a few epochs is beneficial for two reasons. (1) It is essential for continuous online updates, and the meta-learned base model has the potential to enable on-device life-long learning. (2) It enables the use of higher capacity models where each training epoch can take a significant amount of time. 


\subsection{Subject Adaptation}

Table~\ref{tab:subject} shows the subject adaptation results and MetaEMG is on average better than all of the baselines, suggesting that meta-learning yields benefits in subject adaptation over the other methods. Subject adaptation is generally a harder problem compared to session adaptation because the distributional shift across subjects is larger than that across different sessions from the same subject. This is shown by the performance drop of all methods from the session adaptation experiments except for \textit{No-pretraining (3 epochs)}. However, meta-learning is still able to produce a base model that adapts better with limited subject-specific data and very few epochs by transferring knowledge across similar subjects. 



In order to further understand if different pretraining methods would improve as we add more subjects and data diversity to our database, we conduct another ablation experiment where we vary the number of pretraining subjects presented in the meta-training task collection, as shown in Fig.~\ref{fig:fewer_subjects}. We compare \textit{Conventional-pretraining (3 epochs)} and MetaEMG with only 1, 2, 3, and 4 subjects in our meta-training tasks and with the rest of the subjects in the meta-testing tasks. MetaEMG shows a more significant improvement in intent inferral accuracy when the number of pretraining subjects increases, and we believe this indicates that with more subjects added to our corpus of offline data, the advantage of MetaEMG will become even more pronounced. 


\section{Chapter Summary}

Different from SemiEMG as discussed in the last chapter, which utilizes unlabeled data, MetaEMG treats each session/subject as a single task, enabling fast adaptation to a new session/subject through meta-learning. This naturally and effectively handles intersession drift, which is a limitation of SemiEMG. MetaEMG further reduces the training burden on the stroke patients, and mediates the challenge of data scarcity of intent inferral. To our knowledge, this is the first time meta-learning has been explored as a way to improve intent inferral in stroke survivors, and we hope that it inspires other health domains where large-scale data collection is difficult. 

For all the learning paradigms discussed so far, the learning happens unidirectionally, meaning the classifier is learning from the subject. In the next chapter, we will discuss a new paradigm that enables bi-directional learning.

%% file: chapters/reciprocal.tex
\chapter{Reciprocal Learning for Bi-directional Adaptation Between Patients and Device}
\label{chap:reciprocal_learning}

In this section, we propose an entirely new paradigm, dubbed reciprocal learning. Different from works discussed before, reciprocal learning enables bi-drectional learning between patients and devices. 

\section{Motivation}

Up till now, we have explored multiple learning paradigms to address the challenge of data scarcity in rehabilitation robots, specifically, intent inferral to control a hand orthosis for stroke. However, for all these learning paradigms, the learning is unidirectional, meaning that the models learn from the subjects through collected biosignals. In this reciprocal learning work, we aim to bridge the gap and enable the opposite learning direction. We achieve this through augmented visual feedback added on top of our device, providing intuition on the states of the classifier, such that the subjects can adapt to the behaviors of the models, and then generate higher-quality data. In this way, we effectively handle the challenge of data scarcity through bi-directional adaptation and generating more distinguishable data samples. 

\begin{figure}[t]
    \centering
    \includegraphics[width=0.65\textwidth]{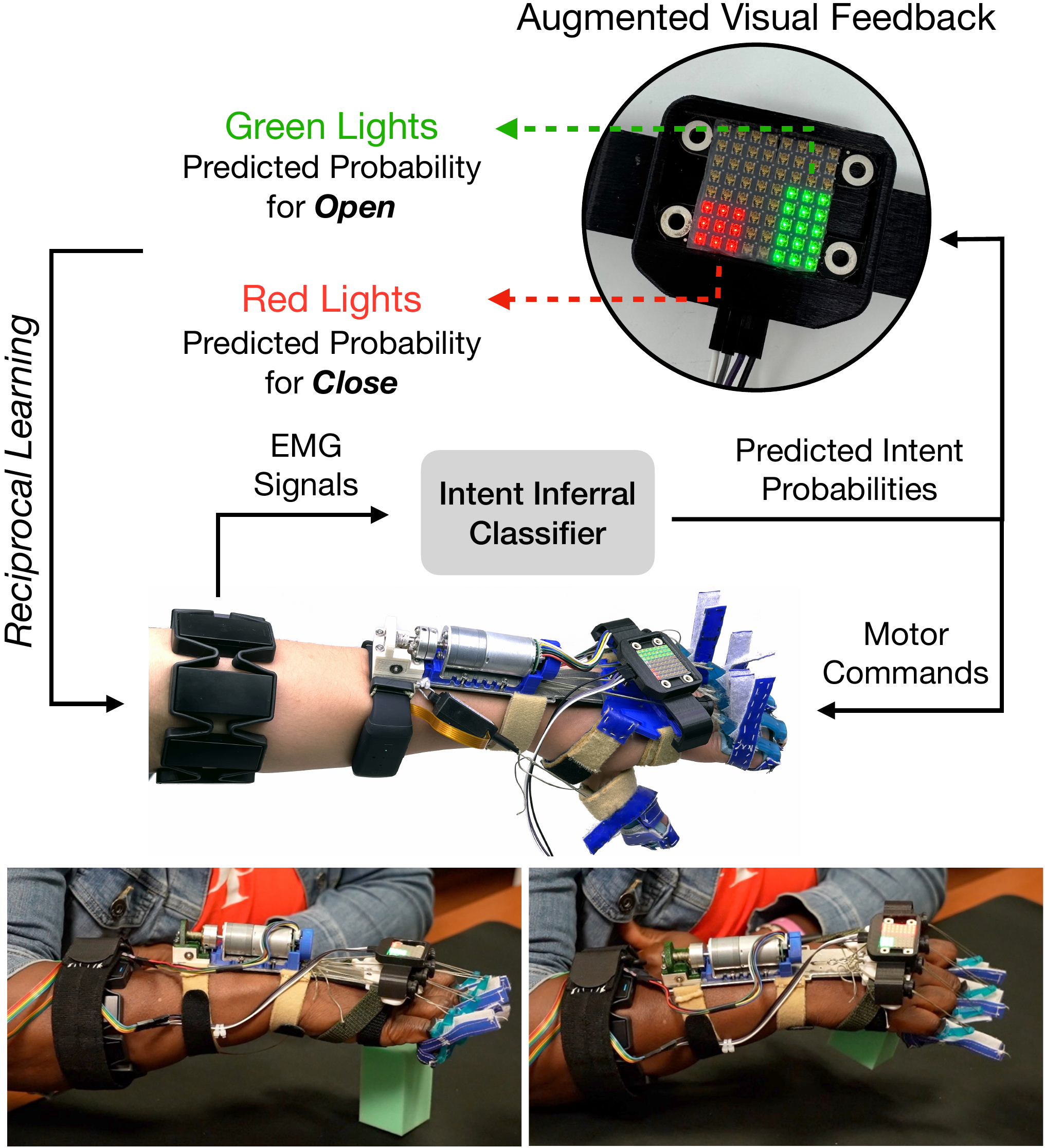}
    \caption{Reciprocal learning with robotic hand orthosis and augmented visual feedback. Our hand orthosis assists stroke survivors in opening the hand when an \textit{open} intent is detected from EMG signals, and allows the hand to close when it detects the \textit{close} intent. Augmented feedback consists of LED display bars corresponding to the predicted probabilities of \textit{open} and \textit{close} from the intent classifier running on the orthosis. Subjects use visual feedback to adapt to the intent inferral model during reciprocal learning practice. The bottom row shows a stroke survivor using our device with augmented feedback for a functional pick-and-place task.
    }
    \label{fig:teaser}
\end{figure}

Typical intent inferral methods take in user intents as unidirectional ground truths, but wearable interfaces are inherently bidirectional systems. Humans modify their control inputs in response to sensory feedback of device behavior, such as whether the device moves according to expectations~\cite{Proulx2022interaction}. Device developers have sought to provide augmented feedback, or feedback in addition to intrinsic human sensing, to communicate information about biosignals that are not directly observable in order to aid human adaptation. Augmented feedback can compensate for reduced proprioception associated with sensory impairments after stroke~\cite{Ingemanson2019somatosensory}, and when applied to EMG signals can offer insights into muscle coactivations~\cite{Wright2014mci}. Multiple groups~\cite{Madduri2023coadaptation, Hu2023coadaptation} have suggested augmented feedback of biosignals as a potential modality to mediate co-adaptive learning and enable personalized control, such as human and device simultaneously adjusting control parameters in response to the other agent's behavior. However, co-adaptation in practical deployments such as physically assistive devices or rehabilitative interventions has yet to be realized for non-able-bodied users.

Reciprocal learning uses bidirectional intent inferral on a robotic device to dynamically encourage greater separability of control input patterns by stroke-impaired users, instead of solely relying on static thresholding of physiological or kinematic signals~\cite{cisnal2023interaction}. Furthermore, we can immediately deploy this approach for robot-assisted movement. We treat the human as a dynamic co-learner alongside the classifier algorithm, prompting each to update their understanding of the other's behavior. The combination of an adaptive classification algorithm and augmented feedback can increase the coherence of sensory input and action output, which in turn helps users develop greater automaticity when controlling the hand. In reciprocal learning, the user learns to generate better biosignals for ML models while practicing actively using the hand. In summary, our main contributions are as follows:

\begin{itemize}
    \item We introduce a novel paradigm for training EMG-based intent inferral, consisting of interwoven sessions that alternate between updating ML models based on human biosignal generation and encouraging human exploration and adaptation to models with the use of augmented feedback.
    \item As far as we know, this bidirectional training system is a completely unexplored area in the context of upper-limb rehabilitation after stroke. We are the first to use augmented feedback to communicate the inner state of the learning model running on a wearable robotic orthosis as it assists hand movement.
    \item Our results demonstrate that reciprocal learning leads to better predictions for intent inferral on a subset of stroke subjects by aiding in the generation of more separable muscle activation patterns and biosignals. 
\end{itemize}

\section{Reciprocal Learning Paradigm}

\begin{figure*}[h]
    \centering
    \includegraphics[width=0.9\linewidth]{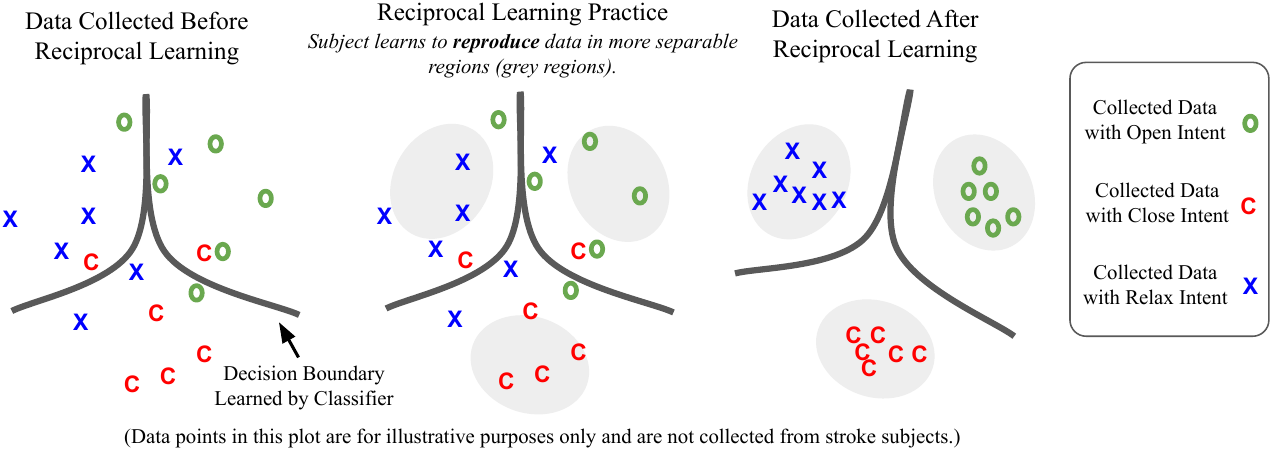}
    \caption{The goal of reciprocal learning is to guide the subject in generating more separable signals. Before reciprocal learning practice, the collected data are more diverse for each intent, making it difficult for the classifier to learn a perfect decision boundary. During reciprocal learning practice, the subject adapts to the intent inferral classifier and learns to generate data in more separable regions. After reciprocal learning practice, we retrain the classifier using the newly generated more separable data. This figure is intended to illustrate the concept, and does not show EMG signals collected from real subjects; separability data from real stroke subjects will be shown in Fig.~\ref{fig:separability}.}
    \label{fig:illustration}
\end{figure*}

\begin{figure*}[h]
    \centering
    \includegraphics[width=\textwidth]{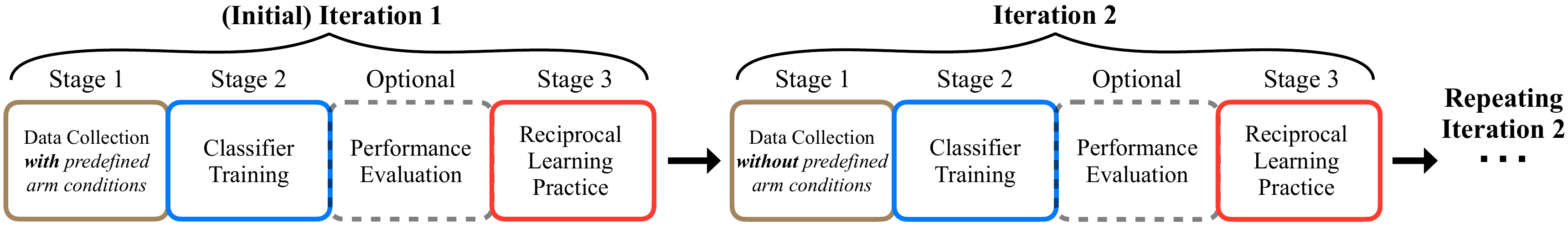}
    \caption{Reciprocal learning paradigm. An iteration consists of data collection, classifier training, performance evaluation, and reciprocal learning practice.}
    \label{fig:paradigm}
\end{figure*}

The ultimate goal of this work is to develop algorithms that can infer discrete intent (one of \{\textit{open}, \textit{relax}, \textit{close}\}) from EMG signals generated by stroke-impaired users attempting to move their hands, so that our robotic device can provide appropriate assistance. If the predicted intent is \textit{open}, the robot extends the user's fingers to open the hand; if the predicted intent is \textit{close}, the robot releases to allow the user to use their own strength to close the hand; if the predicted intent is \textit{relax}, the robot simply maintains the previous state. 

The classical way to train intent inferral algorithms is completely unidirectional. A researcher would issue cues to open, close, or relax the hand to a user wearing the device, then would record the real-time EMG signals and label the data using the provided cues as ground truth intent. In this unidirectional process, ML models learn personalized patterns through the labeled dataset and generate predictions. Although the user can observe the robot opening or closing, they do not have knowledge about the underlying ML predictions or EMG signals that led to the robot taking action.

Our approach aims for a different bidirectional paradigm, dubbed reciprocal learning. We hypothesize that by providing insights about the internal state of the model in an intuitive and self-explanatory way, a user can adapt to the behavior of the classifier and reproduce more distinguishable control inputs. This idea is illustrated conceptually in Fig.~\ref{fig:illustration}.


Reciprocal learning is an iterative process and can have a potentially unlimited number of iterations (shown in Fig.~\ref{fig:paradigm}). Each iteration consists of multiple stages, listed below: 

\begin{itemize}
    \item \textbf{Stage 1: Data Collection.} We collect a labeled dataset from a subject by providing verbal cues of \textit{open}, \textit{relax} and \textit{close} while simultaneously recording EMG signals.  
    \item \textbf{Stage 2: Classifier Training.} We train an intent inferral classifier using the collected dataset. This is where one direction of the learning happens---the ML model learns from the subject through the labeled dataset. 
    \item \textbf{Optional: Performance Evaluation.} The purpose of this optional stage is to use any metric of our choice to track the performance of the trained intent inferral classifiers across iterations. 
    \item \textbf{Stage 3: Reciprocal Learning Practice.} We run the learned intent inferral classifiers on the orthosis and provide the augmented visual feedback to communicate the inner state of the ML model that is not directly observable. The subject practices to generate distinguishable biosignals for \textit{open} and \textit{close} intents, using augmented feedback as guidance. This is where the opposite direction of learning happens---the subject explores and adapts to the ML model.
    \item \textbf{Repeat...} We continue to the next iteration, and collect a new batch of data from the subject who has supposedly learned to produce more distinguishable EMG signals.
\end{itemize}

\section{Methods and Implementation}

Having described our general concept and approach, we now expand on the method and implementation of each stage.

\subsection{Data Collection}
In the data collection stage of each iteration, we collect a labeled dataset by providing verbal cues for subjects to open, close, and relax the hand. We simultaneously record the EMG signals and the given verbal cues as ground truth labels. The EMG armband \mbox{(Myo, Thalmic Labs)} has eight electrodes encircling the forearm. It streams filtered and rectified 8-channel surface EMG signals at 50Hz. 

Note that the arm conditions under which the data is collected differ between the first iteration and the following iterations. A \textit{condition} is a particular combination of hand position and whether the orthosis is actively providing assistance. In the initial iteration, the orthosis has no existing classifiers to run and thus no augmented visual feedback (see Sec.~\ref{sec:feedback} for details on the feedback mechanism) to provide. The subject has not done any reciprocal learning practice and thus has not found any muscle activations that consistently generate distinguishable EMG signals. As a result, we collect data from a diverse set of predefined arm conditions in the hope that this set provides enough diversity of muscle activation patterns such that one of them or interpolation between them is easily reproducible and separable. 

The four different arm conditions of the initial iteration are as follows. (1) \textit{Arm on table, motor off}: the assisted arm of the subject rests on the table, and the motor of the orthosis is disengaged, providing no physical assistance to the subject. (2) \textit{Arm on table, motor on}: the assisted arm of the subject rests on the table, and the motor of the orthosis is engaged, allowing the device to provide active grasp assistance to the user. (3) \textit{Arm off table, motor off}: the subject keeps their arm raised to shoulder level for the duration of data collection, and the motor is disengaged, providing no physical assistance to the subject. (4) \textit{Arm off table, motor on}: the subject keeps their arm raised to shoulder level for the duration of data collection, and the motor of the orthosis is engaged, allowing the device to provide active hand-opening assistance.

For each condition, we collect two recordings, one for training and the other one held out for testing. A \textit{recording} is defined to be a continuous and uninterrupted recorded sequence of EMG signals. For each recording, we instruct the subject to open and close their hand 3 times. Each verbal cue lasts for 5 seconds, with a \textit{relax} cue between each \textit{open} and \textit{close} cue. As we have not yet trained a user orthosis controller, to collect data for active-motor conditions we manually command the orthosis to move approximately 1 second after the verbal cue. 

In the data collection stage of the following iterations, there are no more predefined arm conditions. The orthosis runs the trained classifier, providing both active assistance and augmented visual feedback at a 5 Hz operational rate. The subject's goal is to reproduce the muscle activation patterns learned in the reciprocal learning practice of the previous iteration. We collect four recordings: two for training and the other two for testing.

\subsection{Classifier Training}

In each classifier training stage, we train a Linear Discriminant Analysis (LDA) classifier. We choose LDA because it is fast, trackable, and widely used in biomedical research~\cite{meeker2017emg,xu2022adaptive,la2024meta,xu2024chatemg}. In the optional performance evaluation stage, we evaluate the performance of the classifiers both qualitatively and quantitatively (see Sec.~\ref{sec:exp} for details).

The classifiers take in a single time step of an 8-channel, 50Hz EMG signal. We clip the signal to $[0, 1000]$ and then normalize it to $[-1, 1]$. Prediction on a single step is noisy, so during online inference we apply a median filter of window size 20 to the predicted probabilities of \textit{relax}, \textit{open}, and \textit{close} over time. This median filter reduces unwanted oscillations in predictions when changing intents.

\subsection{Reciprocal Learning Practice}
\label{sec:feedback}

Our augmented visual feedback consists of two progress-bar displays mounted on the dorsal side of the hand orthosis so that the subject can observe them comfortably during reciprocal learning, as shown in Fig.~\ref{fig:teaser}. The relative heights of the green and red LED bars correspond to the model's predicted probabilities for \textit{open} and \textit{close} intents, respectively.

In the reciprocal learning practice stage, we ask the subject to practice opening and closing their hands guided by the augmented visual feedback. We run the trained classifier on the device to provide appropriate assistance. In this stage, we instruct the subject to trigger different device states (\textit{open} and \textit{close}) by maximizing the respective green or red LED bars. 

This is the exploration stage, where the subject adapts to the behavior of the device, and tries to discover and reinforce the muscle activation pattern that is easily reproducible and distinguishable. In this stage, the subject uses augmented visual feedback to learn to generate more separable EMG signals for each intent. This idea is conceptually illustrated in Fig.~\ref{fig:illustration}. Reciprocal learning practice essentially helps the subject locate the EMG subspace that is both reproducible and separable (i.e., far from the \textit{decision boundary} learned by the classifier of the previous iteration). Then, in the next iteration, the subject can reproduce data in these more separable regions, leading to new retrained classifiers that can more accurately predict user intent.

\section{Experimental Setup}

We tested reciprocal learning in single-session experiments with five stroke subjects, and compared intent inferral accuracies from two iterations. Reciprocal learning supports multiple iterations; however, here we only complete two iterations due to session time constraints. We leave the study of reciprocal learning with more iterations as future work.
\begin{table}[b]
\centering
 \begin{tabular}{c c c c} 
 \toprule
 Subject ID & Age & Gender & FM-UE \\
 \midrule
 S1 & 46 & Male & 27 \\ 
 S2 & 48 & Male & 26 \\
 S3 & 53 & Female & 26 \\
 S4 & 31 & Male & 50 \\
 S5 & 53 & Male & 47 \\ 
 \bottomrule
 \end{tabular}
 \caption{Subject information.}
 \label{tab:subjects}
\end{table}

\subsection{Subjects}

We recruited stroke survivors who presented with chronic hemiparesis and moderate muscle tone defined as Modified Ashworth Scale (MAS) scores $\leq$ 2 in the upper extremity. Our subjects can fully close their hands but are unable to completely extend their fingers without assistance. The passive range of motion in the fingers is within functional limits. Testing was approved by the Columbia University Institutional Review Board (IRB-AAAS8104) and was performed under the clinical supervision of an occupational therapist. See Table~\ref{tab:subjects} for details on subjects.

Our subjects have different degrees of hand impairment severity, as measured on the Fugl-Meyer scale for upper extremity (FM-UE). Subjects S1, S2, and S3 have no active finger extension (minimal observable movement) and lower corresponding FM-UE scores (27, 26, 26, respectively), whereas S4 and S5 have some residual active finger extension capacity (limited observable movement) and higher corresponding FM-UE scores (50, 47, respectively). 

\subsection{Hardware}

Our hand orthosis~\cite{park2018multimodal,chen2022thumb} is a robotic exotendon device with a forearm EMG armband and LED visual displays, as shown in Fig.~\ref{fig:teaser}. The device assists hand opening by extending all digits simultaneously. It does not directly assist finger flexion and relies on human strength for hand closing.

When operating under user control, the device acts on streaming \emph{open} and \emph{close} predictions at approximately 5 Hz to move to preset hand poses. The user is able to countermand the device mid-motion by generating an opposing intent input. \emph{Relax} predictions do not change device state.

\section{Results and Discussion}
\label{sec:exp}


\subsection{Intent Inferral Accuracy}

\begin{table}[t]
\centering
 \begin{tabular}{c c c c c c} 
 \toprule
 Subject & S1 & S2 & S3 & S4 & S5 \\
 \midrule
 Iteration 1 & 0.61 & 0.69 & \textbf{0.70} & 0.86 & \textbf{0.82} \\ 
 Iteration 2 & \textbf{0.88} & \textbf{0.71} & 0.68 & \textbf{0.94} & 0.80 \\
 \bottomrule
 \end{tabular}
 \caption{Mean intent inferral accuracy.}
 \label{tab:accuracy}
\end{table}

During the performance evaluation stage of each iteration, we use the test set to compute the classifier's per-timestep intent inferral accuracy against the ground truth labels. Table~\ref{tab:accuracy} reports the mean accuracies across samples. While the intent inferral accuracies for S2, S3, and S5 remain similar across the first two iterations, we observe an accuracy improvement for S1 and S4, from 0.61 to 0.88 and from 0.86 to 0.94, respectively. Augmented visual feedback during reciprocal learning practice may have helped S1 and S4 generate more consistent and distinguishable muscle activation patterns, creating more separable intent signals. 
As a result, the data collected after the first reciprocal learning practice become easier to classify. We note that S1 has more limited hand movement capacity compared to S4 (also reflected by their intent inferral accuracy). This suggests that whether reciprocal learning can improve intent inferral accuracy may not depend on stroke subjects' residual hand function. 

One possibility for why reciprocal learning does not improve performance across all subjects is the randomness of exploration in the practice stage. Although the prescribed arm conditions during the data collection stage of the first iteration provide data variance, due to the variation of impairment across stroke subjects, it is not guaranteed that the initial dataset is diverse enough to contain reproducible and distinguishable muscle activation patterns for all intents. In addition, even though the initial dataset collected with predefined arm conditions contains many activation patterns, due to randomness of exploration, it is not guaranteed for a subject to successfully locate or reproduce a specific pattern. 

While advancing the proposed paradigm to have a positive effect on most participants is obviously desirable and is the focus of our future work, we believe that an approach that improves performance on a subset of participants without hindering performance on the others can also have practical usefulness. Furthermore, a detailed study of the differences between cases showing successes versus more limited effects can suggest future areas of exploration.

\subsection{Data Separability}

\begin{figure*}
    \centering
    \includegraphics[width=0.9\textwidth]{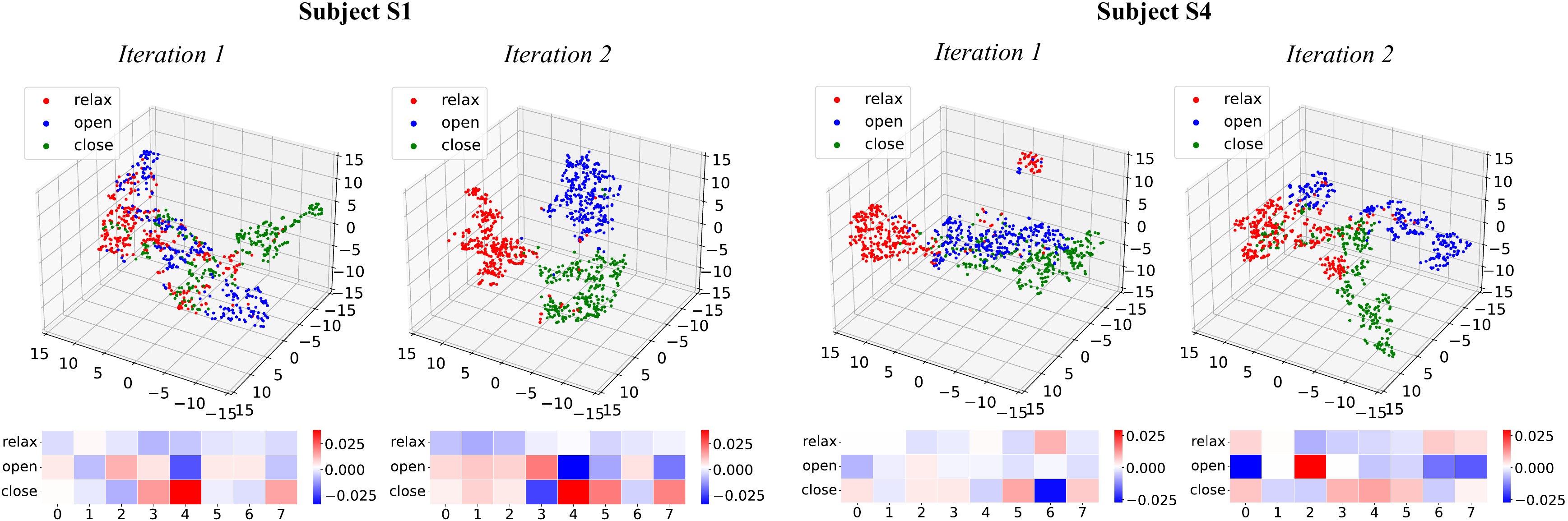}
    \caption{Reciprocal learning results on two stroke subjects. Top: comparing data separability. After reciprocal learning, clusters associated with each intent become more easily separable. Bottom: LDA classifier weights visualization indicating a specific EMG electrode (0 through 7) is negatively (blue) or positively (red) associated with a specific intent. Reciprocal learning strengthens such associations, which helps separability.}
    \label{fig:separability}
\end{figure*}

\begin{table}[t]
\centering
 \begin{tabular}{c c c c c c} 
 \toprule
 Subject & S1 & S2 & S3 & S4 & S5 \\
 \midrule
 Iteration 1 & $7.8\mathrm{e}{-6}$ & $1.0\mathrm{e}{-4}$ & $8.8\mathrm{e}{-4}$ & $1.1\mathrm{e}{-4}$ & $1.0\mathrm{e}{-4}$ \\ 
 Iteration 2 & $\mathbf{2.5\mathrm{\textbf{e}}{-4}}$ & $\mathbf{4.6\mathrm{\textbf{e}}{-4}}$ & $\mathbf{1.2\mathrm{\textbf{e}}{-3}}$ & $\mathbf{2.8\mathrm{\textbf{e}}{-4}}$ & $\mathbf{2.0\mathrm{\textbf{e}}{-4}}$ \\
 \bottomrule
 \end{tabular}
 \caption{Variance of LDA weights for the open intent.}
 \label{tab:variance}
\end{table} 

We further investigate the data of subjects S1 and S4 to show how reciprocal learning may lead to biosignals with more discriminative information for classifying intent. We visualize 1000 randomly sampled 8D EMG signals for each intent before and after the reciprocal learning practice in 3D space using t-distributed stochastic neighbor embedding \mbox{(t-SNE)}~\cite{van2008visualizing}, as shown in Fig.~\ref{fig:separability}. Each data point is labeled using its ground truth label, which is only available during training. After reciprocal learning, the data clusters associated with each intent improve in separability, which could explain the increased classification accuracy and lend support to the hypothesis that reciprocal learning may help some subjects generate signals that are more easily interpretable by intent inferral algorithms. 

We also visualize the LDA weights associated with each EMG electrode in Fig~\ref{fig:separability} to show the spatial distribution of predicted intents on the forearm. Positive values (red) mean that muscle activity under that particular electrode is positively associated with a specific intent, while negative values (blue) mean that muscle activity contributes negatively to that intent. We notice that after reciprocal learning practice, the weights (especially for the \textit{open} intent) become more diverse, meaning that the positive weights are more positive, and the negative weights are more negative. In fact, the variance of the 8D weights for the \textit{open} intent increases consistently for all subjects, as shown in Table~\ref{tab:variance}. These results suggest that subjects have stronger muscle activation patterns after reciprocal learning practice and explain improvements in intent inferral accuracy from another angle.

\section{Chapter Summary}

Different from SemiEMG and MetaEMG, which propose machine learning algorithms that work with limited labeled data, reciprocal learning addresses the data scarcity challenge by enabling a brand new learning direction, making subjects adapt to the behavior of the intent classifiers. This addresses the data scarcity challenge by generating higher-quality training data and reinforcing more distinguishable movement patterns, giving the classifier an easier time to learn. 

In the next chapter, instead of generating higher-quality training data from stroke subjects, we discuss how to dream up synthetic data. These data encourage the use of high-capacity models such as Transformers in our classifier, which has not yet been exploited in this thesis, due to the data-hungry nature of such large models.

%% file: chapters/chatemg.tex
\chapter{Generative AI for Expanding Datasets with Synthetic Data}
\label{chap:chatemg}

In this chapter, we discuss ChatEMG, an autoregressive generative model that can produce synthetic EMG samples to address the challenge of data scarcity.

\section{Motivation}

Many of the learning paradigms that we have discussed so far (semi-supervised learning, meta-learning, reciprocal learning) still rely on mid-capacity or low-capacity models such as MLP, LDA, or SVM. While larger capacity models such as deep neural networks and Transformers have revolutionized the computer vision and natural language processing fields, they require a lot of data to realize their full potential, which is difficult to get in the domain of rehabilitation robots. Low-capacity models are preferable when data is scarce, as they offer greater traceability. 
In order to bring the advantages of the latest and greatest high-capacity model advances into our application of intent inferral, we propose ChatEMG, which can generate synthetic samples to expand the limited training dataset. With the addition of extra synthetic samples, intent classifiers, especially the higher-capacity models, can achieve higher intent inferral accuracy. 

Like many applications in the assistive and rehabilitation devices domain, a fundamental challenge in intent inferral for stroke is the difficulty of collecting training data. The variation in EMG signals across conditions, sessions, and subjects makes the challenge even more pronounced. 
Firstly, EMG signal presents different patterns across subjects for the same intent due to variations in neuromuscular control impairments~\cite{tang2018surface,kyranou2018causes}.
In addition, defining a use \textit{session} as a single use of the device between donning and doffing, even for the same subject, the muscle tone and spasticity can vary across different sessions~\cite{la2024meta}. Furthermore, the signals are non-stationary and could change over time within a single session, depending on the use \textit{conditions}, such as the hand position and whether the motor is engaged and providing active grasping assistance~\cite{meeker2017emg,xu2022adaptive}. Due to such variation, intent classifiers trained for a specific condition/session/subject, do not generalize well, and classical solutions often require tedious data collection on every new condition/session/subject, introducing a significant burden on the participants. 

\begin{figure}
    \centering
    \includegraphics[width=0.65\textwidth]{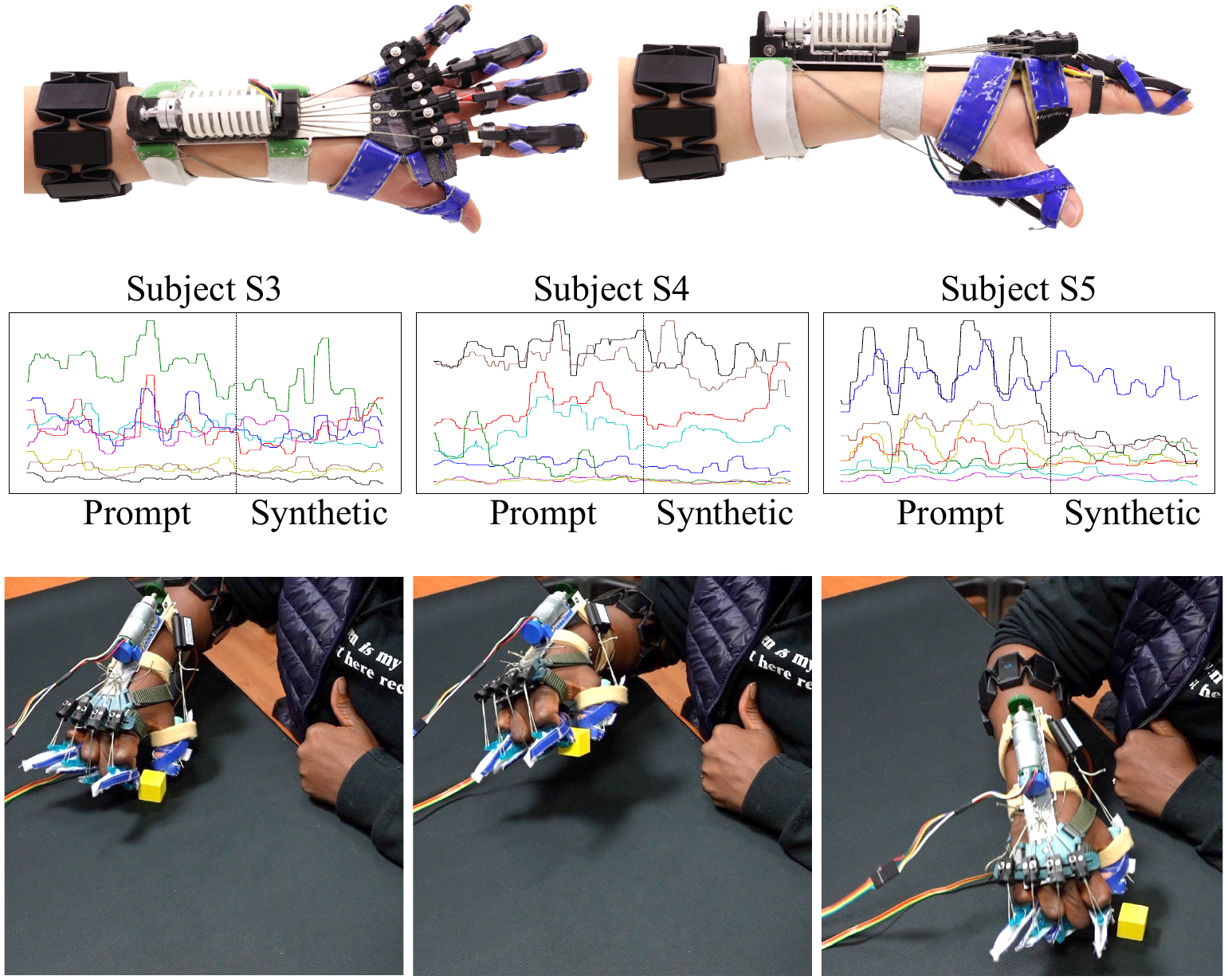}
    \caption{Approach overview. Our hand orthosis (top row) collects EMG data from a forearm armband and uses this data to infer the patient's intent. ChatEMG models trained on a large corpus of offline data can generate synthetic data (middle row) for a new patient conditioned on a prompt from a small dataset of the new patient, and specific to an intended arm movement. The synthetic and real data are then used to jointly train an intent classifier, which, in the course of the same session, enables functional pick-and-place tasks (bottom row) with the orthosis.}
    \label{fig:teaser_chatemg}
\end{figure}

In ChatEMG, we aim to reduce the burden of data collection from stroke subjects by generating synthetic data. We propose ChatEMG, an autoregressive generative model that understands the broad behavior of forearm EMG signals from a corpus of offline data across different stroke subjects and then can generate \textit{personalized} (i.e., condition-, session-, and subject-specific) synthetic samples conditioned on \textit{prompts} (i.e., a given sequence of EMG signals) sampled from a very limited dataset of a new condition, session, or subject. 


ChatEMG is Transformer-based, trained autoregressively, and temporal in its generative nature, meaning that each block of generated signals is conditioned on the previous blocks. The ability of ChatEMG to condition on a given sequence of EMG signals to generate synthetic signals of unlimited length is crucial to our application. Due to the significant variations of the EMG signals, the synthetic samples have to be personalized in order to be useful. As a result, ChatEMG leverages experience from previous data and produces synthetic samples conditioned on new data. In summary, our contributions are as follows:

\begin{itemize}
    \item We propose ChatEMG, an approach for producing synthetic EMG data via generative training on data also collected from stroke patients. Unlike previous models, ChatEMG can be conditioned on limited data from a new condition/session/subject in order to generate personalized synthetic sequences of arbitrary lengths. 
    \item We show that data generated by ChatEMG improves intent inferral performance for a broad range of intent classifiers. To the best of our knowledge, this is the first time that synthetic data has been shown to improve intent inferral performance when using real data from stroke patients.
    \item Our complete new patient protocol (collecting limited new data, using ChatEMG to generate personalized synthetic data, and then training an intent classifier), can be integrated into a single patient session. This increases the applicability of our method for functional tasks with real-world patients. To the best of our knowledge, this is the first time that an intent classifier trained partially on synthetic data has been deployed for functional orthosis control by a stroke patient.
\end{itemize}

\section{Overview}
\label{sec:overview}

The ultimate goal of this project is to develop intent inferral classifiers that can predict stroke subjects' intent, so that our orthosis can provide meaningful functional assistance. Specifically, based on EMG data, we aim to predict which movement out of $\{open,close,relax\}$ the subject intends to perform with their hand. If the classifier predicts that the user intends to open, the device retracts the tendon, extending the fingers. If the user intends to close, the device extends the tendon, allowing the user to use their own grip strength to close their hand. If the predicted intent is to relax, the device maintains its previous state.

\begin{figure*}
    \centering
    \includegraphics[width=\textwidth]{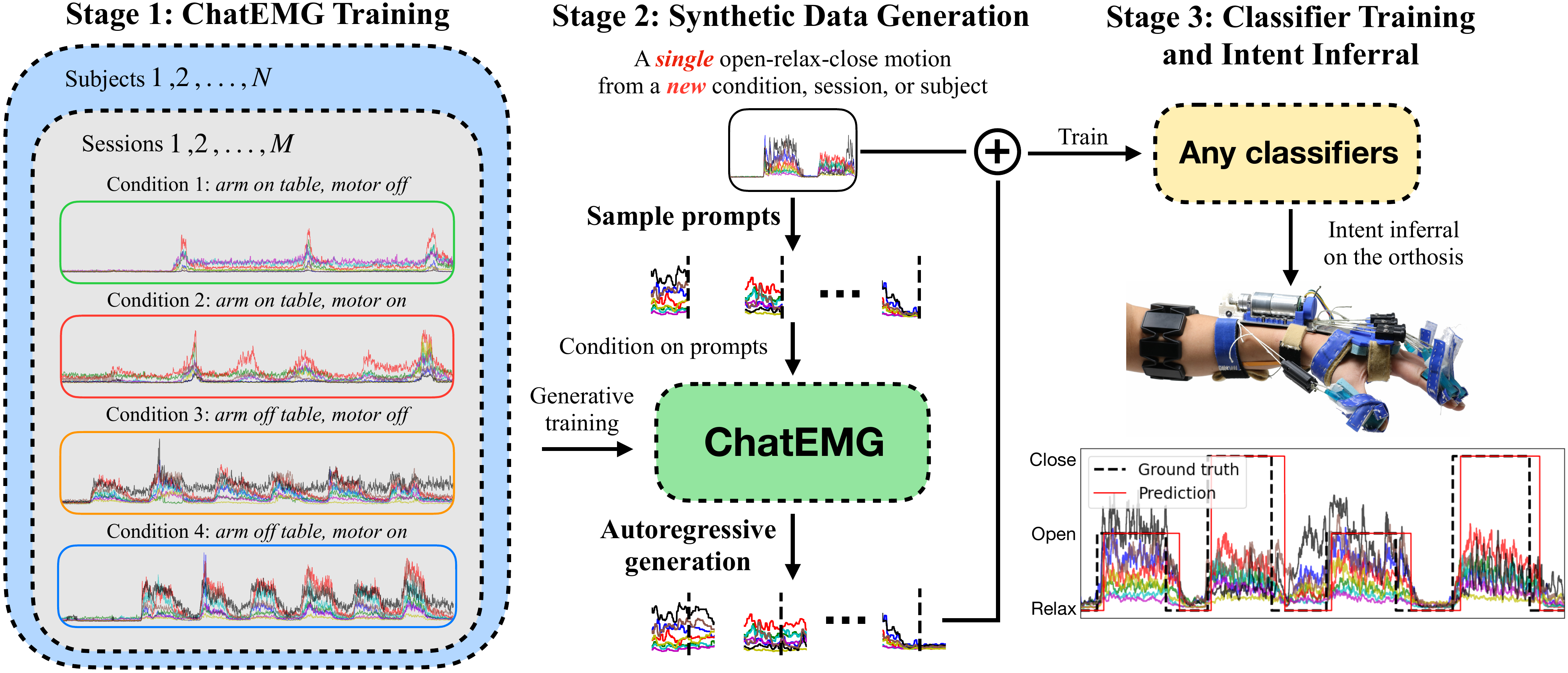}
    \caption{ChatEMG overview. Stage 1: ChatEMG is trained on large offline data from different conditions, sessions, and subjects. We visualize the EMG recordings from different conditions of the same subject in a single session. As shown here, there is a drastic variation in EMG signals for different conditions. Stage 2: we only need a very limited labeled dataset from a new condition, session, and subject, and we use ChatEMG to expand this limited dataset with synthetic samples. These synthetic samples are conditioned on prompts from the new condition/session/subject. Stage 3: we train intent inferral classifiers using both the synthetic samples and the original limited dataset. Running the classifier, our orthosis can then provide active assistance for the stroke subjects in functional tasks.}
    \label{fig:pipeline}
\end{figure*}

The most direct way to achieve this goal is to collect a set of training data (EMG signals) labeled with ground truth intent. This can be done by instructing the patient to attempt one of the three hand movements of interest, while simultaneously recording EMG data and labeling it with the prescribed intent. Once enough labeled data is collected for each of the three possible intent classes, we can train a classifier to distinguish between them. 

However, this traditional approach suffers from two key limitations: (1) The process of collecting labeled training data is burdensome and time-consuming for both the patient and the experimenter. It uses up precious session time and also fatigues the patients, leading to increased muscle tone and spasticity. (2) EMG signals exhibit significant variations between different conditions, sessions, and subjects. Training data collected in one condition/session/subject is unlikely to apply to a different one, leading to very poor generalization performance of the classifier. 

Our approach aims for a different paradigm. Its goal is to quickly adapt to a new condition, session or subject, using only a very small amount of newly collected, labeled training data. To achieve that, it relies on synthetic data from a generative model trained on a very large corpus of previously collected labeled data from a variety of conditions, sessions, and subjects. Concretely, our approach consists of three stages, illustrated in Fig.~\ref{fig:pipeline} and described below:

\subsubsection{ChatEMG Generative Training on Large Offline Data} In the first stage, a number of generative ChatEMG models are trained on a large corpus of offline data $\mathcal{D}_{\text{offline}}$ collected from different stroke subjects, which includes various conditions and sessions. One ChatEMG model is trained for each intent (open/close/relax). Once trained, each such model is able to generate synthetic data matching its respective intent. 

\subsubsection{Synthetic Data Generation Conditioned on Small Prompts} When a new condition/session/subject is started, we collect a very small labeled dataset $\mathcal{D}_{\text{new}}^{\text{orig}}$ in the new setting. We then use the ChatEMG models to extend this dataset through synthetic data generation. Concretely, for each possible intent, we use its respective model to generate additional synthetic data, referred to as $\mathcal{D}_{\text{new}}^{\text{synth}}$. 

Critically, $\mathcal{D}_\text{new}^\text{synth}$ is generated by models trained on $\mathcal{D}_{\text{offline}}$, but prompted with data sampled from $\mathcal{D}_\text{new}^\text{orig}$. The autoregressive nature of ChatEMG enables it to generate synthetic data of unlimited length conditioned on an existing piece of EMG sequence, which we call prompts. The ability to condition on prompts means that our synthetic data is based on knowledge mined from a large repository of previous data but applied in the context of the current condition, session, or subject. \textit{This is the essence of ChatEMG: it can leverage a vast repository of previous data via generative training while still remaining condition-, session- and subject-specific via prompting.}
    
\subsubsection{Classifier Training and Intent Inferral} Once the personalized synthetic data has been generated, we are ready to train an intent inferral classifier for the current situation. We train this classifier on both $\mathcal{D}_\text{new}^\text{orig}$ and $\mathcal{D}_\text{new}^\text{synth}$. We can then use this classifier for live intent inferral on our orthosis. 

It is worth noting that our approach is agnostic to the type of classifier used here. As we will show in the results section, this approach can be used with a variety of classifier architectures and generally improves their performance.

\section{ChatEMG Generative Models}

The role of a ChatEMG model is to take in a sequence of EMG signals as input and predict the next EMG signal as output, where one signal consists of the 8-channel data from our EMG armband. We use the Myo armband from the Thalmic Labs, which has 8 electrodes covering the forearm and collects signals at 100Hz. Such a model can then be used autoregressively to generate synthetic data of arbitrary length, all conditioned on the given prompt. We note that this approach is similar in concept to language models such as ChatGPT~\cite{openai2024chatgpt}, capable of generating text in response to a text prompt. However, ChatEMG operates on the ``language'' of EMG data, hence its chosen moniker.

We train one ChatEMG model on data corresponding to each user intent. The goal is then for each of these models to generate synthetic data corresponding to the intent on which it has been trained. While each of these models is trained on different data, their architecture is the same.

\subsection{Architecture}

\begin{figure}
    \centering
    \includegraphics[width=0.65\linewidth]{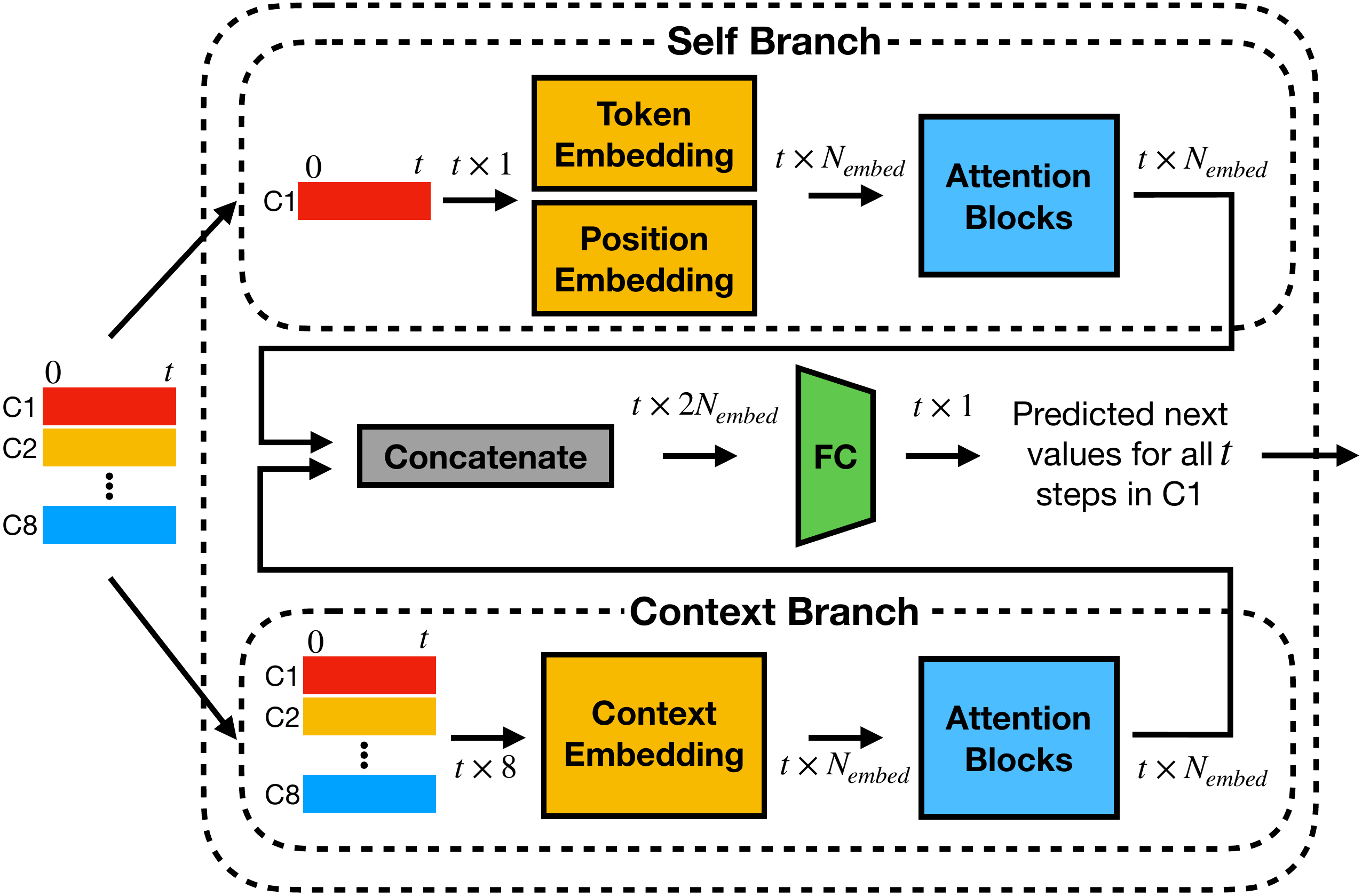}
    \caption{ChatEMG model architecture. ChatEMG has two branches: the self branch that takes in the first channel (C1) and the context branch that takes in all 8-channel EMG signals. 
    }
    \label{fig:architecture}
\end{figure}

\begin{figure}
    \centering
    \includegraphics[width=0.65\linewidth]{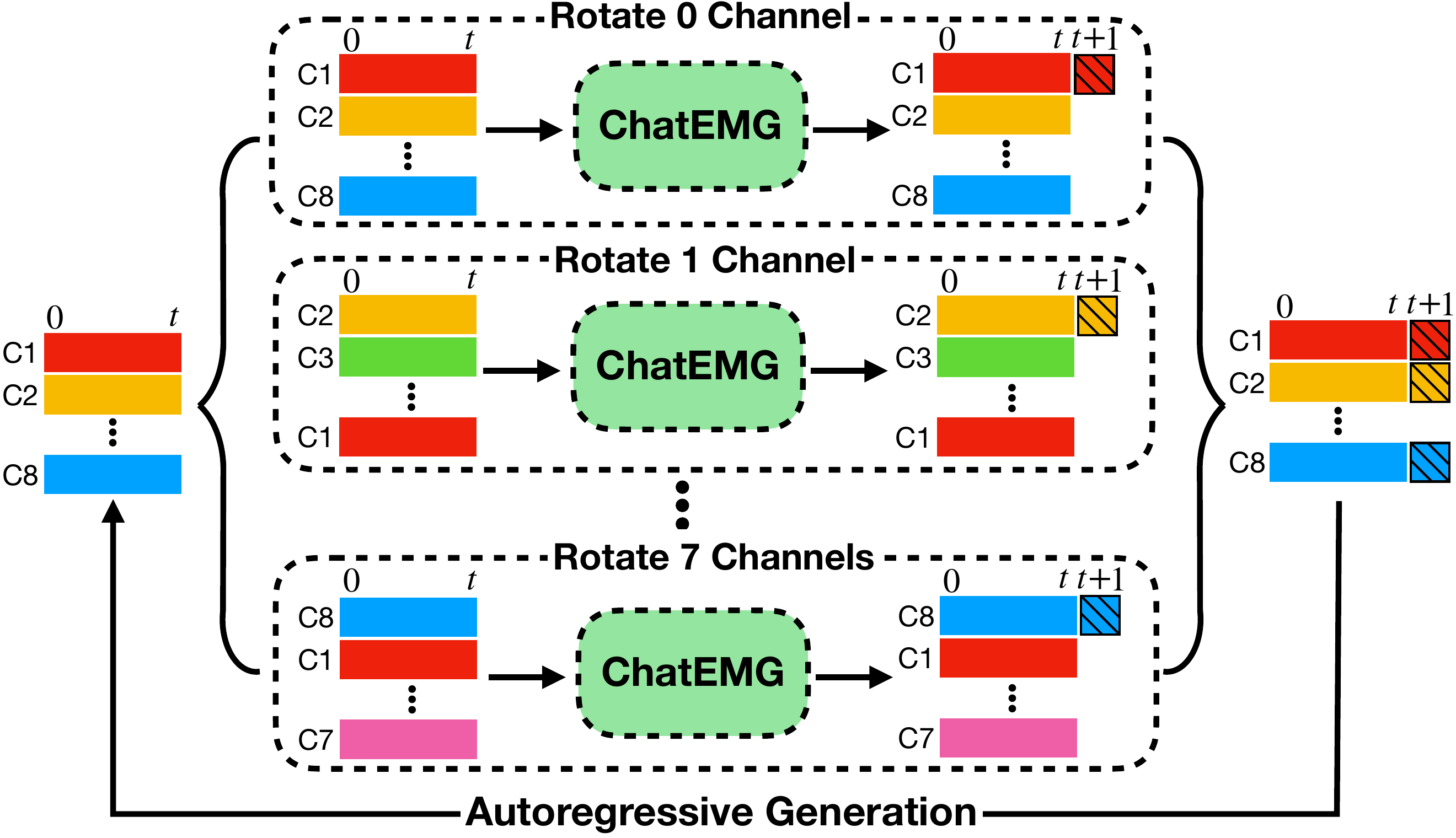}
    \caption{Autoregressive synthetic data generation. ChatEMG only predicts the next EMG value for the first channel (C1) and in order to generate the complete 8-channel EMG signal, we rotate the input signals 7 times such that all other channels will become the first channel once. 
    }
    \label{fig:autoregressive}
\end{figure}

Each ChatEMG model is a Transformer-based decoder-only model with only self-attention mechanisms, similar to ChatGPT, as shown in Fig.~\ref{fig:architecture}. The input to this model consists of a time sequence of EMG signals. The attention mechanism of Transformers allows the input to be arbitrary length. Given an input sequence of length $t$, the output is a vector, also of length $t$, which contains the predicted next values for all $t$ steps in the first channel. ChatEMG only predicts the EMG value of the first channel such that the output value follows a discrete vocabulary of limited size, which we discuss later in this section. The $i$th element of this output vector is the predicted $(i+1)$th element of the first channel, with attention up to the $i$th signals in the input sequence. Details of the attention mechanism can be found in Vaswani et al.~\cite{vaswani2017attention}. 

During training, the whole output vector is compared with the ground truth next values to compute the loss. During synthetic data generation, we only use the last prediction of this output vector. In order to generate the next complete 8-channel EMG signal, we rotate the input EMG signals 7 times (one channel per time) so that each of the other 7 channels can become the first channel of the input EMG signals. We then append this newly generated signal to the input signals and continue the generation process, as shown in Fig.~\ref{fig:autoregressive}. The autoregressive nature of this architecture allows us to generate an output sequence of arbitrary length. 

One key consideration for designing the ChatEMG architecture is its ability to generate diverse output given the same input signals. As a result, the output of the model is sampled from a probability distribution over a discrete vocabulary, and during training, we compute the cross-entropy loss between the predicted distribution and the one-hot label. Furthermore, we can generate an arbitrary number of ``likely'' next signals simply by repeating the sampling process. Each sequence can then be continued in an autoregressive fashion. This means that one generative model can use a single prompt to generate an arbitrary number of likely next sequences, each of arbitrary length. This is an important feature of our model, given that we will need large amounts of synthetic data in order to train downstream classifiers\footnote{As an alternative, we also tried a variational regression setup that predicts a mean and variance for the next signal, and used the reparameterization trick to backpropagate the loss, which also allows sampling an arbitrary number of completions for a single prompt. However, we found empirically that the balance between the KL (Kullback-Leibler) loss and the reconstruction loss is hard to tune, and its generated signals are not as good as the classification version.}.

However, this architecture requires a discrete vocabulary of finite size, which is challenging given that our raw data consists of an 8-channel EMG signal. For preprocessing, we smooth out the data using a median filter of size 9 and then bin and clip each channel of the EMG signal to be an integer between 0 and 1000. Under this range, if we model the whole 8-channel signal as one ``token'' in our vocabulary, then our vocabulary size becomes $1000^8$, which is too large to predict a probability distribution and sample from. As a result, ChatEMG always predicts the next EMG value for only the first channel, making the vocabulary size 1000.

\subsection{Modelling Inter-channel Relationship}

Although our model predicts one channel at a time, it still considers the interchannel relationship. Our EMG armband has 8 electrodes surrounding the forearm, and each electrode covers a particular area of the forearm muscle. The interchannel relationship across different electrodes can be useful information to leverage for understanding user intent. 

The model of ChatEMG has two branches: the self branch, which takes in the first channel for which we are predicting the next EMG value, and the context branch, which takes in all 8-channel EMG signals (shown in Fig.~\ref{fig:architecture}). We use an embedding size of $N_{embed} = 256$.
The self branch uses both token and position embedding layers to compute the embedding. The context embedding block consists of 8 separate token embedding layers for each channel and one shared positional embedding layer. The channel-specific token embeddings are then summed with the shared positional embedding to create the final embedding of the context branch. Each branch has 12 attention blocks, and each block uses an 8-head attention mechanism. The output from both branches is concatenated to pass through another 3-layer fully connected network (FC).

Our model is trained with EMG sequences of length $t = 256$ (2.56 seconds at 100Hz). The dimension of the model output is $256 \times 1$, and they are the 256 predicted EMG values for the next time step of the first channel.
During training, we also augment the input signals by rotating the channels seven times to simulate the rotation of the electrodes. This data augmentation strategy enables ChatEMG to be invariant with channel rotation.
During data generation, we sample EMG prompts of length 150 (corresponding to 1.5s) from the very limited dataset of the new condition, session, or subject and use ChatEMG to autoregressively complete the rest of the signal to a length of 256, which is the time-series length that our classification algorithms take.

\section{Experimental Setup}
The ultimate goal of orthosis is to provide meaningful functional assistance for stroke subjects, enabled by intent inferral. Thus, we examine whether the generated synthetic samples by ChatEMG can improve intent inferral accuracy. 

\subsection{Subjects}

We performed experiments with 5 chronic stroke survivors having hemiparesis and moderate muscle tone: Modified Ashworth Scale (MAS) scores $\leq$ 2 in the upper extremity. Our MAS criteria exclude subjects whose fingers are difficult to move passively --- fingers with more severe spasticity cannot be quickly extended with external forces without increasing muscle tone and risking damage to the joints. Our participants can fully close their hands but are unable to completely extend their fingers without assistance. The passive range of motion in the fingers is within functional limits. Testing was approved by the Columbia University Institutional Review Board (IRB-AAAS8104) and was performed under the clinical supervision of an occupational therapist.

Our subjects have different hand impairments, and their Fugl-Meyer scores for upper extremity (FM-UE) vary. Subjects S1, S2, and S3 have no active finger extension (lower functioning) and have a corresponding low FM-UE score (27, 26, 26, respectively), whereas S4 and S5 have some residual active finger extension ability (higher-functioning) with a higher FM-UE score (50, 47, respectively).

\subsection{Data Collection Protocol}
\label{sec:protocol}

For each stroke subject, we collect data from two sessions on two different days using the following protocol. A \textit{session} is defined to be a single use of the device between donning and doffing the device. We intentionally keep the two sessions for each subject at least one week apart to better study the variation of the EMG signals across different days. For each session, we collect data under four different \textit{conditions}: 1) with the arm resting on a table and the orthosis motor off \{\textit{arm on table, motor off}\}, 2) with the arm resting on a table and the orthosis motor on, providing active grasp assistance \{\textit{arm on table, motor on}\}, 3) with the arm raised above the table and the orthosis motor off \{\textit{arm off table, motor off}\}, and 4) with the arm raised above the table and the orthosis motor on \{\textit{arm off table, motor on}\}.

We collect two continuous, uninterrupted \textit{recordings} for each condition, and for each recording, we instruct the subjects to open and close their hands three times by giving verbal cues of open, close, and relax. We simultaneously record the EMG signals and verbal cues as ground truth intent labels. Each verbal cue lasts for 5 seconds, and there is a relax cue between each open and close cue. For conditions where the motor is on, we move the motor approximately one second after the verbal cue is given using a dedicated button. We define each opening and closing hand completion as one round of \textit{open-relax-close motion}. Each recording then contains three open-relax-close motions. We note that this protocol is at the maximum capacity that stroke subjects can follow during a 90-minute session, and we can observe increased spasticity and fatigue at the end. 

\subsection{Assessment Scenarios}

We create different assessment scenarios (listed below) that simulate different use cases of ChatEMG, by selecting different \textit{training recordings} (recordings used to train ChatEMG) and \textit{intent inferral recordings} (recordings used to perform intent inferral evaluation). These scenarios evaluate how well ChatEMG generalizes and adapts to new conditions/sessions/subjects not seen in its training recordings. 

\subsubsection{Condition Adaptation} This scenario studies whether ChatEMG can generalize to a new condition. The training recordings are of condition \{\textit{arm on table, motor off}\} from all five subjects (including both sessions), and the intent inferral recordings are of condition \{\textit{arm off table, motor off}\}. We note that \{\textit{arm on table, motor off}\} is the most effortless condition for us to collect data in, while \{\textit{arm off table, motor off}\} is the closest condition to an ongoing functional pick-and-place task. Thus, this simulates a scenario where ChatEMG is trained on data collected in the effortless condition and used to generate synthetic samples for a drastically different but realistic condition. 

\subsubsection{Session Adaptation} This scenario pertains to the signal variation across different use sessions, and it seeks to simulate using the orthosis on a subject seen previously in a different session on a different day. ChatEMG is trained on recordings of the first session from all subjects, and the intent inferral recordings are those of the second session.

\subsubsection{Subject Adaptation} This scenario simulates onboarding new subjects. We conduct five separate experiments, each one simulating the onboarding of one holdout subject, given that we have seen the other four. In each experiment, we train ChatEMG using all the recordings from the other subjects (including both sessions), and the intent inferral recordings are those of the holdout subject. When adapting to a new subject, it is also implicitly adapting to a new session. However, in our session adaption experiments, we assume it is a different session of a previously seen subject. 

We use a subset of the training recordings (around 314K samples of size 256 by 8) to train ChatEMG, and we use the remaining training recordings (around 204K samples) as the validation set. We early stop the generative training before the validation loss increases to avoid overfitting.

\vspace{-0.1in}
\subsection{Intent Inferral Classifiers}

ChatEMG is classifier-agnostic, and the generated synthetic samples can be integrated with the training set of any classifiers. We study three types of classifiers: linear discriminant analysis (LDA), random forests (RF), and Transformer. They are popular in the biomedical literature and cover both classic machine-learning algorithms and high-capacity neural networks. We feed into each classifier a time series of length 256 (2.56s), in the shape of 256 by 8. The EMG signals are flattened into a single vector for LDA and RF. We use a single 4-head attention block followed by a 3-layer multilayer perception (MLP) for the Transformer classifier. We perform the same preprocessing techniques as training the ChatEMG model, and as an additional step, we normalize the EMG signals into the range of $[-1, 1]$.

\subsection{Baselines}

The intent inferral evaluation is done on individual intent inferral recordings. For each recording, we assume only a small \textit{support set} (i.e., the first open-relax-close motion of the recording) is available for training the classifier. The support set simulates the limited new training samples from a new condition, session, or subject, denoted by $\mathcal{D}_{\text{new}}^{\text{synth}}$ in Sec.~\ref{sec:overview}. We then test the classifier's accuracy using the \textit{query set} (i.e., the second and third open-relax-close motions). 

\subsubsection{Self} This method trains the intent inferral classifier using only the support set of the intent inferral recordings. 
\subsubsection{Fine-tune} This method pre-trains the classifier using all the training recordings of ChatEMG and then fine-tunes on the support set of the intent inferral recordings. This baseline ensures the training data for ChatEMG is also accessible for a fair comparison.
\subsubsection{ChatEMG} This is our proposed method. We repetitively sample prompts of size 150 (1.5s) from the small support set and leverage ChatEMG models to expand the prompts to size 256 (2.56s). These synthetic samples are then combined with the original support set to train intent classifiers. For each intent, we add 1000 synthetic samples.

\section{Results and Discussion}

In this section, we first discuss the intent inferral accuracy and then analyze the synthetic samples. Finally, we show that ChatEMG can help improve the performance of functional pick-and-place tasks in real-world hospital testing. 
Visit our project website at \url{https://jxu.ai/chatemg} for hospital testing demonstrations and additional information.

\subsection{Intent Inferral Performance}

\begin{table*}
    \setlength{\tabcolsep}{3pt}
    \footnotesize
    \renewcommand{\arraystretch}{0.6}
    \centering
    \begin{tabular}{c|c|cccccc|c}
    \toprule
    & & \textbf{S1} & \textbf{S2} & \textbf{S3} & \textbf{S4} & \textbf{S5} & \textbf{Average} & \begin{tabular}[c]{@{}c@{}}\textbf{$p$-value w\slash}\\\textbf{\textit{ChatEMG}} \end{tabular} \\
    \midrule
    \multirow{3}{*}{LDA} & \textit{Self} & $0.37 \pm 0.15$ & $0.54 \pm 0.07$ & $0.46 \pm 0.10$ & $0.64 \pm 0.22$ & $0.36 \pm 0.07$ & $0.48$ & $\mathbf{3\mathrm{\textbf{e}}{-4}}$ \\
    & \textit{Fine-tune} & $0.34 \pm 0.12$ & $0.64 \pm 0.10 $ & $0.54 \pm 0.05$ & $0.71 \pm 0.05$ & $0.48 \pm 0.15$ & $0.54$ & $\mathbf{4\mathrm{\textbf{e}}{-4}}$\\
    & \textit{ChatEMG} & $0.45 \pm 0.06$ & $0.69 \pm 0.02$ & $0.70 \pm 0.09$ & $0.90 \pm 0.04$ & $0.68 \pm 0.04$ & $\textbf{0.68}$ & --- \\
    \midrule
    \multirow{3}{*}{RF} & \textit{Self} & $0.52 \pm 0.07$ & $0.72 \pm 0.08$ & $0.64 \pm 0.10$ & $0.90 \pm 0.08$ & $0.77 \pm 0.06$ & $0.71$ & $1\mathrm{e}{-1}$\\
    & \textit{Fine-tune} & $0.53 \pm 0.11$ & $0.68 \pm 0.06$ & $0.63 \pm 0.04$ & $0.89 \pm 0.08$ & $0.64 \pm 0.17$ & $0.67$ & $\mathbf{5\mathrm{\textbf{e}}{-2}}$\\
    & \textit{ChatEMG} & $0.53 \pm 0.14$ & $0.70 \pm 0.11$ & $0.71 \pm 0.11$ & $0.92 \pm 0.04$ & $0.77 \pm 0.05$ & $\textbf{0.73}$ & --- \\
    \midrule
    \multirow{3}{*}{Transformer} & \textit{Self} & $0.55 \pm 0.05$ & $0.64 \pm 0.05$ & $0.58 \pm 0.10$ & $0.84 \pm 0.06$ & $0.73 \pm 0.07$ & $0.67$ & $\mathbf{3\mathrm{\textbf{e}}{-2}}$ \\
    & \textit{Fine-tune} & $0.57 \pm 0.06$ & $0.64 \pm 0.05$ & $0.70 \pm 0.03$ & $0.88 \pm 0.04$ & $0.68 \pm 0.06$ & $0.69$ & $2\mathrm{e}{-1}$\\
    & \textit{ChatEMG} & $0.60 \pm 0.15$ & $0.72 \pm 0.08$ & $0.65 \pm 0.01$ & $0.86 \pm 0.04$ & $0.78 \pm 0.04$ & $\textbf{0.72}$ & --- \\
    \bottomrule
    \end{tabular}
    \caption{Condition adaptation experiment results.}
    \label{tab:cross_condition}
\end{table*}

\begin{table*}
    \setlength{\tabcolsep}{3pt}
    \footnotesize
    \renewcommand{\arraystretch}{0.6}
    \centering
    \begin{tabular}{c|c|cccccc|c}
    \toprule
    & & \textbf{S1} & \textbf{S2} & \textbf{S3} & \textbf{S4} & \textbf{S5} & \textbf{Average} & \begin{tabular}[c]{@{}c@{}}\textbf{$p$-value w\slash}\\\textbf{\textit{ChatEMG}} \end{tabular} \\
    \midrule
    \multirow{3}{*}{LDA} & \textit{Self} & $0.40 \pm 0.16$ & $0.50 \pm 0.09$ & $0.52 \pm 0.07$ & $0.50 \pm 0.13$ & $0.81 \pm 0.05$ & $0.54$ & $\mathbf{6\mathrm{\textbf{e}}{-5}}$ \\
    & \textit{Fine-tune} & $0.64 \pm 0.04$ & $0.49 \pm 0.07$ & $0.65 \pm 0.05$ & $0.65 \pm 0.02$ & $0.64 \pm 0.13$ & $0.61$ & $\mathbf{1\mathrm{\textbf{e}}{-2}}$\\
    & \textit{ChatEMG} & $0.54 \pm 0.07$ & $0.55 \pm 0.05$ & $0.64 \pm 0.07$ & $0.77 \pm 0.08$ & $0.83 \pm 0.08$ & $\textbf{0.67}$ & --- \\
    \midrule
    \multirow{3}{*}{RF} & \textit{Self} & $0.55 \pm 0.08$ & $0.57 \pm 0.07$ & $0.77 \pm 0.11$ & $0.82 \pm 0.13$ & $0.77 \pm 0.06$ & $0.69$ & $1\mathrm{e}{-1}$ \\
    & \textit{Fine-tune} & $0.58 \pm 0.06$ & $0.62 \pm 0.02$ & $0.74 \pm 0.16$ & $0.72 \pm 0.16$ & $0.78 \pm 0.05$ & $0.69$ & $2\mathrm{e}{-1}$ \\
    & \textit{ChatEMG} & $0.57 \pm 0.07$ & $0.66 \pm 0.03$ & $0.72 \pm 0.15$ & $0.82 \pm 0.13$ & $0.79 \pm 0.07$ & $\textbf{0.71}$ & --- \\
    \midrule
    \multirow{3}{*}{Transformer} & \textit{Self} & $0.66 \pm 0.06$ & $0.53 \pm 0.05$ & $0.77 \pm 0.09$ & $0.67 \pm 0.15$ & $0.72 \pm 0.15$ & $0.67$ & $\mathbf{2\mathrm{\textbf{e}}{-2}}$ \\
    & \textit{Fine-tune} & $0.72 \pm 0.06$ & $0.57 \pm 0.12$ & $0.78 \pm 0.12$ & $0.79 \pm 0.09$ & $0.79 \pm 0.08$ & $\textbf{0.73}$ & $5\mathrm{e}{-1}$ \\
    & \textit{ChatEMG} & $0.64 \pm 0.12$ & $0.67 \pm 0.05$ & $0.77 \pm 0.14$ & $0.79 \pm 0.04$ & $0.77 \pm 0.12$ & $\textbf{0.73}$ & --- \\
    \bottomrule
    \end{tabular}
    \caption{Session adaptation experiment results.}
    \label{tab:cross_session}
\end{table*}

\begin{table*}
    \setlength{\tabcolsep}{3pt}
    \footnotesize
    \renewcommand{\arraystretch}{0.6}
    \centering
    \begin{tabular}{c|c|cccccc|c}
    \toprule
    & & \textbf{S1} & \textbf{S2} & \textbf{S3} & \textbf{S4} & \textbf{S5} & \textbf{Average} & \begin{tabular}[c]{@{}c@{}}\textbf{$p$-value w\slash}\\\textbf{\textit{ChatEMG}} \end{tabular} \\
    \midrule
    \multirow{3}{*}{LDA} & \textit{Self} & $0.33 \pm 0.06$ & $0.61 \pm 0.11$ & $0.51 \pm 0.11$ & $0.52 \pm 0.26$ & $0.65 \pm 0.18$ & $0.51$ & $\mathbf{1\mathrm{\textbf{e}}{-2}}$ \\
    & \textit{Fine-tune} & $0.60 \pm 0.08$ & $0.62 \pm 0.10$ & $0.49 \pm 0.06$ & $0.74 \pm 0.03$ & $0.40 \pm 0.17$ & $0.56$ & $3\mathrm{e}{-1}$ \\
    & \textit{ChatEMG} & $0.37 \pm 0.18$ & $0.63 \pm 0.11$ & $0.56 \pm 0.16$ & $0.68 \pm 0.15$ & $0.70 \pm 0.14$ & $\textbf{0.58}$ & --- \\
    \midrule
    \multirow{3}{*}{RF} & \textit{Self} & $0.60 \pm 0.01$ & $0.64 \pm 0.02$ & $0.59 \pm 0.02$ & $0.80 \pm 0.08$ & $0.71 \pm 0.07$ & $0.66$ & $3\mathrm{e}{-1}$ \\
    & \textit{Fine-tune} & $0.54 \pm 0.02$ & $0.65 \pm 0.02$ & $0.55 \pm 0.12$ & $0.78 \pm 0.07$ & $0.75 \pm 0.13$ & $0.65$ & $\mathbf{5\mathrm{\textbf{e}}{-2}}$ \\
    & \textit{ChatEMG} & $0.52 \pm 0.01$ & $0.65 \pm 0.06$ & $0.55 \pm 0.12$ & $0.86 \pm 0.06$ & $0.80 \pm 0.08$ & $\textbf{0.67}$ & --- \\
    \midrule
    \multirow{3}{*}{Transformer} & \textit{Self} & $0.57 \pm 0.05$ & $0.63 \pm 0.03$ & $0.63 \pm 0.06$ & $0.72 \pm 0.04$ & $0.65 \pm 0.16$ & $0.64$ & $\mathbf{1\mathrm{\textbf{e}}{-2}}$ \\
    & \textit{Fine-tune} & $0.56 \pm 0.08$ & $0.64 \pm 0.02$ & $0.60 \pm 0.10$ & $0.68 \pm 0.08$ & $0.69 \pm 0.04$ & $0.64$ & $\mathbf{4\mathrm{\textbf{e}}{-2}}$ \\
    & \textit{ChatEMG} & $0.58 \pm 0.13$ & $0.64 \pm 0.07$ & $0.67 \pm 0.06$ & $0.76 \pm 0.03$ & $0.74 \pm 0.14$ & $\textbf{0.68}$ & --- \\
    \bottomrule
    \end{tabular}
    \caption{Subject adaptation experiment results.}
    \label{tab:cross_subject}
\end{table*}

The results are shown in Table~\ref{tab:cross_condition},~\ref{tab:cross_session} and~\ref{tab:cross_subject}.
For each subject, we evaluate three intent inferral recordings, and we present the average accuracy and one standard deviation. ChatEMG is able to improve the average intent inferral accuracy across five subjects for all classifiers under all assessment scenarios. This shows that ChatEMG can successfully generalize to a new condition, session, or subject despite not seeing them in its training recordings. 

Despite that the improvement in intent inferral accuracy is consistent across 17/18 comparisons between ChatEMG and Fine-tune/Self, we further investigate the statistical significance of such improvement by performing a one-sided Wilcoxon rank-sum test on the results aggregated across all subjects (three intent inferral recordings per subject). We choose a non-parametric statistical test because we do not assume an underlying normal distribution. We report the computed $p$-values for pairwise differences between our ChatEMG and the other methods. With a commonly used hypothesis threshold of $\alpha = 5\mathrm{e}{-2}$, 11/17 improvements (\mbox{$p$-values} in bold) are statistically significant.

We notice that if trained only on the small support set (Self), RF tends to have the highest performance, while when the classifier has access to larger datasets (Fine-tune or ChatEMG), Transformer tends to perform better. This matches our intuition that larger-capacity models can realize their potential only when given enough data, and ChatEMG achieves that through synthetic data generation. 

Subject adaptation is the most difficult scenario with the lowest intent inferral accuracy. It simulates the scenarios of onboarding a new stroke subject, and we only collect one round of open-relax-close motion from this new subject as our support set. It is the most tricky scenario because variation in EMG signals is larger among different subjects than among different conditions or sessions of the same subject. However, ChatEMG can still understand the broad signal patterns of different intents from past subjects and apply that knowledge by generating synthetic samples conditioned on prompts from the new subject. S4 and S5 tend to have higher intent inferral accuracy than S1, S2, and S3, which matches the hand-functionality measured by the FM-UE scores.

\subsection{Synthetic Sample Visualization and Analysis}

\begin{figure}
    \centering
    \includegraphics[width=0.65\linewidth]{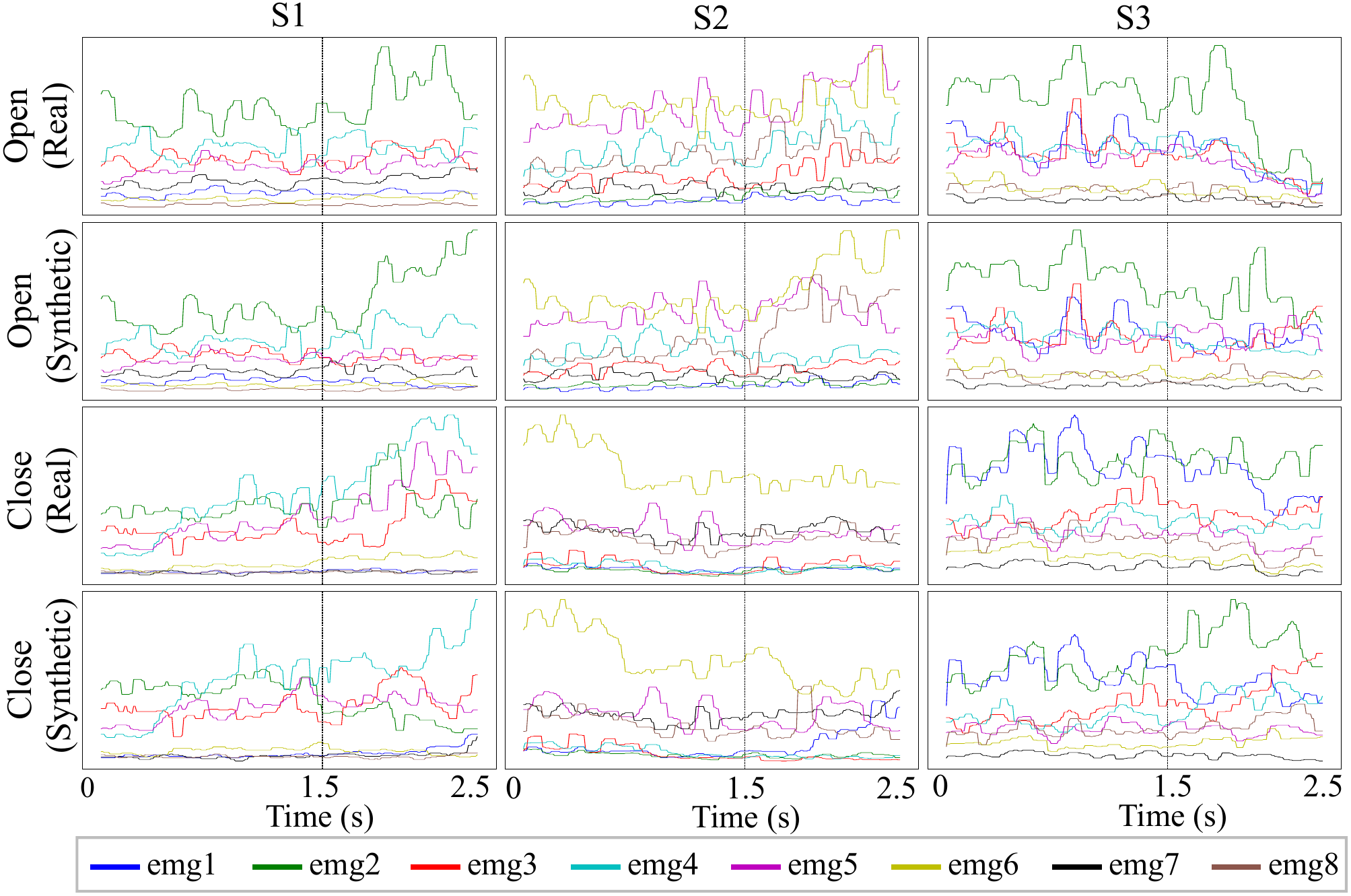}
    \caption{Comparison between the real and synthetic samples on open and close intents of subjects S1, S2, and S3. The vertical line indicates the switch from the provided prompt to the generated synthetic sequence. These samples also demonstrate the significant variations in EMG signals across different stroke subjects.}
    \label{fig:samples}
\end{figure}

\begin{figure}
    \centering
    \includegraphics[width=0.65\linewidth]{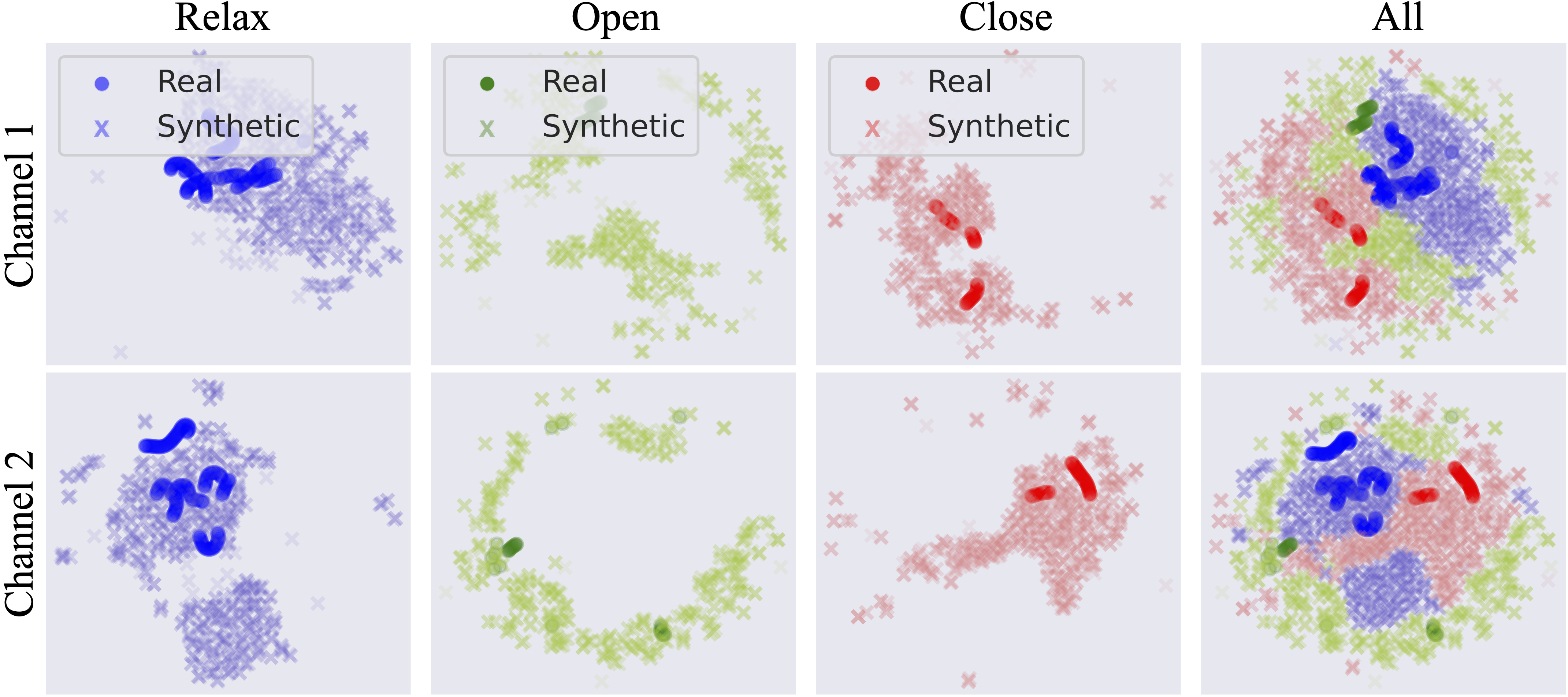}
    \caption{t-SNE visualization. We compare the t-SNE embedding space of the first 2 EMG channels between the synthetic samples and real samples from the same recording of subject S4.}
    \label{fig:tsne}
\end{figure}


We show generated samples for lower-functioning subjects S1, S2, and S3 in Fig.~\ref{fig:samples}. The first 150 steps (1.5s) are the sampled prompts from the limited dataset and are identical between real and synthetic data. For the synthetic samples, ChatEMG generates the last 106 steps (1.06s). When presented with both plots in parallel, without knowing in advance, it is very challenging to identify the synthetic one. There is no significant transition at 1.5s when the sequence switches from real to synthetic. This shows that ChatEMG can capture the characteristics of EMG signals, such as the amplitude, frequency, fluctuation pattern, etc. ChatEMG can also maintain the relative position of different channels very well. We compute the normalized root mean squared error (NRMSE) between the synthetic and real samples across all subjects and intents. The NRMSE for close, open, and relax intents across all subjects are 6\%, 5\%, and 3\%, respectively.

More importantly, ChatEMG not only learns to babble EMG signals by following the previous trends,
but it also learns to reproduce common trends that do not show up in the prompt at all. For example, in the prompt of S1's open sample, there is no indication of the green channel (emg2) going up, but the generated sequence shows such a trend, which turns out to be correct. 

We further visualize the generated samples of subject S4 in a low-dimensional space using \mbox{t-SNE~\cite{van2008visualizing}}, shown in Fig.~\ref{fig:tsne}. For each class, we generate 1000 synthetic samples using the support set and then randomly select 100 samples from the query set of the same recording. We embed each channel of the 256-step EMG sequence separately into a 2D space, and visualize that of the first two channels. We observe that: (1) The synthetic samples of different intents are very separable from each other, meaning that ChatEMG captures distinct patterns of different intents. (2) The embedding space of the generated samples almost always covers the real samples from the query set. This shows that ChatEMG captures the distribution of the test samples correctly. Thus, adding these synthetic samples to the limited support set can improve the intent inferral accuracy on the query set.

\subsection{Integration in Complete Subject Protocol}

We deploy ChatEMG to help an unseen stroke subject complete a functional pick-and-place task using a robotic hand orthosis, as shown in Fig.~\ref{fig:teaser_chatemg}. 
We integrate the pipeline of collecting a limited support set, using ChatEMG to generate synthetic samples, and training Tranformer classifiers within a single hospital session.
Visit our project website for video demonstrations.
This preliminary experiment uses the ChatEMG models trained with data from S1, S2, S4, and S5, and excludes data from the test subject S3. Without adding the synthetic data, classifiers trained with only one open-close-relax motion cannot predict the open intent at all. However, when the classifier is trained with synthetic sample augmentation, S3 can complete multiple rounds of pick-and-place tasks. These qualitative results suggest that the improvement in classification accuracy can translate to the improvement of meaningful daily functional tasks.

\section{Chapter Summary}

ChatEMG addresses the data scarcity issue through synthetic data generation. It allows us to collect only a very small dataset from the new condition/session/subject and expand it with synthetic samples. ChatEMG learns the broad behavior of forearm EMG signals from a vast corpus of previous data while remaining context-specific via prompting. We show that these synthetic samples are classifier-agnostic and can improve the intent inferral accuracy of different types of classifiers. We are the first to deploy an intent classifier trained partially on synthetic data on a hand orthosis to help an unseen stroke subject complete pick-and-place tasks, showing that the improvement in classification accuracy can lead to improvement in meaningful functional tasks.

%% file: Conclusion.tex
\begin{center}
\pagebreak
\vspace*{5\baselineskip}
\textbf{\large Conclusion}
\end{center}

\hspace{10mm} Data is at the center of any AI breakthroughs, and robotics is certainly no exception. While robot learning starts to revolutionize how we develop generalist robots that work for humans, with humans, and are part of humans, data will continue to be the foundation for such advancements. Unfortunately, unlike text, images, or videos that are abundant on the Internet, high-quality robotics data remains sparse and scarce. This thesis is an attempt to address the data sparsity and scarcity issues in robot learning, focusing on two representative domains: tactile manipulation (Part~\ref{part:sparsity}) and rehabilitation robots (Part~\ref{part:scarcity}).

\hspace{10mm} In tactile manipulation, inspired by humans' ability to use tactile sensing under the complete absence of vision, such as effortlessly retrieving a small object from our pockets, we study the data sparsity problem by exploring two tasks, tactile object identification and tactile object retrieval from granular media. In tactile object identification, touch signals such as contact locations and normals are so sparse and local, but we address this challenge by methodically choosing the next action to take that collects the most useful information. We co-train this intelligent and efficient exploration policy with a decision-making policy, and they work together to recognize objects using so few touches that even humans might struggle to do so with the same limited information. We name this co-training framework TANDEM~\cite{xu2022tandem} (Chapter~\ref{chap:tandem}).

\hspace{10mm} We then extend this co-training framework to 3D environments where the data sparsity problem is even more pronounced. 3D objects present more complicated geometry, and we need a better encoder to compute 3D representations from a sequence of sparse tactile data. Built on top of the co-training framework, we propose TANDEM3D~\cite{xu2023tandem3d} (Chapter~\ref{chap:tandem3d}). TANDEM3D further addresses the data sparsity challenge by enabling 6DOF movements of the tactile sensor and is able to discover discriminative points through small-angle adjustments, taking advantage of the tactile finger with all-around sensing coverage. It jointly learns an encoder with the decision-making and exploration policies, which builds the 3D object representation from the contact positions and surface normals acquired via tactile sensing.

\hspace{10mm} Compared to tactile object identification, object retrieval from granular media is undoubtedly a more challenging task, and granular media is such a sensor-deprived environment that vision is completely occluded. In addition to the data sparsity problem that is fundamental to tactile manipulation, this task presents additional challenges such as large uncertainty and significant sensor noise, introduced by the ubiquitous contacts inside granular media. GEOTACT~\cite{xu2024tactile} (Chapter~\ref{chap:geotact}) formulates this task as a model-free reinforcement learning problem and trains it end-to-end in simulation with a curriculum strategy. Our formulation of the learning task action space leads to emergent pushing behaviors that efficiently explore under granular media with sparse tactile signals and help reduce uncertainty. GEOTACT is the first robotic system that can retrieve novel objects completely inside granular media using only touch sensing. 

\hspace{10mm} While data sparsity is a challenge from the angle of the data quality/representation, data scarcity focuses on data quantity, meaning that the sheer amount of data we can collect from physical hardware and humans is very limited. In the second part of the thesis, we switch gears and address the data scarcity issue in rehabilitation robots, which is arguably the most impacted by this challenge, due to the extreme difficulty in collecting biosignals from people with disabilities. Specifically, we study intent inferral for stroke. Widely considered to be a key problem in assistive and rehabilitative robotics, an effective intent inferral mechanism can be an intuitive way to control a robotic device.

\hspace{10mm} Concept drift exacerbates the data scarcity problem in intent inferral, as it requires the training set to incorporate labeled data with as much signal variation as possible. Such data collection comes at a high cost since it is manually labeled, putting a lot of burden on the stroke subjects. SemiEMG~\cite{xu2021learned} (Chapter~\ref{chap:semiemg}) addresses intrasession concept drift through exploiting unlabeled data. It is a disagreement-based semi-supervised algorithm that automatically labels a stream of new unlabeled data in an online fashion.  To our knowledge, this is the first time a semi-supervised learning algorithm has been proposed and used for a hand orthosis based on multimodal ipsilateral sensing. 

\hspace{10mm} While SemiEMG can help adapt to concept drift within the same session, the heuristics to label incoming unlabeled data will generally fail for a new session. In order to address the data scarcity problem by quickly adapting to not only the intrasession but also the intersession drift, we propose MetaEMG~\cite{la2024meta} (Chapter~\ref{chap:metaemg}), a meta-learning (or learning to learn) approach that treats intent inferral as a multitask learning problem with each session or patient being a task. MetaEMG naturally and effectively handles intersession drift, further reduces the training burden on the stroke patients, and mediates the challenge of data scarcity.

\hspace{10mm} Different from both SemiEMG and MetaEMG, which propose machine learning algorithms that work with limited labeled data, reciprocal learning~\cite{xu2025reciprocal} (Chapter~\ref{chap:reciprocal_learning}) addresses the data scarcity challenge by enabling bi-directional learning. This novel paradigm consists of iterative, interwoven stages that alternate between updating machine learning models and guiding human adaptation with the use of augmented visual feedback. It addresses the data scarcity challenge by generating higher-quality training data and reinforcing more distinguishable movement patterns, giving the classifier an easier time to learn.

\hspace{10mm} All the learning paradigms discussed above either try to find a smart workaround with limited data (SemiEMG, MetaEMG) or improve the quality of the limited data (reciprocal learning); however, ChatEMG~\cite{xu2024chatemg} (Chapter~\ref{chap:chatemg}) addresses the data scarcity issue by dreaming up synthetic data. It allows us to collect only a very small dataset from the new condition/session/subject and expand it with synthetic samples. ChatEMG learns the broad behavior of forearm EMG signals from a vast corpus of previous data while remaining context-specific via prompting. We show that these synthetic samples are classifier-agnostic and can improve the intent inferral accuracy of different types of classifiers. We are the first to deploy an intent classifier trained partially on synthetic data on a hand orthosis to help an unseen stroke subject complete pick-and-place tasks, showing that the improvement in classification accuracy can lead to improvement in meaningful functional tasks.

\hspace{10mm} We believe the insights gained in this thesis extend beyond the two domains and are applicable to a broad range of robotics applications facing the fundamental challenges of data sparsity and scarcity. I hope this thesis can contribute meaningfully to the advancement of robotics research, and the advent of truly cognitively and physically intelligent generalist robots.



%% file: references.bib
@misc{openai2024chatgpt,
  author       = {OpenAI},
  title        = {{ChatGPT}},
  year         = {2024},
  howpublished = {\url{https://chat.openai.com}},
  note         = {Accessed: 2025-07-21}
}

@misc{sora2024,
  author       = {OpenAI},
  title        = {Sora},
  year         = {2024},
  howpublished = {\url{https://openai.com/sora}},
  note         = {Video generation model}
}

@misc{claude2023,
  author       = {Anthropic},
  title        = {Claude},
  year         = {2023},
  howpublished = {\url{https://www.anthropic.com/index/introducing-claude}},
  note         = {Large language model}
}

@inproceedings{watkins2019multi,
  title={Multi-modal geometric learning for grasping and manipulation},
  author={Watkins-Valls, David and Varley, Jacob and Allen, Peter},
  booktitle={2019 Intl. Conf. on robotics and automation},
  pages={7339--7345},
  year={2019},
  organization={IEEE}
}

@article{allen1988integrating,
  title={Integrating vision and touch for object recognition tasks},
  author={Allen, Peter K},
  journal={The Intl. Journal of Robotics Research},
  volume={7},
  number={6},
  pages={15--33},
  year={1988},
  publisher={Sage Publications Sage CA: Thousand Oaks, CA}
}

@inproceedings{yu2015shape,
  title={Shape and pose recovery from planar pushing},
  author={Yu, Kuan-Ting and Leonard, John and Rodriguez, Alberto},
  booktitle={2015 IEEE/RSJ Intl. Conf. on Intelligent Robots and Systems},
  pages={1208--1215},
  year={2015},
  organization={IEEE}
}

@article{suresh2020tactile,
  title={Tactile SLAM: Real-time inference of shape and pose from planar pushing},
  author={Suresh, Sudharshan and Bauza, Maria and Yu, Kuan-Ting and Mangelson, Joshua G and Rodriguez, Alberto and Kaess, Michael},
  journal={arXiv preprint arXiv:2011.07044},
  year={2020}
}

@inproceedings{schneider2009object,
  title={Object identification with tactile sensors using bag-of-features},
  author={Schneider, Alexander and Sturm, J{\"u}rgen and Stachniss, Cyrill and Reisert, Marco and Burkhardt, Hans and Burgard, Wolfram},
  booktitle={2009 IEEE/RSJ Intl. Conf. on Intelligent Robots and Systems},
  pages={243--248},
  year={2009},
  organization={IEEE}
}

@article{pezzementi2011tactile,
  title={Tactile-object recognition from appearance information},
  author={Pezzementi, Zachary and Plaku, Erion and Reyda, Caitlin and Hager, Gregory D},
  journal={IEEE Transactions on Robotics},
  volume={27},
  number={3},
  pages={473--487},
  year={2011},
  publisher={IEEE}
}

@inproceedings{xu2013tactile,
  title={Tactile identification of objects using Bayesian exploration},
  author={Xu, Danfei and Loeb, Gerald E and Fishel, Jeremy A},
  booktitle={2013 IEEE Intl. Conf. on Robotics and Automation},
  pages={3056--3061},
  year={2013},
  organization={IEEE}
}

@inproceedings{martinez2013active,
  title={Active contour following to explore object shape with robot touch},
  author={Martinez-Hernandez, Uriel and Metta, Giorgio and Dodd, Tony J and Prescott, Tony J and Natale, Lorenzo and Lepora, Nathan F},
  booktitle={2013 World Haptics Conf. (WHC)},
  pages={341--346},
  year={2013},
  organization={IEEE}
}

@inproceedings{strub2014using,
  title={Using haptics to extract object shape from rotational manipulations},
  author={Strub, Claudius and W{\"o}rg{\"o}tter, Florentin and Ritter, Helge and Sandamirskaya, Yulia},
  booktitle={2014 Intl. Conf. on Intelligent Robots and Systems},
  pages={2179--2186}
}

@inproceedings{schmitz2014tactile,
  title={Tactile object recognition using deep learning and dropout},
  author={Schmitz, Alexander and Bansho, Yusuke and Noda, Kuniaki and Iwata, Hiroyasu and Ogata, Tetsuya and Sugano, Shigeki},
  booktitle={2014 IEEE-RAS Intl. Conf. on Humanoid Robots},
  pages={1044--1050},
}

@article{schulman2017proximal,
  title={Proximal policy optimization algorithms},
  author={Schulman, John and Wolski, Filip and Dhariwal, Prafulla and Radford, Alec and Klimov, Oleg},
  journal={arXiv preprint arXiv:1707.06347},
  year={2017}
}

@article{piacenza2020sensorized,
  title={A sensorized multicurved robot finger with data-driven touch sensing via overlapping light signals},
  author={Piacenza, Pedro and Behrman, Keith and Schifferer, Benedikt and Kymissis, Ioannis and Ciocarlie, Matei},
  journal={IEEE/ASME Transactions on Mechatronics},
  volume={25},
  number={5},
  pages={2416--2427},
  year={2020},
  publisher={IEEE}
}

@inproceedings{johnson2009retrographic,
  title={Retrographic sensing for the measurement of surface texture and shape},
  author={Johnson, Micah K and Adelson, Edward H},
  booktitle={2009 IEEE Conf. on Computer Vision and Pattern Recognition},
  pages={1070--1077},
  year={2009},
  organization={IEEE}
}

@article{lambeta2020digit,
  title={Digit: A novel design for a low-cost compact high-resolution tactile sensor with application to in-hand manipulation},
  author={Lambeta, Mike and Chou, Po-Wei and Tian, Stephen and Yang, Brian and Maloon, Benjamin and Most, Victoria Rose and Stroud, Dave and Santos, Raymond and Byagowi, Ahmad and Kammerer, Gregg and others},
  journal={IEEE Robotics and Automation Letters},
  volume={5},
  number={3},
  pages={3838--3845},
  year={2020},
  publisher={IEEE}
}

@article{meier2011probabilistic,
  title={A probabilistic approach to tactile shape reconstruction},
  author={Meier, Martin and Schopfer, Matthias and Haschke, Robert and Ritter, Helge},
  journal={IEEE Transactions on Robotics},
  volume={27},
  number={3},
  pages={630--635},
  year={2011},
  publisher={IEEE}
}

@inproceedings{bierbaum2009grasp,
  title={Grasp affordances from multi-fingered tactile exploration using dynamic potential fields},
  author={Bierbaum, Alexander and Rambow, Matthias and Asfour, Tamim and Dillmann, R{\"u}diger},
  booktitle={2009 9th IEEE-RAS Intl. Conf. on Humanoid Robots},
  pages={168--174},
  year={2009},
  organization={IEEE}
}

@article{rajeswar2021touch,
  title={Touch-based Curiosity for Sparse-Reward Tasks},
  author={Rajeswar, Sai and Ibrahim, Cyril and Surya, Nitin and Golemo, Florian and Vazquez, David and Courville, Aaron and Pinheiro, Pedro O},
  journal={arXiv preprint arXiv:2104.00442},
  year={2021}
}

@inproceedings{zhang2017active,
  title={Active end-effector pose selection for tactile object recognition through monte carlo tree search},
  author={Zhang, Mabel M and Atanasov, Nikolay and Daniilidis, Kostas},
  booktitle={2017 IEEE/RSJ Intl. Conf. on Intelligent Robots and Systems},
  pages={3258--3265},
  year={2017},
  organization={IEEE}
}

@inproceedings{hebert2013next,
  title={The next best touch for model-based localization},
  author={Hebert, Paul and Howard, Thomas and Hudson, Nicolas and Ma, Jeremy and Burdick, Joel W},
  booktitle={2013 IEEE Intl. Conf. on Robotics and Automation},
  pages={99--106},
  year={2013},
  organization={IEEE}
}

@article{pastor2019using,
  title={Using 3d convolutional neural networks for tactile object recognition with robotic palpation},
  author={Pastor, Francisco and Gandarias, Juan M and Garc{\'\i}a-Cerezo, Alfonso J and G{\'o}mez-de-Gabriel, Jes{\'u}s M},
  journal={Sensors},
  volume={19},
  number={24},
  pages={5356},
  year={2019},
  publisher={Multidisciplinary Digital Publishing Institute}
}

@article{allen1988haptic,
  title={Haptic object recognition using a multi-fingered dextrous hand},
  author={Allen, Peter K and Roberts, Kenneth S},
  year={1988}
}

@article{okamura2001feature,
  title={Feature detection for haptic exploration with robotic fingers},
  author={Okamura, Allison M and Cutkosky, Mark R},
  journal={The Intl. Journal of Robotics Research},
  volume={20},
  number={12},
  pages={925--938},
  year={2001},
  publisher={SAGE Publications}
}

@article{allen1985object,
  title={Object recognition using vision and touch},
  author={Allen, Peter K},
  year={1985}
}

@article{gaston1984tactile,
  title={Tactile recognition and localization using object models: The case of polyhedra on a plane},
  author={Gaston, Peter C and Lozano-Perez, Tomas},
  journal={IEEE transactions on pattern analysis and machine intelligence},
  number={3},
  pages={257--266},
  year={1984},
  publisher={IEEE}
}

@article{skiena1989problems,
  title={Problems in geometric probing},
  author={Skiena, Steven},
  journal={Algorithmica},
  volume={4},
  number={4},
  pages={599--605},
  year={1989},
  publisher={Citeseer}
}

@article{fishel2012bayesian,
  title={Bayesian exploration for intelligent identification of textures},
  author={Fishel, Jeremy A and Loeb, Gerald E},
  journal={Frontiers in neurorobotics},
  volume={6},
  pages={4},
  year={2012},
  publisher={Frontiers}
}

@inproceedings{lepora2013active,
  title={Active touch for robust perception under position uncertainty},
  author={Lepora, Nathan F and Martinez-Hernandez, Uriel and Prescott, Tony J},
  booktitle={2013 IEEE Intl. Conf. on Robotics and Automation},
  pages={3020--3025},
  year={2013},
  organization={IEEE}
}

@article{martinez2017active,
  title={Active sensorimotor control for tactile exploration},
  author={Martinez-Hernandez, Uriel and Dodd, Tony J and Evans, Mathew H and Prescott, Tony J and Lepora, Nathan F},
  journal={Robotics and Autonomous Systems},
  volume={87},
  pages={15--27},
  year={2017},
  publisher={Elsevier}
}

@article{kaboli2017tactile,
  title={A tactile-based framework for active object learning and discrimination using multimodal robotic skin},
  author={Kaboli, Mohsen and Feng, Di and Yao, Kunpeng and Lanillos, Pablo and Cheng, Gordon},
  journal={IEEE Robotics and Automation Letters},
  volume={2},
  number={4},
  pages={2143--2150},
  year={2017},
  publisher={IEEE}
}

@article{kaboli2019tactile,
  title={Tactile-based active object discrimination and target object search in an unknown workspace},
  author={Kaboli, Mohsen and Yao, Kunpeng and Feng, Di and Cheng, Gordon},
  journal={Autonomous Robots},
  volume={43},
  number={1},
  pages={123--152},
  year={2019},
  publisher={Springer}
}

@inproceedings{driess2017active,
  title={Active learning with query paths for tactile object shape exploration},
  author={Driess, Danny and Englert, Peter and Toussaint, Marc},
  booktitle={2017 IEEE/RSJ Intl. Conf. on Intelligent Robots and Systems},
  pages={65--72},
  year={2017},
  organization={IEEE}
}

@article{xu2024chatemg,
  title={ChatEMG: Synthetic data generation to control a robotic hand orthosis for stroke},
  author={Xu, Jingxi and Wang, Runsheng and Shang, Siqi and Chen, Ava and Winterbottom, Lauren and Hsu, To-Liang and Chen, Wenxi and Ahmed, Khondoker and La Rotta, Pedro Leandro and Zhu, Xinyue and others},
  journal={IEEE Robotics and Automation Letters},
  year={2024},
  publisher={IEEE}
}

@ARTICLE{beckerle2017,
  AUTHOR={Beckerle, Philipp and Salvietti, Gionata and Unal, Ramazan and
                  Prattichizzo, Domenico and Rossi, Simone and
                  Castellini, Claudio and Hirche, Sandra and Endo,
                  Satoshi and Amor, Heni Ben and Ciocarlie, Matei and
                  Mastrogiovanni, Fulvio and Argall, Brenna D. and
                  Bianchi, Matteo},   
	 TITLE={A Human–Robot Interaction Perspective on Assistive and
                  Rehabilitation Robotics},      	
JOURNAL={Frontiers in Neurorobotics},      	
VOLUME={11},      
PAGES={24},     	
YEAR={2017},      
}

@inproceedings{xu2022adaptive,
  title={Adaptive Semi-Supervised Intent Inferral to Control a Powered Hand Orthosis for Stroke},
  author={Xu, Jingxi and Meeker, Cassie and Chen, Ava and Winterbottom, Lauren and Fraser, Michaela and Park, Sangwoo and Weber, Lynne M and Miya, Mitchell and Nilsen, Dawn and Stein, Joel and others},
  booktitle={2022 International Conference on Robotics and Automation (ICRA)},
  pages={8097--8103},
  year={2022},
  organization={IEEE}
}

@article{park2018multimodal,
  title={Multimodal sensing and interaction for a robotic hand orthosis},
  author={Park, Sangwoo and Meeker, Cassie and Weber, Lynne M and Bishop, Lauri and Stein, Joel and Ciocarlie, Matei},
  journal={IEEE Robotics and Automation Letters},
  volume={4},
  number={2},
  pages={315--322},
  year={2018},
  publisher={IEEE}
}

@inproceedings{meeker2017emg,
  title={EMG pattern classification to control a hand orthosis for functional grasp assistance after stroke},
  author={Meeker, Cassie and Park, Sangwoo and Bishop, Lauri and Stein, Joel and Ciocarlie, Matei},
  booktitle={2017 international conference on rehabilitation robotics (ICORR)},
  pages={1203--1210},
  year={2017},
  organization={IEEE}
}

@article{chen2022thumb,
  title={Thumb Stabilization and Assistance in a Robotic Hand Orthosis for Post-Stroke Hemiparesis},
  author={Chen, Ava and Winterbottom, Lauren and Park, Sangwoo and Xu, Jingxi and Nilsen, Dawn M and Stein, Joel and Ciocarlie, Matei},
  journal={IEEE Robotics and Automation Letters},
  volume={7},
  number={3},
  pages={8276--8282},
  year={2022},
  publisher={IEEE}
}

@article{vaswani2017attention,
  title={Attention is all you need},
  author={Vaswani, Ashish and Shazeer, Noam and Parmar, Niki and Uszkoreit, Jakob and Jones, Llion and Gomez, Aidan N and Kaiser, {\L}ukasz and Polosukhin, Illia},
  journal={Advances in neural information processing systems},
  volume={30},
  year={2017}
}

@article{cisnal2023interaction,
  title={Interaction with a hand rehabilitation exoskeleton in emg-driven bilateral therapy: Influence of visual biofeedback on the users’ performance},
  author={Cisnal, Ana and Gordaliza, Paula and P{\'e}rez Turiel, Javier and Fraile, Juan Carlos},
  journal={Sensors},
  volume={23},
  number={4},
  pages={2048},
  year={2023},
  publisher={MDPI}
}

@article{zhai2017self,
  title={Self-recalibrating surface EMG pattern recognition for neuroprosthesis control based on convolutional neural network},
  author={Zhai, Xiaolong and Jelfs, Beth and Chan, Rosa HM and Tin, Chung},
  journal={Frontiers in neuroscience},
  volume={11},
  pages={379},
  year={2017},
  publisher={Frontiers Media SA}
}

@article{sensinger2009adaptive,
  title={Adaptive pattern recognition of myoelectric signals: exploration of conceptual framework and practical algorithms},
  author={Sensinger, Jonathon W and Lock, Blair A and Kuiken, Todd A},
  journal={IEEE Transactions on Neural Systems and Rehabilitation Engineering},
  volume={17},
  number={3},
  pages={270--278},
  year={2009},
  publisher={IEEE}
}

@article{amsuss2013self,
  title={Self-correcting pattern recognition system of surface EMG signals for upper limb prosthesis control},
  author={Ams{\"u}ss, Sebastian and Goebel, Peter M and Jiang, Ning and Graimann, Bernhard and Paredes, Liliana and Farina, Dario},
  journal={IEEE Transactions on Biomedical Engineering},
  volume={61},
  number={4},
  pages={1167--1176},
  year={2013},
  publisher={IEEE}
}

@article{chen2013application,
  title={Application of a self-enhancing classification method to electromyography pattern recognition for multifunctional prosthesis control},
  author={Chen, Xinpu and Zhang, Dingguo and Zhu, Xiangyang},
  journal={Journal of neuroengineering and rehabilitation},
  volume={10},
  number={1},
  pages={1--13},
  year={2013},
  publisher={BioMed Central}
}

@article{edwards2016application,
  title={Application of real-time machine learning to myoelectric prosthesis control: A case series in adaptive switching},
  author={Edwards, Ann L and Dawson, Michael R and Hebert, Jacqueline S and Sherstan, Craig and Sutton, Richard S and Chan, K Ming and Pilarski, Patrick M},
  journal={Prosthetics and orthotics international},
  volume={40},
  number={5},
  pages={573--581},
  year={2016},
  publisher={SAGE Publications Sage UK: London, England}
}

@inproceedings{zhang2013adaptation,
  title={An adaptation strategy of using LDA classifier for EMG pattern recognition},
  author={Zhang, Haoshi and Zhao, Yaonan and Yao, Fuan and Xu, Lisheng and Shang, Peng and Li, Guanglin},
  booktitle={2013 35th annual international conference of the IEEE engineering in medicine and biology society (EMBC)},
  pages={4267--4270},
  year={2013},
  organization={IEEE}
}

@inproceedings{he2012adaptive,
  title={Adaptive pattern recognition of myoelectric signal towards practical multifunctional prosthesis control},
  author={He, Jiayuan and Zhang, Dingguo and Zhu, Xiangyang},
  booktitle={Intelligent Robotics and Applications: 5th International Conference, ICIRA 2012, Montreal, QC, Canada, October 3-5, 2012, Proceedings, Part I 5},
  pages={518--525},
  year={2012},
  organization={Springer}
}

@article{liu2015towards,
  title={Towards zero retraining for myoelectric control based on common model component analysis},
  author={Liu, Jianwei and Sheng, Xinjun and Zhang, Dingguo and Jiang, Ning and Zhu, Xiangyang},
  journal={IEEE Transactions on Neural Systems and Rehabilitation Engineering},
  volume={24},
  number={4},
  pages={444--454},
  year={2015},
  publisher={IEEE}
}

@article{goodfellow2014generative,
  title={Generative adversarial nets},
  author={Goodfellow, Ian and Pouget-Abadie, Jean and Mirza, Mehdi and Xu, Bing and Warde-Farley, David and Ozair, Sherjil and Courville, Aaron and Bengio, Yoshua},
  journal={Advances in neural information processing systems},
  volume={27},
  year={2014}
}

@inproceedings{shin2018medical,
  title={Medical image synthesis for data augmentation and anonymization using generative adversarial networks},
  author={Shin, Hoo-Chang and Tenenholtz, Neil A and Rogers, Jameson K and Schwarz, Christopher G and Senjem, Matthew L and Gunter, Jeffrey L and Andriole, Katherine P and Michalski, Mark},
  booktitle={Simulation and Synthesis in Medical Imaging: Third International Workshop, SASHIMI 2018, Held in Conjunction with MICCAI 2018, Granada, Spain, September 16, 2018, Proceedings 3},
  pages={1--11},
  year={2018},
  organization={Springer}
}

@article{hazra2020synsiggan,
  title={SynSigGAN: Generative adversarial networks for synthetic biomedical signal generation},
  author={Hazra, Debapriya and Byun, Yung-Cheol},
  journal={Biology},
  volume={9},
  number={12},
  pages={441},
  year={2020},
  publisher={MDPI}
}

@article{anicet2020parkinson,
  title={Parkinson’s disease EMG data augmentation and simulation with DCGANs and style transfer},
  author={Anicet Zanini, Rafael and Luna Colombini, Esther},
  journal={Sensors},
  volume={20},
  number={9},
  pages={2605},
  year={2020},
  publisher={MDPI}
}

@article{coelho2023novel,
  title={A novel sEMG data augmentation based on WGAN-GP},
  author={Coelho, Fabricio and Pinto, Milena F and Melo, Aur{\'e}lio G and Ramos, Gabryel S and Marcato, Andr{\'e} LM},
  journal={Computer Methods in Biomechanics and Biomedical Engineering},
  volume={26},
  number={9},
  pages={1008--1017},
  year={2023},
  publisher={Taylor \& Francis}
}

@article{bird2021synthetic,
  title={Synthetic biological signals machine-generated by GPT-2 improve the classification of EEG and EMG through data augmentation},
  author={Bird, Jordan J and Pritchard, Michael and Fratini, Antonio and Ek{\'a}rt, Anik{\'o} and Faria, Diego R},
  journal={IEEE Robotics and Automation Letters},
  volume={6},
  number={2},
  pages={3498--3504},
  year={2021},
  publisher={IEEE}
}

@article{van2008visualizing,
  title={Visualizing data using t-SNE.},
  author={Van der Maaten, Laurens and Hinton, Geoffrey},
  journal={Journal of machine learning research},
  volume={9},
  number={11},
  year={2008}
}

@article{tang2018surface,
  title={Surface electromyographic examination of poststroke neuromuscular changes in proximal and distal muscles using clustering index analysis},
  author={Tang, Weidi and Zhang, Xu and Tang, Xiao and Cao, Shuai and Gao, Xiaoping and Chen, Xiang},
  journal={Frontiers in Neurology},
  volume={8},
  pages={731},
  year={2018},
  publisher={Frontiers Media SA}
}

@article{la2024meta,
  title={Meta-Learning for Fast Adaptation in Intent Inferral on a Robotic Hand Orthosis for Stroke},
  author={La Rotta, Pedro Leandro and Xu, Jingxi and Chen, Ava and Winterbottom, Lauren and Chen, Wenxi and Nilsen, Dawn and Stein, Joel and Ciocarlie, Matei},
  journal={arXiv preprint arXiv:2403.13147},
  year={2024}
}

@article{kyranou2018causes,
  title={Causes of performance degradation in non-invasive electromyographic pattern recognition in upper limb prostheses},
  author={Kyranou, Iris and Vijayakumar, Sethu and Erden, Mustafa Suphi},
  journal={Frontiers in neurorobotics},
  volume={12},
  pages={58},
  year={2018},
  publisher={Frontiers Media SA}
}

@article{piacentino2021generating,
  title={Generating synthetic ecgs using gans for anonymizing healthcare data},
  author={Piacentino, Esteban and Guarner, Alvaro and Angulo, Cecilio},
  journal={Electronics},
  volume={10},
  number={4},
  pages={389},
  year={2021},
  publisher={MDPI}
}

@article{festag2022generative,
  title={Generative adversarial networks for biomedical time series forecasting and imputation},
  author={Festag, Sven and Denzler, Joachim and Spreckelsen, Cord},
  journal={Journal of Biomedical Informatics},
  volume={129},
  pages={104058},
  year={2022},
  publisher={Elsevier}
}

@article{yang2023ts,
  title={Ts-gan: Time-series gan for sensor-based health data augmentation},
  author={Yang, Zhenyu and Li, Yantao and Zhou, Gang},
  journal={ACM Transactions on Computing for Healthcare},
  volume={4},
  number={2},
  pages={1--21},
  year={2023},
  publisher={ACM New York, NY}
}

@inproceedings{xu2021learned,
title={How Are Learned Perception-Based Controllers Impacted by the Limits of Robust Control?},
author={Xu, Jingxi and Lee, Bruce and Matni, Nikolai and Jayaraman, Dinesh},
booktitle={Learning for Dynamics and Control},
pages={954--966},
year={2021},
organization={PMLR}
}

@inproceedings{xu2023tandem3d,
  title={Tandem3d: Active tactile exploration for 3d object recognition},
  author={Xu, Jingxi and Lin, Han and Song, Shuran and Ciocarlie, Matei},
  booktitle={2023 IEEE International Conference on Robotics and Automation (ICRA)},
  pages={10401--10407},
  year={2023},
  organization={IEEE}
}

@article{park2019,
  title   = "Multimodal Sensing and Interaction for a Robotic Hand Orthosis",
   author  = "Park, Sangwoo and Meeker, Cassie and Weber, Lynne M and Bishop,
              Lauri and Stein, Joel and Ciocarlie, Matei",
   journal = "IEEE Robotics and Automation Letters",
   volume  =  4,
   number  =  2,
   pages   = "315--322",
   year    =  2019
 }

@article{xu2024tactile,
  title={Tactile-based Object Retrieval From Granular Media},
  author={Xu, Jingxi and Jia, Yinsen and Yang, Dongxiao and Meng, Patrick and Zhu, Xinyue and Guo, Zihan and Song, Shuran and Ciocarlie, Matei},
  journal={Autonomous Robots},
  year={2025}
}

@inproceedings{akinola2021dynamic,
  title={Dynamic grasping with reachability and motion awareness},
  author={Akinola*, Iretiayo and Xu*, Jingxi and Song, Shuran and Allen, Peter K},
  booktitle={2021 IEEE/RSJ International Conference on Intelligent Robots and Systems (IROS)},
  pages={9422--9429},
  year={2021},
  organization={IEEE}
}

@inproceedings{akinola2020accelerated,
  title={Accelerated robot learning via human brain signals},
  author={Akinola, Iretiayo and Wang, Zizhao and Shi, Junyao and He, Xiaomin and Lapborisuth, Pawan and Xu, Jingxi and Watkins-Valls, David and Sajda, Paul and Allen, Peter},
  booktitle={2020 IEEE international conference on robotics and automation (ICRA)},
  pages={3799--3805},
  year={2020},
  organization={IEEE}
}

@inproceedings{watkins2020learning,
  title={Learning your way without map or compass: Panoramic target driven visual navigation},
  author={Watkins-Valls, David and Xu, Jingxi and Waytowich, Nicholas and Allen, Peter},
  booktitle={2020 IEEE/RSJ International Conference on Intelligent Robots and Systems (IROS)},
  pages={5816--5823},
  year={2020},
  organization={IEEE}
}

@article{ha2020learning,
  title={Learning a decentralized multi-arm motion planner},
  author={Ha, Huy and Xu, Jingxi and Song, Shuran},
  journal={2020 Conference on Robot Learning (CoRL)},
  year={2020}
}

@article{xu2022tandem,
  title={Tandem: Learning joint exploration and decision making with tactile sensors},
  author={Xu, Jingxi and Song, Shuran and Ciocarlie, Matei},
  journal={IEEE Robotics and Automation Letters},
  volume={7},
  number={4},
  pages={10391--10398},
  year={2022},
  publisher={IEEE}
}

@inproceedings{chang2024investigation,
  title={An investigation of multi-feature extraction and super-resolution with fast microphone arrays},
  author={Chang, Eric T and Wang, Runsheng and Ballentine, Peter and Xu, Jingxi and Smith, Trey and Coltin, Brian and Kymissis, Ioannis and Ciocarlie, Matei},
  booktitle={2024 IEEE International Conference on Robotics and Automation (ICRA)},
  pages={3388--3394},
  year={2024},
  organization={IEEE}
}

@article{chen2024volitional,
  title={Volitional Control of the Paretic Hand Post-Stroke Increases Finger Stiffness and Resistance to Robot-Assisted Movement},
  author={Chen, Ava and Lee, Katelyn and Winterbottom, Lauren and Xu, Jingxi and Lee, Connor and Munger, Grace and Deli-Ivanov, Alexandra and Nilsen, Dawn M and Stein, Joel and Ciocarlie, Matei},
  journal={2024 IEEE International Conference on Biomedical Robotics and Biomechatronics (BioRob)},
  year={2024}
}

@article{jia2022autonomous,
  title={Autonomous Tactile Localization and Mapping of Objects Buried in Granular Materials},
  author={Jia, Shengxin and Zhang, Lionel and Santos, Veronica J},
  journal={IEEE Robotics and Automation Letters},
  volume={7},
  number={4},
  pages={9953--9960},
  year={2022},
  publisher={IEEE}
}

@article{jia2021tactile,
  title={Tactile perception for teleoperated robotic exploration within granular media},
  author={Jia, Shengxin and Santos, Veronica J},
  journal={ACM Transactions on Human-Robot Interaction (THRI)},
  volume={10},
  number={4},
  pages={1--27},
  year={2021},
  publisher={ACM New York, NY, USA}
}

@inproceedings{patel2021digger,
  title={Digger finger: Gelsight tactile sensor for object identification inside granular media},
  author={Patel, Radhen and Ouyang, Rui and Romero, Branden and Adelson, Edward},
  booktitle={Experimental Robotics: The 17th International Symposium},
  pages={105--115},
  year={2021},
  organization={Springer}
}

@article{millard2023granular,
  title={Granular Gym: High Performance Simulation for Robotic Tasks with Granular Materials},
  author={Millard, David and Pastor, Daniel and Bowkett, Joseph and Backes, Paul and Sukhatme, Gaurav S},
  journal={arXiv preprint arXiv:2306.01369},
  year={2023}
}

@article{thompson2022lammps,
  title={LAMMPS-a flexible simulation tool for particle-based materials modeling at the atomic, meso, and continuum scales},
  author={Thompson, Aidan P and Aktulga, H Metin and Berger, Richard and Bolintineanu, Dan S and Brown, W Michael and Crozier, Paul S and in't Veld, Pieter J and Kohlmeyer, Axel and Moore, Stan G and Nguyen, Trung Dac and others},
  journal={Computer Physics Communications},
  volume={271},
  pages={108171},
  year={2022},
  publisher={Elsevier}
}

@article{xian2023fluidlab,
  title={Fluidlab: A differentiable environment for benchmarking complex fluid manipulation},
  author={Xian, Zhou and Zhu, Bo and Xu, Zhenjia and Tung, Hsiao-Yu and Torralba, Antonio and Fragkiadaki, Katerina and Gan, Chuang},
  journal={arXiv preprint arXiv:2303.02346},
  year={2023}
}

@article{kloss2012models,
  title={Models, algorithms and validation for opensource DEM and CFD--DEM},
  author={Kloss, Christoph and Goniva, Christoph and Hager, Alice and Amberger, Stefan and Pirker, Stefan},
  journal={Progress in Computational Fluid Dynamics, an International Journal},
  volume={12},
  number={2-3},
  pages={140--152},
  year={2012},
  publisher={Inderscience Publishers}
}

@article{fang2021chrono,
  title={Chrono:: GPU: An open-source simulation package for granular dynamics using the discrete element method},
  author={Fang, Luning and Zhang, Ruochun and Vanden Heuvel, Colin and Serban, Radu and Negrut, Dan},
  journal={Processes},
  volume={9},
  number={10},
  pages={1813},
  year={2021},
  publisher={MDPI}
}

@article{servin2021multiscale,
  title={A multiscale model of terrain dynamics for real-time earthmoving simulation},
  author={Servin, Martin and Berglund, Tomas and Nystedt, Samuel},
  journal={Advanced Modeling and Simulation in Engineering Sciences},
  volume={8},
  number={1},
  pages={1--35},
  year={2021},
  publisher={SpringerOpen}
}

@misc{ansys,
  title = {Ansys Rocky — Particle Dynamics Simulation Software},
  howpublished = {\url{https://www.ansys.com/products/fluids/ansys-rocky}},
  note = {Accessed: 2023-11-23}
}

@inproceedings{haeri2020efficient,
  title={Efficient numerical methods for accurate modeling of soil cutting operations},
  author={Haeri, Amin and Tremblay, Dominique and Skonieczny, Krzysztof and Holz, Daniel and Teichmann, Marek},
  booktitle={ISARC. Proceedings of the International Symposium on Automation and Robotics in Construction},
  volume={37},
  pages={608--615},
  year={2020},
  organization={IAARC Publications}
}

@article{hu2019difftaichi,
  title={Difftaichi: Differentiable programming for physical simulation},
  author={Hu, Yuanming and Anderson, Luke and Li, Tzu-Mao and Sun, Qi and Carr, Nathan and Ragan-Kelley, Jonathan and Durand, Fr{\'e}do},
  journal={arXiv preprint arXiv:1910.00935},
  year={2019}
}

@article{yuan2017gelsight,
  title={Gelsight: High-resolution robot tactile sensors for estimating geometry and force},
  author={Yuan, Wenzhen and Dong, Siyuan and Adelson, Edward H},
  journal={Sensors},
  volume={17},
  number={12},
  pages={2762},
  year={2017},
  publisher={MDPI}
}

@inproceedings{dong2017improved,
  title={Improved gelsight tactile sensor for measuring geometry and slip},
  author={Dong, Siyuan and Yuan, Wenzhen and Adelson, Edward H},
  booktitle={2017 IEEE/RSJ International Conference on Intelligent Robots and Systems (IROS)},
  pages={137--144},
  year={2017},
  organization={IEEE}
}

@inproceedings{sarata2004trajectory,
  title={Trajectory arrangement based on resistance force and shape of pile at scooping motion},
  author={Sarata, Shigeru and Osumi, Hisashi and Kawai, Yoshihiro and Tomita, Fumiaki},
  booktitle={IEEE International Conference on Robotics and Automation, 2004. Proceedings. ICRA'04. 2004},
  volume={4},
  pages={3488--3493},
  year={2004},
  organization={IEEE}
}

@inproceedings{wu2020squirl,
  title={Squirl: Robust and efficient learning from video demonstration of long-horizon robotic manipulation tasks},
  author={Wu, Bohan and Xu, Feng and He, Zhanpeng and Gupta, Abhi and Allen, Peter K},
  booktitle={2020 IEEE/RSJ International Conference on Intelligent Robots and Systems (IROS)},
  pages={9720--9727},
  year={2020},
  organization={IEEE}
}

@article{yamaguchi2015pouring,
  title={Pouring skills with planning and learning modeled from human demonstrations},
  author={Yamaguchi, Akihiko and Atkeson, Christopher G and Ogasawara, Tsukasa},
  journal={International Journal of Humanoid Robotics},
  volume={12},
  number={03},
  pages={1550030},
  year={2015},
  publisher={World Scientific}
}

@inproceedings{matl2020inferring,
  title={Inferring the material properties of granular media for robotic tasks},
  author={Matl, Carolyn and Narang, Yashraj and Bajcsy, Ruzena and Ramos, Fabio and Fox, Dieter},
  booktitle={2020 ieee international conference on robotics and automation (icra)},
  pages={2770--2777},
  year={2020},
  organization={IEEE}
}

@article{kobayakawa2020interaction,
  title={Interaction between dry granular materials and an inclined plate (comparison between large-scale DEM simulation and three-dimensional wedge model)},
  author={Kobayakawa, Murino and Miyai, Shinichiro and Tsuji, Takuya and Tanaka, Toshitsugu},
  journal={Journal of Terramechanics},
  volume={90},
  pages={3--10},
  year={2020},
  publisher={Elsevier}
}

@article{swick1988model,
  title={A model for predicting soil-tool interaction},
  author={Swick, WC and Perumpral, JV},
  journal={Journal of Terramechanics},
  volume={25},
  number={1},
  pages={43--56},
  year={1988},
  publisher={Elsevier}
}

@article{gravish2014force,
  title={Force and flow at the onset of drag in plowed granular media},
  author={Gravish, Nick and Umbanhowar, Paul B and Goldman, Daniel I},
  journal={Physical Review E},
  volume={89},
  number={4},
  pages={042202},
  year={2014},
  publisher={APS}
}

@article{li2013terradynamics,
  title={A terradynamics of legged locomotion on granular media},
  author={Li, Chen and Zhang, Tingnan and Goldman, Daniel I},
  journal={science},
  volume={339},
  number={6126},
  pages={1408--1412},
  year={2013},
  publisher={American Association for the Advancement of Science}
}

@inproceedings{hauser2016friction,
  title={Friction and damping of a compliant foot based on granular jamming for legged robots},
  author={Hauser, Simon and Eckert, Peter and Tuleu, Alexandre and Ijspeert, Auke},
  booktitle={2016 6th IEEE international conference on biomedical robotics and biomechatronics (BioRob)},
  pages={1160--1165},
  year={2016},
  organization={Ieee}
}

@article{tuomainen2022manipulation,
  title={Manipulation of granular materials by learning particle interactions},
  author={Tuomainen, Neea and Blanco-Mulero, David and Kyrki, Ville},
  journal={IEEE Robotics and Automation Letters},
  volume={7},
  number={2},
  pages={5663--5670},
  year={2022},
  publisher={IEEE}
}

@article{zhu2023few,
  title={Few-shot Adaptation for Manipulating Granular Materials Under Domain Shift},
  author={Zhu, Yifan and Thangeda, Pranay and Ornik, Melkior and Hauser, Kris},
  journal={arXiv preprint arXiv:2303.02893},
  year={2023}
}

@inproceedings{zhu2019data,
  title={A data-driven approach for fast simulation of robot locomotion on granular media},
  author={Zhu, Yifan and Abdulmajeid, Laith and Hauser, Kris},
  booktitle={2019 international conference on robotics and automation (ICRA)},
  pages={7653--7659},
  year={2019},
  organization={IEEE}
}

@inproceedings{brown2020soft,
  title={Soft haptic interface based on vibration and particle jamming},
  author={Brown, Joshua P and Farkhatdinov, Ildar},
  booktitle={2020 IEEE Haptics Symposium (HAPTICS)},
  pages={1--6},
  year={2020},
  organization={IEEE}
}

@inproceedings{stanley2013haptic,
  title={Haptic jamming: A deformable geometry, variable stiffness tactile display using pneumatics and particle jamming},
  author={Stanley, Andrew A and Gwilliam, James C and Okamura, Allison M},
  booktitle={2013 World Haptics Conference (WHC)},
  pages={25--30},
  year={2013},
  organization={IEEE}
}

@inproceedings{cheng2012design,
  title={Design and analysis of a robust, low-cost, highly articulated manipulator enabled by jamming of granular media},
  author={Cheng, Nadia G and Lobovsky, Maxim B and Keating, Steven J and Setapen, Adam M and Gero, Katy I and Hosoi, Anette E and Iagnemma, Karl D},
  booktitle={2012 IEEE international conference on robotics and automation},
  pages={4328--4333},
  year={2012},
  organization={IEEE}
}

@inproceedings{karimi2020boundary,
  title={A boundary-constrained swarm robot with granular jamming},
  author={Karimi, Mohammad Amin and Alizadehyazdi, Vahid and Busque, Bruno-Pier and Jaeger, Heinrich M and Spenko, Matthew},
  booktitle={2020 3rd IEEE International Conference on Soft Robotics (RoboSoft)},
  pages={291--296},
  year={2020},
  organization={IEEE}
}

@article{brown2010universal,
  title={Universal robotic gripper based on the jamming of granular material},
  author={Brown, Eric and Rodenberg, Nicholas and Amend, John and Mozeika, Annan and Steltz, Erik and Zakin, Mitchell R and Lipson, Hod and Jaeger, Heinrich M},
  journal={Proceedings of the National Academy of Sciences},
  volume={107},
  number={44},
  pages={18809--18814},
  year={2010},
  publisher={National Acad Sciences}
}

@phdthesis{clarke2019robot,
  title={Robot learning for manipulation of granular materials using vision and sound},
  author={Clarke, Samuel},
  year={2019},
  school={Master’s thesis, Carnegie Mellon University Pittsburgh}
}

@article{clarke2018learning,
  title={Learning audio feedback for estimating amount and flow of granular material},
  author={Clarke, Samuel and Rhodes, Travers and Atkeson, Christopher G and Kroemer, Oliver},
  journal={Proceedings of Machine Learning Research},
  volume={87},
  year={2018}
}

@article{johnson2011microgeometry,
  title={Microgeometry capture using an elastomeric sensor},
  author={Johnson, Micah K and Cole, Forrester and Raj, Alvin and Adelson, Edward H},
  journal={ACM Transactions on Graphics (TOG)},
  volume={30},
  number={4},
  pages={1--8},
  year={2011},
  publisher={ACM New York, NY, USA}
}

@article{zhang2023grains,
  title={GRAINS: Proximity Sensing of Objects in Granular Materials},
  author={Zhang, Zeqing and Jia, Ruixing and Yan, Youcan and Han, Ruihua and Lin, Shijie and Jiang, Qian and Zhang, Liangjun and Pan, Jia},
  journal={arXiv preprint arXiv:2307.05935},
  year={2023}
}

@article{fakhri2023systematic,
  title={A Systematic Approach to Stable Grasping of a Two-Finger Robotic Hand Activated by Jamming of Granular Media},
  author={Fakhri, Osamah and Youssef, George and Nacy, Somer and Jameel Al-Tamimi, Adnan Naji and Hussein, O and Abbood, Wisam T and Abdullah, Oday Ibraheem and Al-Karkhi, Nazar Kais},
  journal={Electronics},
  volume={12},
  number={8},
  pages={1902},
  year={2023},
  publisher={MDPI}
}

@article{makoviychuk2021isaac,
  title={Isaac gym: High performance gpu-based physics simulation for robot learning},
  author={Makoviychuk, Viktor and Wawrzyniak, Lukasz and Guo, Yunrong and Lu, Michelle and Storey, Kier and Macklin, Miles and Hoeller, David and Rudin, Nikita and Allshire, Arthur and Handa, Ankur and others},
  journal={arXiv preprint arXiv:2108.10470},
  year={2021}
}

@inproceedings{zeng2018learning,
  title={Learning synergies between pushing and grasping with self-supervised deep reinforcement learning},
  author={Zeng, Andy and Song, Shuran and Welker, Stefan and Lee, Johnny and Rodriguez, Alberto and Funkhouser, Thomas},
  booktitle={2018 IEEE/RSJ International Conference on Intelligent Robots and Systems (IROS)},
  pages={4238--4245},
  year={2018},
  organization={IEEE}
}

@inproceedings{dogar2010push,
  title={Push-grasping with dexterous hands: Mechanics and a method},
  author={Dogar, Mehmet R and Srinivasa, Siddhartha S},
  booktitle={2010 IEEE/RSJ International Conference on Intelligent Robots and Systems},
  pages={2123--2130},
  year={2010},
  organization={IEEE}
}

@article{mason1986mechanics,
  title={Mechanics and planning of manipulator pushing operations},
  author={Mason, Matthew T},
  journal={The International Journal of Robotics Research},
  volume={5},
  number={3},
  pages={53--71},
  year={1986},
  publisher={Sage Publications Sage CA: Thousand Oaks, CA}
}

@article{dogar2012planning,
  title={A planning framework for non-prehensile manipulation under clutter and uncertainty},
  author={Dogar, Mehmet R and Srinivasa, Siddhartha S},
  journal={Autonomous Robots},
  volume={33},
  pages={217--236},
  year={2012},
  publisher={Springer}
}

@article{brost1988automatic,
  title={Automatic grasp planning in the presence of uncertainty},
  author={Brost, Randy C},
  journal={The International Journal of Robotics Research},
  volume={7},
  number={1},
  pages={3--17},
  year={1988},
  publisher={Sage Publications Sage CA: Thousand Oaks, CA}
}

@inproceedings{omrvcen2009autonomous,
  title={Autonomous acquisition of pushing actions to support object grasping with a humanoid robot},
  author={Omr{\v{c}}en, Damir and B{\"o}ge, Christian and Asfour, Tamim and Ude, Ale{\v{s}} and Dillmann, R{\"u}diger},
  booktitle={2009 9th IEEE-RAS International Conference on Humanoid Robots},
  pages={277--283},
  year={2009},
  organization={IEEE}
}

@inproceedings{clavera2017policy,
  title={Policy transfer via modularity and reward guiding},
  author={Clavera, Ignasi and Held, David and Abbeel, Pieter},
  booktitle={2017 IEEE/RSJ International Conference on Intelligent Robots and Systems (IROS)},
  pages={1537--1544},
  year={2017},
  organization={IEEE}
}

@inproceedings{boularias2015learning,
  title={Learning to manipulate unknown objects in clutter by reinforcement},
  author={Boularias, Abdeslam and Bagnell, James and Stentz, Anthony},
  booktitle={Proceedings of the AAAI Conference on Artificial Intelligence},
  volume={29},
  number={1},
  year={2015}
}

@article{yang2021collaborative,
  title={Collaborative pushing and grasping of tightly stacked objects via deep reinforcement learning},
  author={Yang, Yuxiang and Ni, Zhihao and Gao, Mingyu and Zhang, Jing and Tao, Dacheng},
  journal={IEEE/CAA Journal of Automatica Sinica},
  volume={9},
  number={1},
  pages={135--145},
  year={2021},
  publisher={IEEE}
}

@inproceedings{deng2019deep,
  title={Deep reinforcement learning for robotic pushing and picking in cluttered environment},
  author={Deng, Yuhong and Guo, Xiaofeng and Wei, Yixuan and Lu, Kai and Fang, Bin and Guo, Di and Liu, Huaping and Sun, Fuchun},
  booktitle={2019 IEEE/RSJ International Conference on Intelligent Robots and Systems (IROS)},
  pages={619--626},
  year={2019},
  organization={Ieee}
}

@inproceedings{tang2021learning,
  title={Learning collaborative pushing and grasping policies in dense clutter},
  author={Tang, Bingjie and Corsaro, Matthew and Konidaris, George and Nikolaidis, Stefanos and Tellex, Stefanie},
  booktitle={2021 IEEE International Conference on Robotics and Automation (ICRA)},
  pages={6177--6184},
  year={2021},
  organization={IEEE}
}

@article{yu2023novel,
  title={A Novel Robotic Pushing and Grasping Method Based on Vision Transformer and Convolution},
  author={Yu, Sheng and Zhai, Di-Hua and Xia, Yuanqing},
  journal={IEEE Transactions on Neural Networks and Learning Systems},
  year={2023},
  publisher={IEEE}
}

@article{hochreiter1997long,
  title={Long short-term memory},
  author={Hochreiter, Sepp and Schmidhuber, J{\"u}rgen},
  journal={Neural computation},
  volume={9},
  number={8},
  pages={1735--1780},
  year={1997},
  publisher={MIT press}
}

@article{tripura2022role,
  title={Role of shape on the forces on an intruder moving through a dense granular medium},
  author={Tripura, Bitang Kwrung and Kumar, Sonu and Anki Reddy, K and Talbot, Julian},
  journal={Particulate Science and Technology},
  volume={40},
  number={6},
  pages={651--661},
  year={2022},
  publisher={Taylor \& Francis}
}

@article{andrychowicz2020matters,
  title={What matters in on-policy reinforcement learning? a large-scale empirical study},
  author={Andrychowicz, Marcin and Raichuk, Anton and Sta{\'n}czyk, Piotr and Orsini, Manu and Girgin, Sertan and Marinier, Raphael and Hussenot, L{\'e}onard and Geist, Matthieu and Pietquin, Olivier and Michalski, Marcin and others},
  journal={arXiv preprint arXiv:2006.05990},
  year={2020}
}

@inproceedings{xu2025reciprocal,
  title={Reciprocal learning of intent inferral with augmented visual feedback for stroke},
  author={Xu, Jingxi and Chen, Ava and Winterbottom, Lauren and Palacios, Joaquin and Chivukula, Preethika and Nilsen, Dawn M and Stein, Joel and Ciocarlie, Matei},
  booktitle={2025 International Conference On Rehabilitation Robotics (ICORR)},
  pages={1512--1517},
  year={2025},
  organization={IEEE}
}

@article{powell2013,
  title={A training strategy for learning pattern recognition control for myoelectric prostheses},
  author={Powell, Michael A and Thakor, Nitish V},
  journal={Journal of prosthetics and orthotics: JPO},
  volume={25},
  number={1},
  pages={30},
  year={2013},
}

@article{castellini2009,
  title={Surface EMG in advanced hand prosthetics},
  author={Castellini, Claudio and van der Smagt, Patrick},
  journal={Biological cybernetics},
  volume={100},
  number={1},
  pages={35--47},
  year={2009},
  publisher={Springer}
}

@article{lee2011,
  title={Subject-specific myoelectric pattern classification of functional hand movements for stroke survivors},
  author={Lee, Sang Wook and Wilson, Kristin M and Lock, Blair A and Kamper, Derek G},
  journal={IEEE Transactions on Neural Systems and Rehabilitation Engineering},
  volume={19},
  number={5},
  pages={558--566},
  year={2011}
}

@article{zhou2010semi,
  title={Semi-supervised learning by disagreement},
  author={Zhou, Zhi-Hua and Li, Ming},
  journal={Knowledge and Information Systems},
  volume={24},
  number={3},
  pages={415--439},
  year={2010},
  publisher={Springer}
}

@inproceedings{blum1998combining,
  title={Combining labeled and unlabeled data with co-training},
  author={Blum, Avrim and Mitchell, Tom},
  booktitle={Proceedings of the eleventh annual conference on Computational learning theory},
  pages={92--100},
  year={1998}
}

@article{kolter2007dynamic,
  title={Dynamic weighted majority: An ensemble method for drifting concepts},
  author={Kolter, J Zico and Maloof, Marcus A},
  journal={Journal of Machine Learning Research},
  volume={8},
  number={Dec},
  pages={2755--2790},
  year={2007}
}

@inproceedings{javed2005online,
  title={Online detection and classification of moving objects using progressively improving detectors},
  author={Javed, Omar and Ali, Saad and Shah, Mubarak},
  booktitle={2005 IEEE Computer Society Conference on Computer Vision and Pattern Recognition (CVPR'05)},
  volume={1},
  pages={696--701},
  year={2005},
  organization={IEEE}
}

@inproceedings{jain2012improving,
  title={Improving long term myoelectric decoding, using an adaptive classifier with label correction},
  author={Jain, Sarthak and Singhal, Girish and Smith, Ryan J and Kaliki, Rahul and Thakor, Nitish},
  booktitle={2012 4th IEEE RAS \& EMBS International Conference on Biomedical Robotics and Biomechatronics (BioRob)},
  pages={532--537},
  year={2012},
  organization={IEEE}
}

@inproceedings{jiang2012semi,
  title={A semi-supervised ensemble learning algorithm},
  author={Jiang, Zhen and Zhang, Shiyong},
  booktitle={2012 IEEE 2nd International Conference on Cloud Computing and Intelligence Systems},
  volume={2},
  pages={913--918},
  year={2012},
  organization={IEEE}
}

@inproceedings{wang2006exploiting,
  title={Exploiting ensemble method in semi-supervised learning},
  author={Wang, Jiao and Luo, Si-wei},
  booktitle={2006 International Conference on Machine Learning and Cybernetics},
  pages={1104--1107},
  year={2006},
  organization={IEEE}
}

@article{miller2012,
  title={Involuntary paretic wrist/finger flexion forces and EMG increase with shoulder abduction load in individuals with chronic stroke},
  author={Miller, Laura C and Dewald, Julius PA},
  journal={Clinical Neurophysiology},
  volume={123},
  number={6},
  pages={1216--1225},
  year={2012},
  publisher={Elsevier}
}

@article{zhou2005tri,
  title={Tri-training: Exploiting unlabeled data using three classifiers},
  author={Zhou, Zhi-Hua and Li, Ming},
  journal={IEEE Transactions on knowledge and Data Engineering},
  volume={17},
  number={11},
  pages={1529--1541},
  year={2005},
  publisher={IEEE}
}

@article{riley2002changes,
  title={Changes in upper limb joint torque patterns and EMG signals with fatigue following a stroke},
  author={Riley, NA and Bilodeau, M},
  journal={Disability and rehabilitation},
  volume={24},
  number={18},
  pages={961--969},
  year={2002},
  publisher={Taylor \& Francis}
}

@inproceedings{finn2017model,
  title={Model-agnostic meta-learning for fast adaptation of deep networks},
  author={Finn, Chelsea and Abbeel, Pieter and Levine, Sergey},
  booktitle={International conference on machine learning},
  pages={1126--1135},
  year={2017},
  organization={PMLR}
}

@article{banluesombatkul2020metasleeplearner,
  title={MetaSleepLearner: A pilot study on fast adaptation of bio-signals-based sleep stage classifier to new individual subject using meta-learning},
  author={Banluesombatkul, Nannapas and Ouppaphan, Pichayoot and Leelaarporn, Pitshaporn and Lakhan, Payongkit and Chaitusaney, Busarakum and Jaimchariyatam, Nattapong and Chuangsuwanich, Ekapol and Chen, Wei and Phan, Huy and Dilokthanakul, Nat and others},
  journal={IEEE Journal of Biomedical and Health Informatics},
  volume={25},
  number={6},
  pages={1949--1963},
  year={2020},
  publisher={IEEE}
}

@inproceedings{prorokovic2020meta,
  title={Meta-learning for recalibration of EMG-based upper limb prostheses},
  author={Prorokovi{\'c}, Krsto and Wand, Michael and Schmidhuber, J{\"u}rgen},
  booktitle={4th Lifelong Machine Learning Workshop at ICML 2020},
  year={2020}
}

@inproceedings{prorokovic2019,
  author={Proroković, Krsto and Wand, Michael and Schultz, Tanja and Schmidhuber, Jürgen},
  booktitle={2019 IEEE Global Conference on Signal and Information Processing (GlobalSIP)}, 
  title={Adaptation of an EMG-Based Speech Recognizer via Meta-Learning}, 
  year={2019},
  volume={},
  number={},
  pages={1-5},
  doi={10.1109/GlobalSIP45357.2019.8969231}}

@incollection{wilcoxon1992individual,
  title={Individual comparisons by ranking methods},
  author={Wilcoxon, Frank},
  booktitle={Breakthroughs in statistics},
  pages={196--202},
  year={1992},
  publisher={Springer}
}

@article{qi2017pointnet++,
  title={Pointnet++: Deep hierarchical feature learning on point sets in a metric space},
  author={Qi, Charles Ruizhongtai and Yi, Li and Su, Hao and Guibas, Leonidas J},
  journal={Advances in neural information processing systems},
  volume={30},
  year={2017}
}

@article{calli2017yale,
  title={Yale-CMU-Berkeley dataset for robotic manipulation research},
  author={Calli, Berk and Singh, Arjun and Bruce, James and Walsman, Aaron and Konolige, Kurt and Srinivasa, Siddhartha and Abbeel, Pieter and Dollar, Aaron M},
  journal={The International Journal of Robotics Research},
  volume={36},
  number={3},
  pages={261--268},
  year={2017},
  publisher={SAGE Publications Sage UK: London, England}
}

@inproceedings{qi2017pointnet,
  title={Pointnet: Deep learning on point sets for 3d classification and segmentation},
  author={Qi, Charles R and Su, Hao and Mo, Kaichun and Guibas, Leonidas J},
  booktitle={Proceedings of the IEEE conference on computer vision and pattern recognition},
  pages={652--660},
  year={2017}
}

@incollection{mamou2016volumetric,
  title={Volumetric hierarchical approximate convex decomposition},
  author={Mamou, Khaled and Lengyel, E and Peters, A},
  booktitle={Game Engine Gems 3},
  pages={141--158},
  year={2016},
  publisher={AK Peters}
}

@ARTICLE{Wright2014mci,
  title    = "Reducing Abnormal Muscle Coactivation After Stroke Using a
              Myoelectric-Computer Interface: A Pilot Study",
  author   = "Wright, Zachary A and Rymer, W Zev and Slutzky, Marc W",
  journal  = "Neurorehabil. Neural Repair",
  volume   =  28,
  number   =  5,
  pages    = "443--451",
  year     =  2014,
  language = "en"
}

@ARTICLE{Ingemanson2019somatosensory,
  title    = "Somatosensory system integrity explains differences in treatment
              response after stroke",
  author   = "Ingemanson, Morgan L and Rowe, Justin R and Chan, Vicky and
              Wolbrecht, Eric T and Reinkensmeyer, David J and Cramer, Steven C",
  journal  = "Neurology",
  volume   =  92,
  number   =  10,
  pages    = "e1098--e1108",
  year     =  2019,
  language = "en"
}

@ARTICLE{Madduri2023coadaptation,
  title     = "Biosignal-based co-adaptive user-machine interfaces for motor
               control",
  author    = "Madduri, Maneeshika M and Burden, Samuel A and Orsborn, Amy L",
  journal   = "Curr. Opin. Biomed. Eng.",
  publisher = "Elsevier BV",
  volume    =  27,
  pages     =  100462,
  year      =  2023,
  language  = "en"
}

@ARTICLE{Proulx2022interaction,
  title     = "Somesthetic, visual, and auditory feedback and their interactions
               applied to upper limb neurorehabilitation technology: A narrative
               review to facilitate contextualization of knowledge",
  author    = "Proulx, Camille E and Louis Jean, Manouchka T and Higgins,
               Johanne and Gagnon, Dany H and Dancause, Numa",
  journal   = "Front. Rehabil. Sci.",
  publisher = "Frontiers Media SA",
  volume    =  3,
  pages     =  789479,
  year      =  2022,
}

@ARTICLE{Hu2023coadaptation,
  title     = "Bridging human-robot co-adaptation via biofeedback for continuous
               myoelectric control",
  author    = "Hu, Xuhui and Song, Aiguo and Zeng, Hong and Wei, Zhikai and
               Deng, Hanjie and Chen, Dapeng",
  journal   = "IEEE Robot. Autom. Lett.",
  publisher = "Institute of Electrical and Electronics Engineers (IEEE)",
  volume    =  8,
  number    =  12,
  pages     = "8573",
  year      =  2023
}

@inproceedings{jia2024dynamic,
  title={Dynamic grasping with a learned meta-controller},
  author={Jia, Yinsen and Xu, Jingxi and Jayaraman, Dinesh and Song, Shuran},
  booktitle={2024 IEEE 20th International Conference on Automation Science and Engineering (CASE)},
  pages={3608--3615},
  year={2024},
  organization={IEEE}
}

@inproceedings{aoyama2024asynchronously,
  title={Asynchronously assigning, monitoring, and managing assembly goals in virtual reality for high-level robot teleoperation},
  author={Aoyama, Shutaro and Liu, Jen-Shuo and Wang, Portia and Jain, Shreeya and Wang, Xuezhen and Xu, Jingxi and Song, Shuran and Tversky, Barbara and Feiner, Steven},
  booktitle={2024 IEEE Conference Virtual Reality and 3D User Interfaces (VR)},
  pages={450--460},
  year={2024},
  organization={IEEE}
}

@article{xu2021active,
  title={Active multitask learning with committees},
  author={Xu, Jingxi and Tang, Da and Jebara, Tony},
  journal={arXiv preprint arXiv:2103.13420},
  year={2021}
}

@article{wang2025reactemg,
  title={ReactEMG: Zero-Shot, Low-Latency Intent Detection via sEMG},
  author={Wang, Runsheng and Zhu, Xinyue and Chen, Ava and Xu, Jingxi and Winterbottom, Lauren and Nilsen, Dawn M and Stein, Joel and Ciocarlie, Matei},
  journal={arXiv preprint arXiv:2506.19815},
  year={2025}
}

@inproceedings{lee2025fabric,
  title={Fabric Sensing of Intrinsic Hand Muscle Activity},
  author={Lee, Katelyn and Wang, Runsheng and Chen, Ava and Winterbottom, Lauren and Leung, Ho Man Colman and DiSalvo, Lisa Maria and Xu, Iris and Xu, Jingxi and Nilsen, Dawn M and Stein, Joel and others},
  booktitle={2025 International Conference On Rehabilitation Robotics (ICORR)},
  pages={1233--1238},
  year={2025},
  organization={IEEE}
}
